\documentclass[10pt,twocolumn,letterpaper]{article}

\usepackage{cvpr}              %

\usepackage[dvipsnames]{xcolor}

\def\naive{na\"{\i}ve}

\def\eg{\emph{e.g.}}

\def\ie{\emph{i.e.}}

\usepackage{tikz}
\usetikzlibrary{spy}
\usepackage{multirow}

\definecolor{cvprblue}{rgb}{0.21,0.49,0.74}
\usepackage[pagebackref,breaklinks,colorlinks,citecolor=cvprblue]{hyperref}

\def\paperID{*****} %
\def\confName{CVPR}
\def\confYear{2024}

\title{Consolidating Attention Features for Multi-view Image Editing}

\author{
Or Patashnik$^1$ \hspace{6mm}  Rinon Gal$^{1,2}$ \hspace{6mm}  Daniel Cohen-Or$^1$ \hspace{6mm} Jun-Yan Zhu$^3$ \hspace{6mm}  Fernando De la Torre$^3$ \\[4pt]
$^1$Tel Aviv University \hspace{10mm} $^2$NVIDIA \hspace{10mm} $^3$Carnegie Mellon University
\\[-10pt]
}

\begin{document}

\twocolumn[{%
	\renewcommand\twocolumn[1][]{#1}%
	\maketitle
	\begin{center}
    \includegraphics[width=0.95\textwidth]{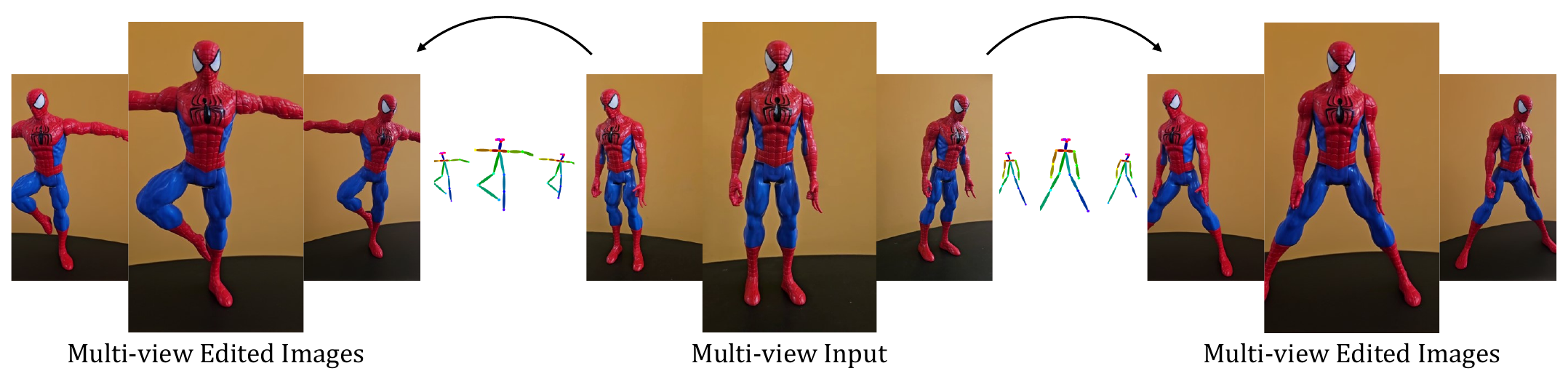}
    \vspace{-8pt}
    \captionof{figure}{Given an object-centric multi-view image set (center), we edit all images simultaneously (left and right), using 3D geometric control, such as changing the body skeleton. 
To promote consistency across different views, we leverage an image diffusion model and introduce QNeRF, a query feature space neural radiance field, to 
progressively consolidate attention features during the generation process.}
    \label{fig:teaser}
	\end{center}
}]

\begin{abstract}
\vspace{-14pt}
Large-scale text-to-image models enable a wide range of image editing techniques, using text prompts or even spatial controls.
However, applying these editing methods to multi-view images depicting a single scene leads to 3D-inconsistent results.
In this work, we focus on spatial control-based geometric 
manipulations and introduce a method to consolidate the editing process across various views. 
We build on two insights: (1) maintaining consistent features throughout the generative process helps attain consistency in multi-view editing, and (2) the queries in self-attention layers significantly influence the image structure. Hence, we propose to improve the geometric consistency of the edited images by enforcing the consistency of the queries.
To do so, we introduce QNeRF, a neural radiance field trained on the internal query features of the edited images. Once trained, QNeRF can render 3D-consistent queries, which are then softly injected back into the self-attention layers during generation, greatly improving multi-view consistency.
We refine the process through a progressive, iterative method that better consolidates queries across the diffusion timesteps.
We compare our method to a range of existing techniques
and demonstrate that it can achieve better multi-view consistency and higher fidelity to the input scene.
These advantages allow us to train NeRFs with fewer visual artifacts, that are better aligned with the target geometry.
\end{abstract}

\section{Introduction}

\begin{figure}
    \setlength{\belowcaptionskip}{0pt}
    \centering
    \setlength{\tabcolsep}{-2.0pt}
    \small{
    \begin{tabular}{c c c c c}
    
        \raisebox{22pt}{\rotatebox[origin=t]{90}{Original}} \hspace{1pt} &
        \begin{tikzpicture}[spy using outlines={}]
            \node {\includegraphics[width=0.20\columnwidth]{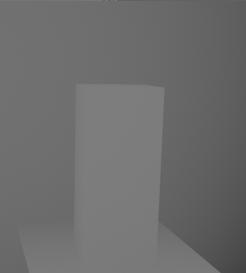}};
        \end{tikzpicture} &
        \begin{tikzpicture}[spy using outlines={}]
            \node {\includegraphics[width=0.20\columnwidth]{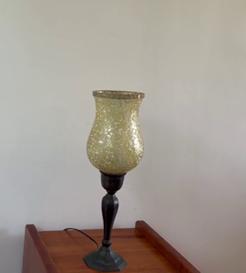}};
        \end{tikzpicture} &
        \begin{tikzpicture}[spy using outlines={}]
            \node {\includegraphics[width=0.20\columnwidth]{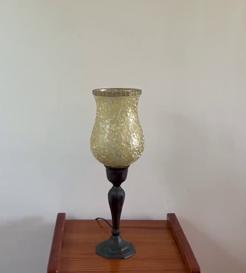}};
        \end{tikzpicture} &

        \begin{tikzpicture}[spy using outlines={}]
            \node {\includegraphics[width=0.20\columnwidth]{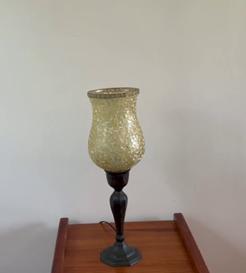}};
        \end{tikzpicture} \\[-8pt]

        \raisebox{23pt}{\rotatebox[origin=t]{90}{MasaCtrl}} \hspace{1pt} &
        \begin{tikzpicture}[spy using outlines={}]
            \node {\includegraphics[width=0.20\columnwidth]{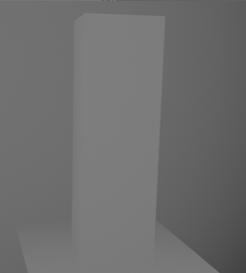}};
        \end{tikzpicture} &
        \begin{tikzpicture}[spy using outlines={circle,yellow,magnification=2,size=0.6cm, connect spies}]
            \node {\includegraphics[width=0.20\columnwidth]{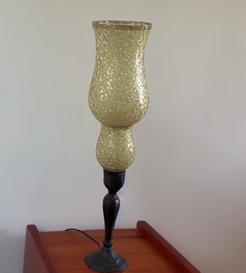}};
        \end{tikzpicture} &

        \begin{tikzpicture}[spy using outlines={circle,yellow,magnification=2,size=0.6cm, connect spies}]
            \node {\includegraphics[width=0.20\columnwidth]{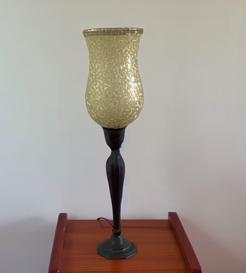}};
        \end{tikzpicture} &

        \begin{tikzpicture}[spy using outlines={circle,yellow,magnification=2,size=0.6cm, connect spies}]
            \node {\includegraphics[width=0.20\columnwidth]{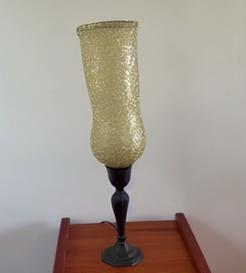}};
        \end{tikzpicture} \\[-6pt]

        \raisebox{33pt}{\rotatebox[origin=t]{90}{Original}} \hspace{1pt} &
        \begin{tikzpicture}[spy using outlines={}]
            \node {\includegraphics[width=0.20\columnwidth]{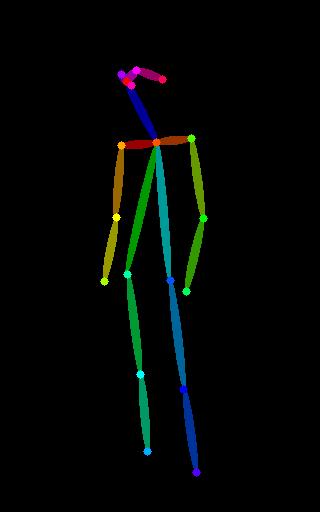}};
        \end{tikzpicture} &
        \begin{tikzpicture}[spy using outlines={}]
            \node {\includegraphics[width=0.20\columnwidth]{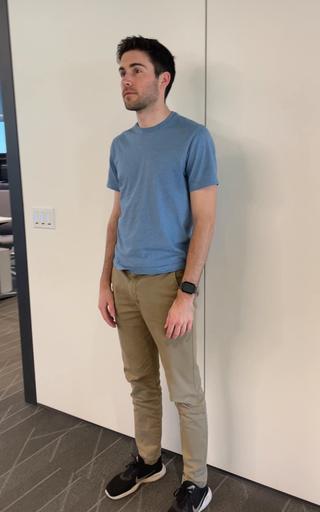}};
        \end{tikzpicture} &
        \begin{tikzpicture}[spy using outlines={}]
            \node {\includegraphics[width=0.20\columnwidth]{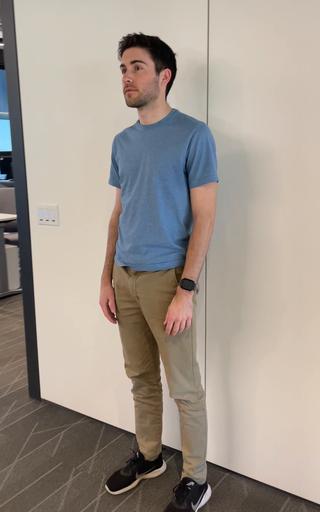}};
        \end{tikzpicture} &

        \begin{tikzpicture}[spy using outlines={}]
            \node {\includegraphics[width=0.20\columnwidth]{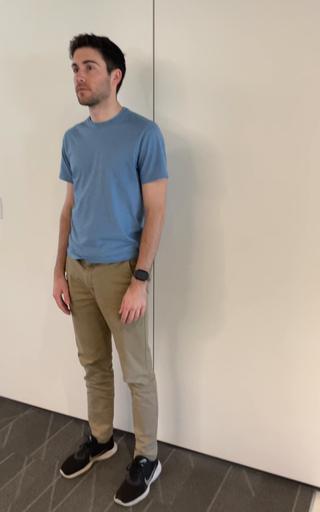}};
        \end{tikzpicture} \\[-8pt]

        \raisebox{35pt}{\rotatebox[origin=t]{90}{MasaCtrl}} \hspace{1pt} &
        \begin{tikzpicture}[spy using outlines={}]
            \node {\includegraphics[width=0.20\columnwidth]{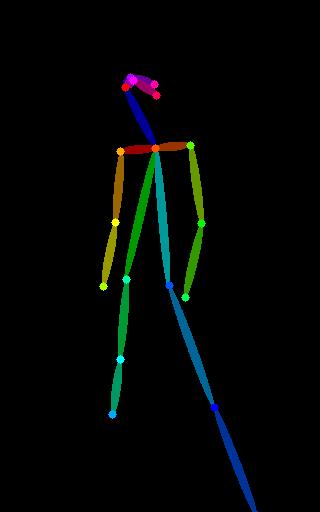}};
        \end{tikzpicture} &
        \begin{tikzpicture}[spy using outlines={circle,yellow,magnification=2,size=0.6cm, connect spies}]
            \node {\includegraphics[width=0.20\columnwidth]{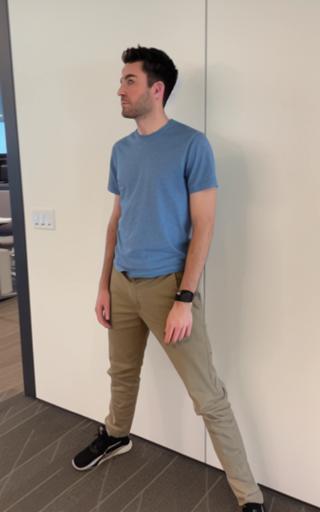}};
            \spy on (-0.18,-0.95) in node [left] at (0.8, 0.9);
        \end{tikzpicture} &

        \begin{tikzpicture}[spy using outlines={circle,yellow,magnification=2,size=0.6cm, connect spies}]
            \node {\includegraphics[width=0.20\columnwidth]{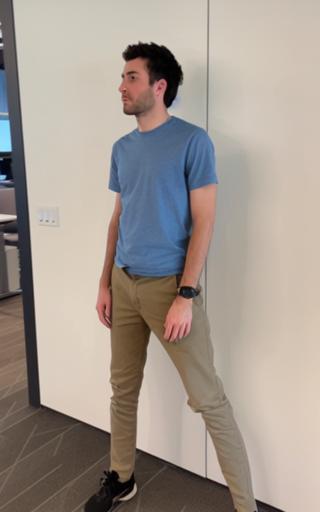}};
            \spy on (-0.12,-1.122) in node [left] at (0.8, 0.9);
        \end{tikzpicture} &

        \begin{tikzpicture}[spy using outlines={circle,yellow,magnification=2,size=0.6cm, connect spies}]
            \node {\includegraphics[width=0.20\columnwidth]{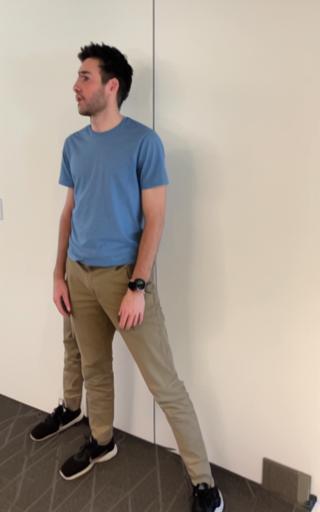}};
        \end{tikzpicture} \\[-6pt]

         & Control & View 1 & View 2 & View 3

    \end{tabular}
    }
    \vspace{-5pt}
    \caption{
    The first and third rows show images captured from different viewpoints.
    When these are individually edited using ControlNet~\cite{zhang2023adding} and MasaCtrl~\cite{cao2023masactrl}, inconsistencies arise. Note the shape of the lamp (top) or the distance of the foot from the wall (bottom). Images were edited using 2D controls projected from a shared 3D model (skeleton, box).
    The leftmost column shows controls corresponding to view 1.
    }
    \vspace{-12pt}
    \label{fig:motivation}
\end{figure}

The advent of large-scale text-to-image models has led to rapid advancements in image editing techniques. Commonly, such techniques are used to modify a \textit{single} image by leveraging the rich visual and semantic prior found in a pre-trained text-to-image diffusion model. However, when considering \textit{sets} of images depicting a shared scene, \naive{} applications of such methods lead to inconsistent edits across the set (see Figure \ref{fig:motivation}). 

\begin{figure}[b]
    \centering
    \includegraphics[width=0.9\linewidth]{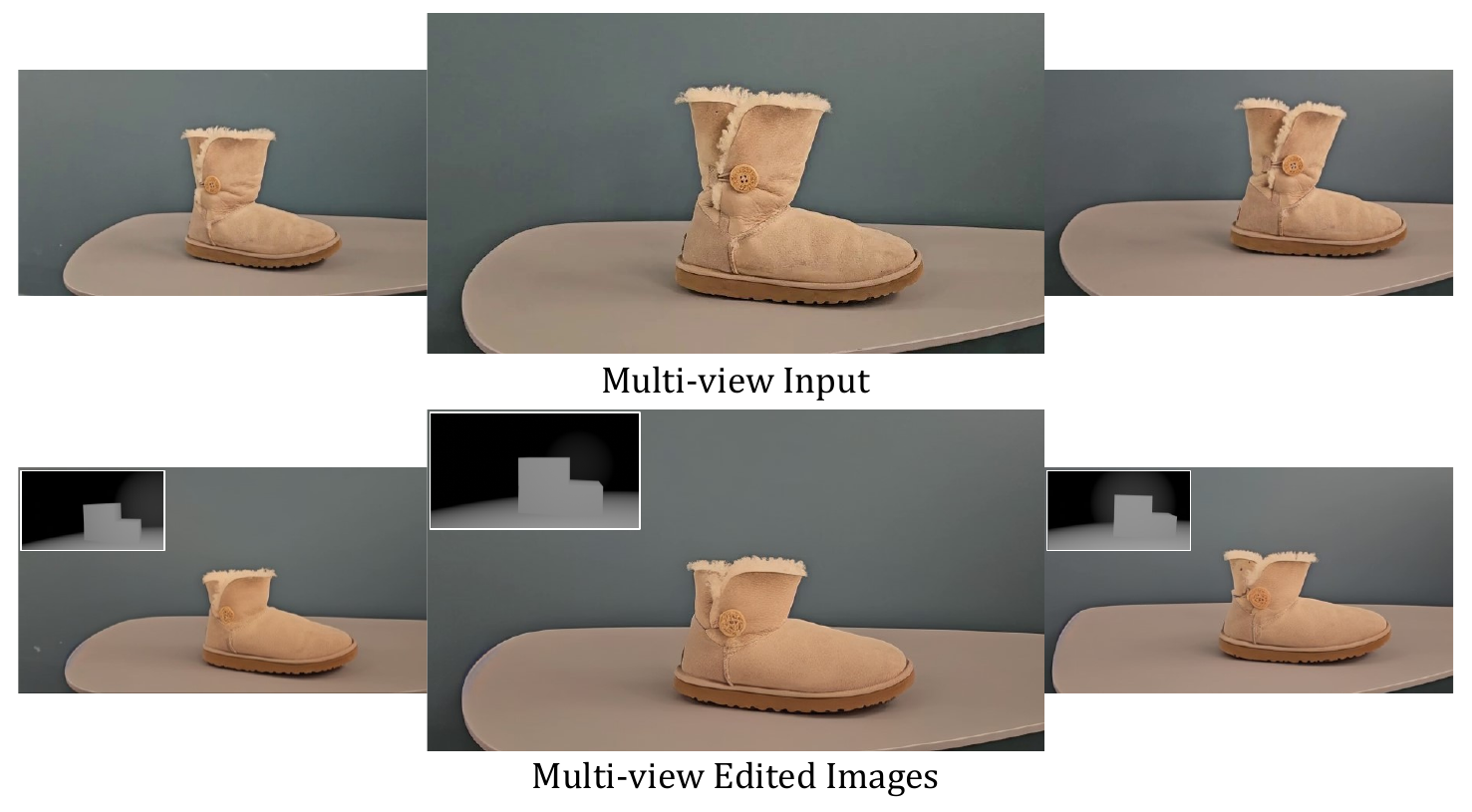}
    \vspace{-8pt}
    \caption{Editing multi-view images of a boot, with a loose depth map~\cite{bhat2023loosecontrol}. We show a sample of three images from the set.}
    \label{fig:teaser-shoe}
\end{figure}

In the realm of multi-view editing, where the image set depicts a single object observed from multiple directions, a recent line of work proposes to leverage the inherent consistency of 3D representations~\cite{mildenhall2020nerf} as a means to consolidate the edits into a more 3D-consistent set~\cite{instructnerf2023}. In practice, existing methods assume that the edits performed are small enough that the underlying 3D representation can successfully average over any inconsistent changes. An assumption that holds well for simpler texture or appearance changes, but fails when dealing with more complex geometric changes. As such, these methods can be used to 
change the style of a person's portrait into a painting, 
but they struggle to make him raise his hands.

In this work, we present an approach for consistent multi-view image editing, focusing on articulations and shape changes, as shown in Figures~\ref{fig:teaser} and \ref{fig:teaser-shoe}.
We use ControlNet~\cite{zhang2023adding}, which was trained to take rough spatial controls (\eg, body skeletons or loose depth maps~\cite{bhat2023loosecontrol}) as an input, and synthesize images aligned with them. Conditioning the generation of the images on these rough controls provides the model with a preliminary understanding of the edited image's coarse geometry. However, relying solely on this coarse geometry signal falls short of attaining high consistency among the edited images. 

\begin{figure}
    \centering
    \includegraphics[width=\columnwidth]{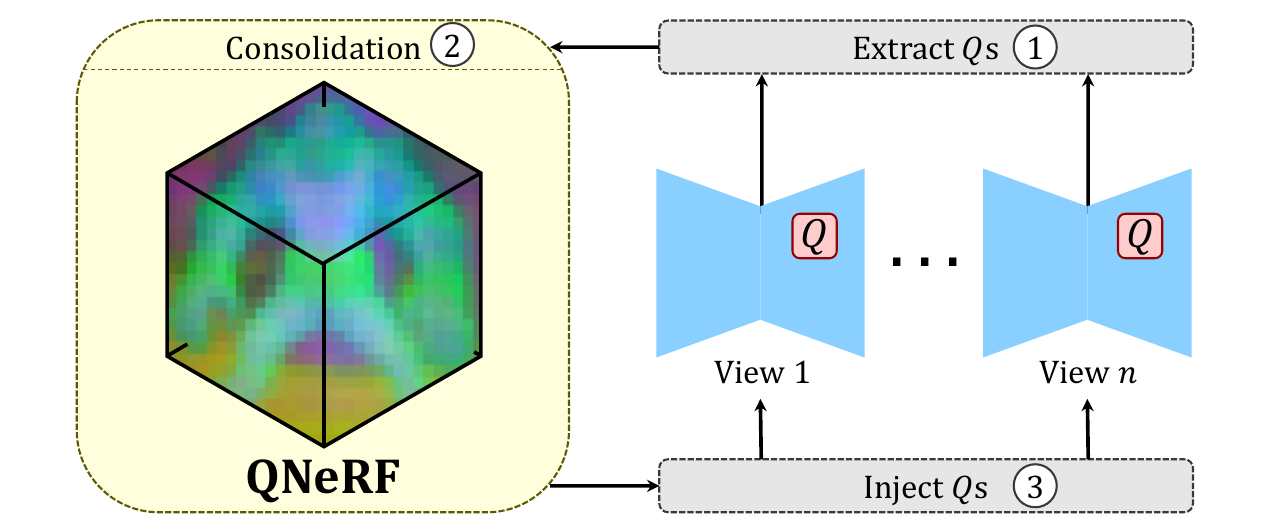}
    \caption{
    We simultaneously generate multi-view edited images with a diffusion model. To consolidate the images, along the denoising process we (1) extract self-attention queries from the network, (2) train a NeRF (termed QNeRF) on the extracted queries and render consolidated queries, and (3) softly inject the rendered queries back to the network for each view. We repeat these steps throughout the denoising process.
    }
    \vspace{-12pt}
    \label{fig:qNerf}
\end{figure}

Our key idea is to encourage the features of ControlNet to be consistent during the generation of the multi-view edited images. As shown by recent works~\cite{tokenflow2023}, increasing the consistency of internal features can help improve the consistency of edited frames in video generation. In particular, we observe that the queries of the self-attention layers within the diffusion model significantly influence the structure of the output image. 
Hence, we propose consolidating the queries of all generated images into a 3D-consistent shared representation, by training a neural field in query feature space, which we term QNeRF.  
We then use the queries rendered from the QNeRF to guide the generation of the edited images, increasing the consistency of the edited multi-view images.
The process of training the QNeRF and using the rendered queries is a progressive process interleaved during the denoising process. As illustrated in Figure \ref{fig:qNerf}, the updated QNeRF is trained with the extracted queries, and the rendered queries guide the features of the generated images within the diffusion network.  

We demonstrate that our approach enables a wide range of articulation and shape-based modifications, while achieving greater visual quality than alternative consistency-preserving approaches. These results are validated through qualitative evaluations, as well as automated metrics and user preference scores. Finally, we demonstrate that the underlying geometry of NeRFs trained on our results exhibits better alignment with the target controls.

\section{Related Work}

\paragraph{Image Editing with Diffusion Models.}
Advancements in large-scale diffusion models have significantly enhanced image editing techniques~\cite{meng2022sdedit, kawar2023imagic, brooks2022instructpix2pix, brack2023sega, hubermanspiegelglas2023edit}. Specifically, the manipulation of internal representations within the diffusion model during the denoising process has been shown to enable high-quality and semantically meaningful edits~\cite{hertz2022prompt, Tumanyan_2023_CVPR, patashnik2023localizing, Parmar_2023, epstein2023selfguidance, ge2023expressive,tokenflow2023}. 
Notably, some recent works focus on self-attention layers, and leverage the roles of queries, keys, and values within self-attention to obtain various edits~\cite{cao2023masactrl, alaluf2023crossimage, hertz2023StyleAligned}. While the above works focus on editing a \emph{single} image, we build on the functionality of self-attention components to achieve consistent multi-view image editing.

\paragraph{Multi-view Data Synthesis and Editing.}
The rich prior embedded in large-scale image diffusion models has paved the way for a new 3D model editing and synthesis paradigm. In this approach, images corresponding to different camera views are edited or generated and then used to construct a 3D model that consolidates them.
In Instruct-NeRF2NeRF~\cite{instructnerf2023}, the authors propose to iteratively update the dataset used to train some initial NeRF. In each update iteration, an image from the dataset is edited with InstructPix2Pix, and then the NeRF is updated accordingly to consolidate the different views. This dataset update approach was widely adopted in follow-up works~\cite{khalid2023latenteditor, shum2023languagedriven, song2023efficient,kim2023collaborative}.
In texturing work, a recent approach is to render a given 3D mesh from multiple views, use a text-guided model to generate UV maps for these views, and then apply them to the 3D model~\cite{Richardson_2023,cao2023texfusion}. This process is performed iteratively with overlapping views, eventually leading to consistent texturing. 
Similarly, Kapon et al.~\cite{kapon2023mas} recently used a human diffusion model to generate 2D views, while employing a 3D representation to ensure consistency.
Instead of editing a given dataset, a concurrent work~\cite{wu2023reconfusion} reconstructs a scene from a few images by generating 3D-consistent fake views. 

Our work also consolidates 2D edits across multiple views. However, we also change the underlying geometry. Moreover, we consolidate the views by training a NeRF on attention features, rather than working in pixel-space. Hence, we are more aligned with the real-image manifold, and less prone to pixel-space averaging artifacts.

Rather than performing the edit in image space, another line of work operates directly on 3D representations.
In particular, with the rapid improvement in implicit 3D representation~\cite{mildenhall2020nerf, kerbl3Dgaussians}, 3D editing methods have emerged~\cite{zhuang2023dreameditor, liu2021editing, yuan2022nerf, feng2023pienerf, sella2023voxe, Koo:2023PDS}. Recent works employ advances in text-based image editing to edit an implicit 3D representation with text~\cite{sella2023voxe, instructnerf2023, bao2023sine, Wang_2022_CLIPNerf, wang2022nerf}. However, these works are guided by prompts, and hence cannot easily handle control-based guidance. Indeed, they typically focus on appearance and small structural modifications, while we target large geometric changes.

Finally, some works perform shape edits by extracting a mesh from the NeRF, editing it with classical techniques, and then deforming the NeRF accordingly~\cite{neumesh, yuan2022nerf, xu2022deforming}. However, such techniques require expert knowledge and may not be as widely applicable as simple control-based methods.

\vspace{-12pt}

\paragraph{Feature NeRFs.}
Previous works showed that NeRFs can represent not only RGB images but also semantic latent features. Some works~\cite{kobayashi2022distilledfeaturefields, lerf2023, tschernezki22neural} distill semantic 2D features into NeRFs, allowing one to obtain semantic 3D information. Other works~\cite{Niemeyer2020GIRAFFE, Chan2021} show that features rendered from NeRFs can be employed for consistent multi-view image generation. In this work, we distill the attention features of a diffusion model into a NeRF and use rendered features during the denoising process to achieve multi-view editing.

\section{Preliminaries}

\subsection{Self-Attention in Diffusion Models}
Recent diffusion models are typically implemented as a UNet~\cite{ronneberger2015u} consisting of cross-attention, self-attention, and convolutional layers. Previous works studied the roles of these components, focusing on attention layers. In our work, we focus on the queries, keys, and values of \textit{self-attention} layers.
Specifically, it has been shown that each query in self-attention layers determines the semantic meaning of the pixel that corresponds to it~\cite{cao2023masactrl, alaluf2023crossimage}. Hence, the queries are associated with the structure of the generated image.
Moreover, the keys and values of self-attention layers determine the appearance of the image, and by using the keys and values of one image in the denoising process of another image, the appearance is transferred~\cite{alaluf2023crossimage}. 
In particular, in MasaCtrl~\cite{cao2023masactrl}, non-rigid edits are applied to an image. To preserve the appearance of the original image, they inject keys and values of self-attention layers from the original image into the generated one. In our method, we employ this technique to preserve the appearance of the original scene.

\subsection{Neural Radiance Fields}
Neural Radiance Field (NeRF) is an implicit 3D representation, parameterized by a network. Given a spatial location $\textbf{x}$ and a viewing direction $\textbf{d}$, the network outputs the density $\sigma(\textbf{x})$ and the RGB value of that location $c(\textbf{x}, \textbf{d})$. 
These can then be used to render an image from a desired viewing direction, using classical volume rendering techniques~\cite{mildenhall2020nerf, max1995optical}. Specifically, given a camera ray $\textbf{r}(t) = \textbf{o} + t\textbf{d}$, the expected color $C(\textbf{r})$ is given by
\begin{equation} \label{eq:vol-render}
    C(\textbf{r}) = 
    \int_{t_n}^{t_f} T(t) \sigma(\textbf{r}(t)) c(\textbf{r}(t), \textbf{d}) dt,
\end{equation}
where
\begin{equation}
    T(t) = \exp \left({- \int_{t_n}^{t} \sigma(\textbf{r}(s)) ds }\right).
\end{equation}
In this work, we train a NeRF on latent representations rather than on RGB values. Following previous works~\cite{lerf2023, kobayashi2022distilledfeaturefields}, we use the same volumetric rendering approach to render latent representations.

\section{Method}

Our method operates on a set of posed images $\{x^v\}_{v=1}^n$ depicting the same scene from multiple viewpoints, along with a set of 2D spatial controls $\{c^v\}_{v=1}^n$ loosely specifying the target geometry of the main object from each view (Figures~\ref{fig:teaser}, \ref{fig:teaser-shoe}). 
These controls are the projection of a low-dimensional 3D model that is easy to manipulate, such as a skeleton or a box. We elaborate on the process of obtaining the edited controls in Appendix~\ref{sec:implementation-details}.

Given these controls, we simultaneously edit all the input images to generate output images ${\{\hat{x}^v}\}^n_{v=1}$. 
These output images should depict the same scene as the input images, with the subject's geometry changed to align with the provided controls.
To edit the images, we leverage a pre-trained Stable Diffusion model~\cite{rombach2021highresolution}, and the MasaCtrl~\cite{cao2023masactrl} approach. There, the images are first inverted with DDIM~\cite{song2021denoising}.
Then, they are re-synthesized following a given control, while preserving the appearance of the original scene by injecting the keys and values of self-attention layers from the original image into the edited one. However, this approach considers each image in isolation, and so the corresponding edit outputs are inconsistent between viewpoints. Our key idea to overcome this hurdle, is to consolidate the edits by improving their multiview consistency in the attention feature space. Specifically, we notice that the inconsistencies are largely in the object shapes. Hence, we propose to align the shapes by consolidating the self-attention queries between the different views. We do so by training a NeRF on the queries during the denoising process, which we term QNeRF. 

A conceptual overview of the consolidation process is illustrated in Figure \ref{fig:qNerf}. There, we depict the parallel networks, each of which denoises a single view. Query features are extracted from the networks and used to train the QNeRF, which consolidates them into a 3D-consistent representation. The consolidated query features are then softly-injected (Section \ref{sec:soft-inj}) back into the denoising network, improving the multi-view consistency of the edited images. 

In practice, we perform the denoising process in intervals. In each interval, we interleave consolidation steps with steps that allow the features to evolve. All consolidation steps in a single interval employ the same trained QNeRF. Additionally, we inject the self-attention keys and values in all steps to preserve the original scene appearance~\cite{cao2023masactrl}.
The details of our QNeRF, feature injection approach, and the structure of intervals are provided below.

\subsection{QNeRF}

The centerpiece of our approach is the QNeRF -- a NeRF~\cite{mildenhall2020nerf} trained on query features extracted from the diffusion model during the denoising process.
The inherent 3D consistency of the QNeRF drives the consolidation of the queries.
Specifically, at the last step of each of our intervals (Section \ref{sec:intervals}), we extract the self-attention queries from the diffusion model along all UNet decoder layers with resolutions $\in$ ($16,32,64$). These yield a total of 9 query sets per denoised-image at a given denoising timestep. These comprise the training set on which we train our QNeRF. 

\begin{figure}
    \centering
    \includegraphics[width=\columnwidth]{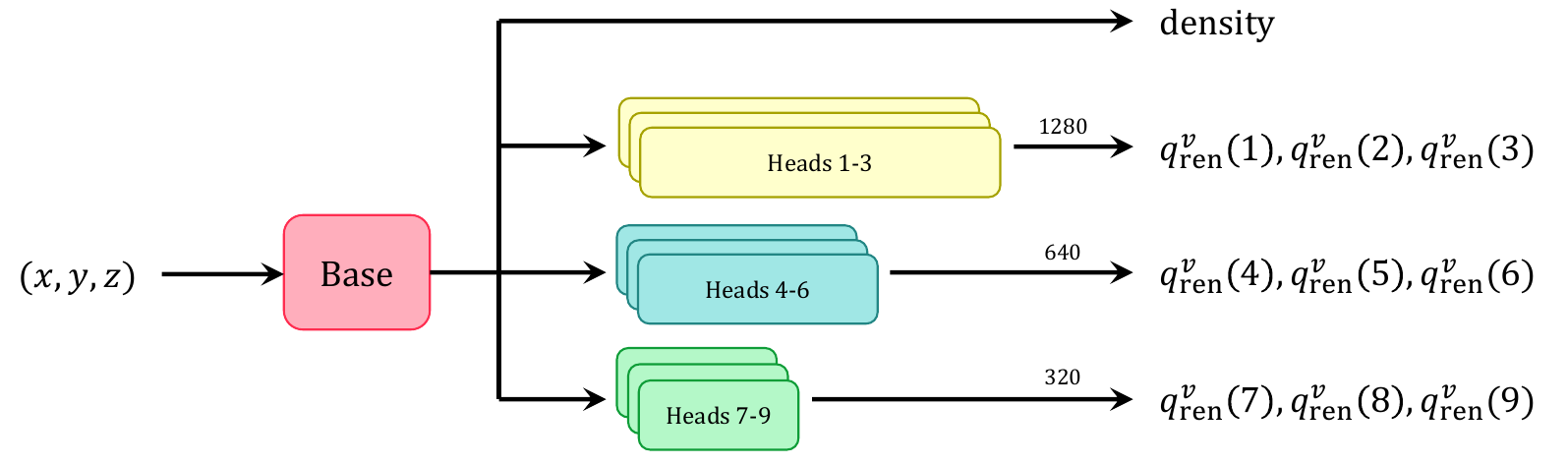}
    \vspace{-10pt}
    \caption{The architecture of QNeRF. Nine heads are attached to the base network, to produce queries corresponding to nine self-attention layers of the diffusion model. Each group of heads corresponds to a self-attention layer of a certain resolution, and the number displayed above the arrow represents the number of channels in that group (1280, 640, 320).}
    \vspace{-12pt}
    \label{fig:qnerf-arch}
\end{figure}

\begin{figure*}
    \centering
    \includegraphics[width=0.95\textwidth]{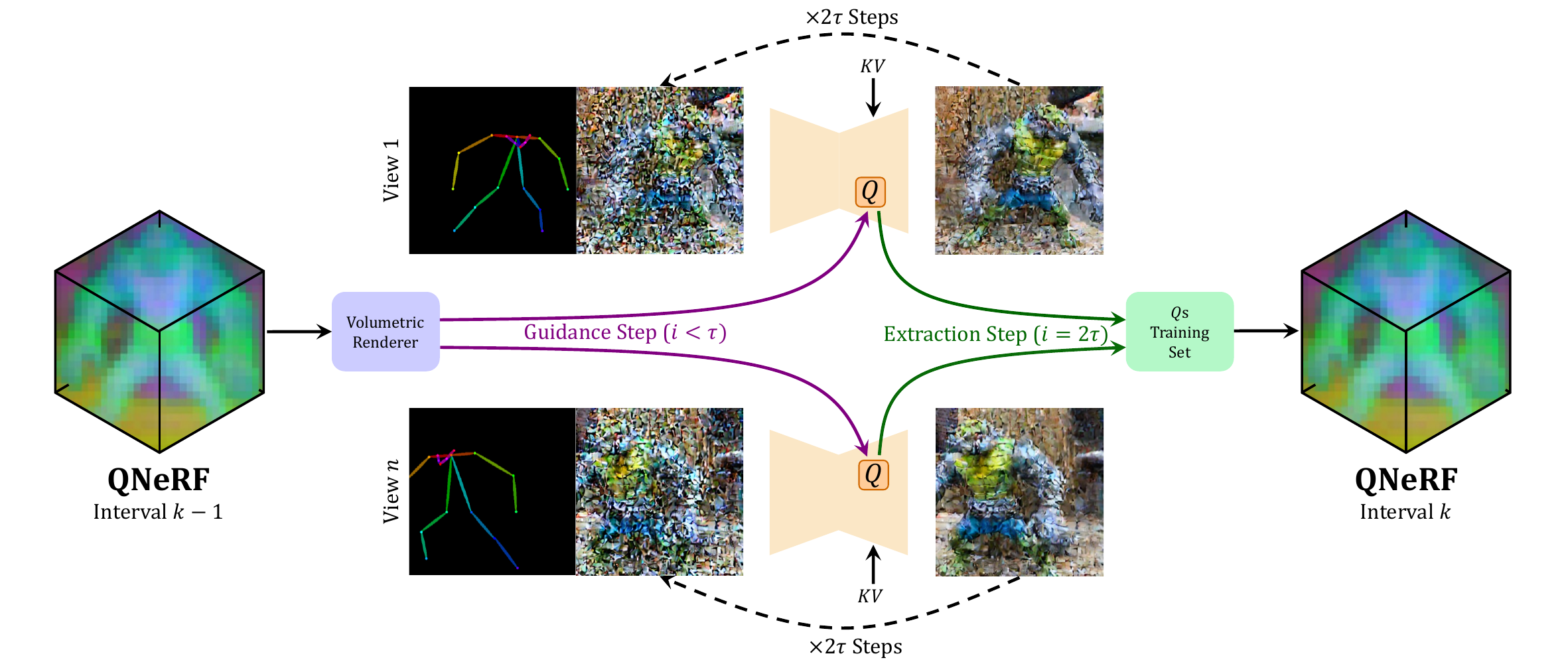}
    \vspace{-5pt}
    \caption{In each multi-view denoising interval we have query-guided steps, followed by steps without guidance. In query-guided steps, we alter the noisy latent code with an objective of proximity between the self-attention queries generated by the latent code, and queries rendered from the QNeRF. At the last step of the interval, we extract the generated queries and use them to train the QNeRF that provides guidance for the next interval. Query guidance consolidates the geometry across the different views. In addition, we inject the keys and values of self-attention layers from the original images to preserve the appearance.}
    \vspace{-9pt}
    \label{fig:method-overview}
\end{figure*}

The QNeRF itself is a depth-nerfacto~\cite{Tancik_2023, mueller2022instant, kangle2021dsnerf}, with a series of adaptations to better fit our use case. First, rather than producing an RGB value for each input coordinate, we output $9$ query values corresponding to the $9$ extracted query layers. We do so by adding $9$ heads to the base nerfacto network, where each head is optimized to output the queries of a specific self-attention layer. Hence, the dimension of each head's output is set to the number of channels in the respective layer's queries. The base nerfacto network predicts the density at each point, as in the original nerfacto architecture. By sharing the density between the different queries, we can better share cross-layer information and stabilize the geometry. Additionally, we omit the dependence of the QNeRF on the viewing direction. This choice embodies the fact that the queries represent geometry, which is not dependent on the viewing direction. The full architecture of our QNeRF is presented in Figure~\ref{fig:qnerf-arch}. To train it, we employ the q-loss:
\begin{equation}
    \mathcal{L}_q = \sum_{\textbf{r} \in \mathcal{R}} \sum_{l}{ \| \hat{\textbf{Q}_{l}}(r) - q^r(l) \|},
\end{equation}
where $\mathcal{R}$ are sampled rays, $q^r(l)$ are the extracted queries corresponding to ray $r$ and layer $l$. $\hat{\textbf{Q}_{l}}(r)$ is defined as in Equation~\ref{eq:vol-render}, where we replace RGB value, $c$, with a self-attention query. Additionally, we use the depth-loss $\mathcal{L}_{\text{depth}}$ proposed by Deng et al.~\cite{kangle2021dsnerf}, and our final loss for training the QNeRF is written as $\mathcal{L} = \mathcal{L}_q + \mathcal{L}_{\text{depth}}$.

Once trained, we can use the QNeRF to render consolidated queries to guide the denoising process. We do so using the standard volumetric rendering technique~\cite{max1995optical, mildenhall2020nerf}.
Finally, since we do not expect the geometry to significantly change between intervals, we initialize each QNeRF from the one trained at the prior interval.

\subsection{Query Guidance} \label{sec:soft-inj}

With the QNeRF in hand, we wish to use consolidated queries produced by it to guide the denoising process. Consider the editing process of a specific frame with a given camera viewpoint. The direct way to use our QNeRF would be to render the queries for this particular viewpoint, and use them to replace the queries naturally created by the UNet during the denoising steps. However, in our initial experiments, we observed that such direct replacement can lead to visual artifacts in the edited frames. Instead, we propose a ``soft-guidance" mechanism inspired by previous works~\cite{dhariwal2021diffusion, chefer2023attendandexcite, Parmar_2023, epstein2023selfguidance}. At each query-guided denoising step, we first do a single forward pass through the UNet and extract all the naturally generated queries. We then perform a single optimization step on the input latents themselves, with the goal of minimizing the distance between these generated queries and the ones rendered from the QNeRF. Formally, the query guidance is defined as:
\begin{equation} \label{eq:q-guidance}
    z_t^v \leftarrow z_t^v - \alpha \nabla_{z_t^v} \sum_l{\| q^v(l) - q^v_{\text{ren}}(l) \|^2},
\end{equation}
where $z_t^v$ is the noisy latent code corresponding to viewpoint $v$ at timestep $t$, and $q^v(l), q^v_{\text{ren}}(l)$ are the generated and rendered self-attention queries of layer $l$, respectively. After this update step, a DDIM~\cite{song2021denoising} denoising step is applied to obtain $z_{t-1}^v$.

\subsection{Multi-view Image Denoising Interval}\label{sec:intervals}

As previously noted, rather than training a QNeRF for every denoising step, we employ an interval-based approach. The structure of each interval is motivated by two observations. On the one hand, it has been observed that in a standard denoising process, the internal UNet features of adjacent denoising timesteps are similar~\cite{li2023faster}, to the extent that their computation can often be skipped and reused. Hence, we expect that guiding several adjacent timesteps with the same query features should not degrade the quality of the results. On the other hand, if we continue to reuse the same queries over an extended number of timesteps, we leave no room for the gradual change that does occur in the diffusion features. In an extreme case, using the same query-guidance for all timesteps would lead to the same queries being used across the entire diffusion path. Ideally, our mechanism should allow the queries to evolve freely along the denoising process.

We thus propose an interleaved process, which breaks image generation into several overlapping intervals. Consider one such interval (illustrated in Figure~\ref{fig:method-overview}), starting at diffusion timestep $T_i$ and spanning $2\tau$ steps. We begin the interval by taking $\tau$ QNeRF-guided steps, using a QNeRF obtained from the end of the previous interval. We store the noisy latents obtained at this point ($T_i - \tau$) for future use. Following these $\tau$ steps, we take another $\tau$ steps where we perform vanilla MasaCtrl editing, without any query guidance. These steps thus allow the queries to evolve freely, matching the more advanced step. At the end of these unguided steps, we extract the updated query features, and use them to optimize a new QNeRF. Finally, we retrieve the latents stored at $T_i - \tau$ and begin a new interval starting from this point, using the updated QNeRF for guidance.

In the special case of the first interval, we do not have a QNeRF at hand. Hence, we perform unguided-editing for $2\tau$ steps (\ie, until $T-2\tau$), train a QNeRF from features extracted at this step, and begin the next interval at $T-\tau$.

\subsection{Progressive Consolidation}
\begin{figure}
    \centering
    \includegraphics[width=0.85\columnwidth]{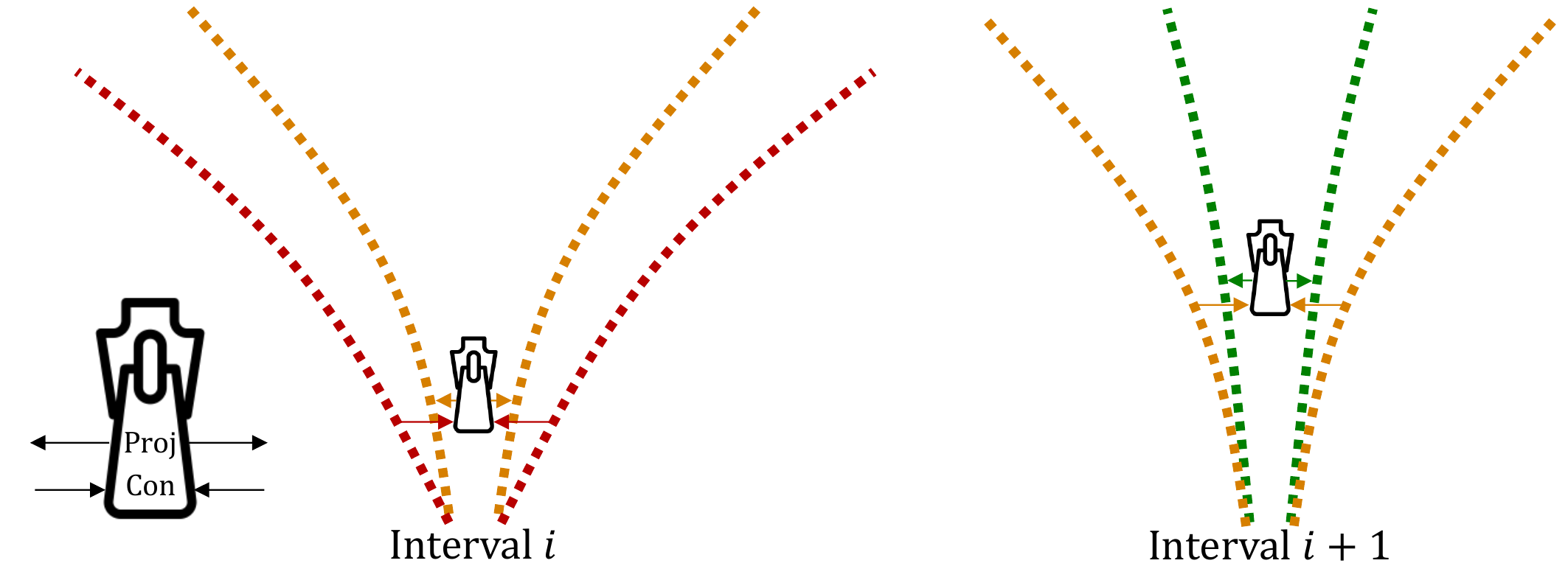}
    \vspace{-6pt}
    \caption{Our progressive denoising process is analogous to a zipper mechanism. Each step of a zipper relies on the closure of all preceding parts, and benefits from the fact that they got closer together. 
    The dotted red curves (left) illustrate the queries generated along the diffusion process of two views. The zipper represents the QNeRF that consolidates the red dots, and projects them to form the orange dots (left) which sit along closer trajectories. After a few steps, we repeat this process with the orange and green curves (right), progressively consolidating the generated queries.}
    \vspace{-12pt}
    \label{fig:zipper}
\end{figure}

The consolidation of the queries along the denoising process is done progressively, by training the QNeRF on-the-fly during intermediate steps of the generation.
Specifically, the queries used to train the QNeRF at each interval, are affected by the consolidated queries rendered from the prior interval's QNeRF.
By combining this approach with the scheduling within each interval, we provide the model with room to freely develop the queries, while ensuring that any injected queries are still the result of consolidated queries formed at the end of the previous interval. Crucially, this prevents the queries from drifting apart too far before they are re-consolidated. To grasp the effectiveness of this progressive process, consider the analogy to a zipper mechanism, as illustrated in Figure~\ref{fig:zipper}. 

\section{Experiments}
Next, we evaluate our method qualitatively and quantitatively, and compare it to other baseline methods. We show results on $7$ multi-view image sets, of $200$-$500$ images each. Of these, 3 sets are used for comparisons. See Appendix~\ref{sec:dataset-details} for additional dataset details.

\subsection{Ablation Studies}
\begin{figure}
    \centering
    \setlength{\tabcolsep}{-2.0pt}
    \small{
    \begin{tabular}{c c c c c c c}
        \raisebox{28pt}{\rotatebox[origin=t]{90}{Original}} \hspace{1pt} &
        \begin{tikzpicture}[spy using outlines={}]
            \node {\includegraphics[width=0.15\columnwidth]{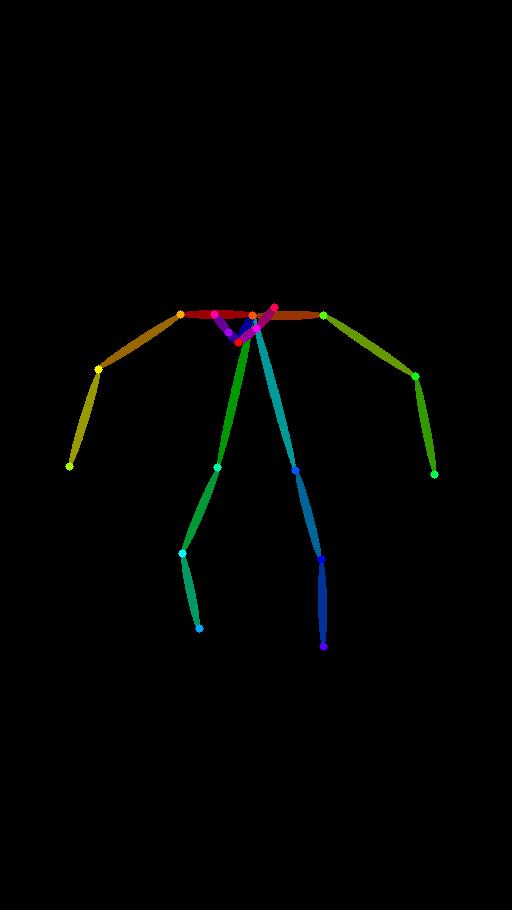}};
        \end{tikzpicture} &
        \begin{tikzpicture}[spy using outlines={}]
            \node {\includegraphics[width=0.15\columnwidth]{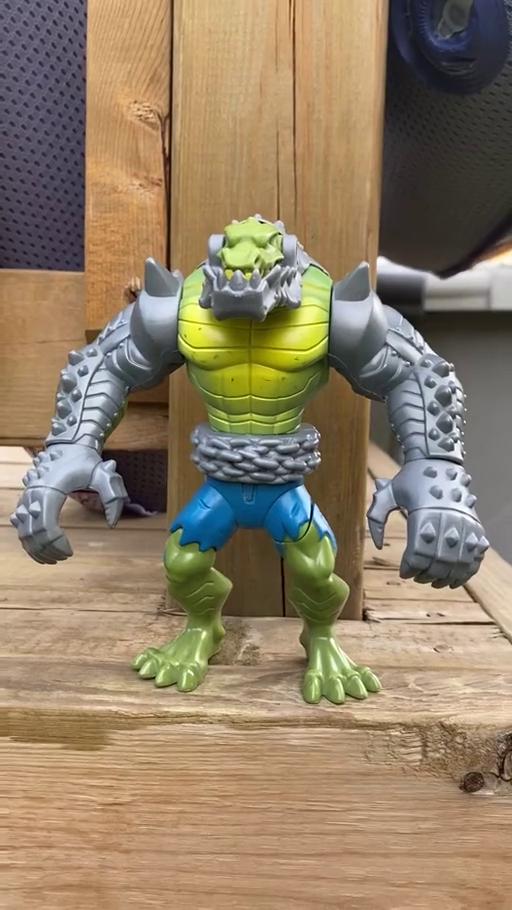}};
        \end{tikzpicture} &
        \begin{tikzpicture}[spy using outlines={}]
            \node {\includegraphics[width=0.15\columnwidth]{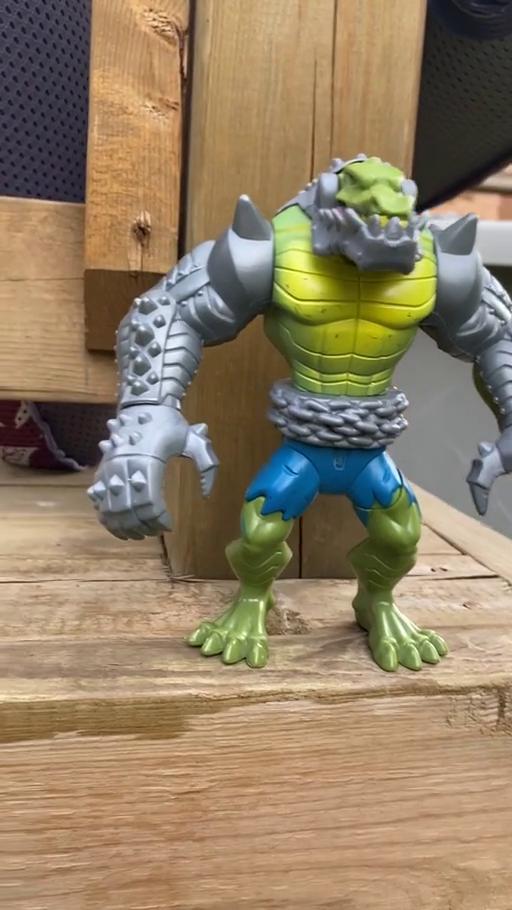}};
        \end{tikzpicture} &

        \begin{tikzpicture}[spy using outlines={}]
            \node {\includegraphics[width=0.15\columnwidth]{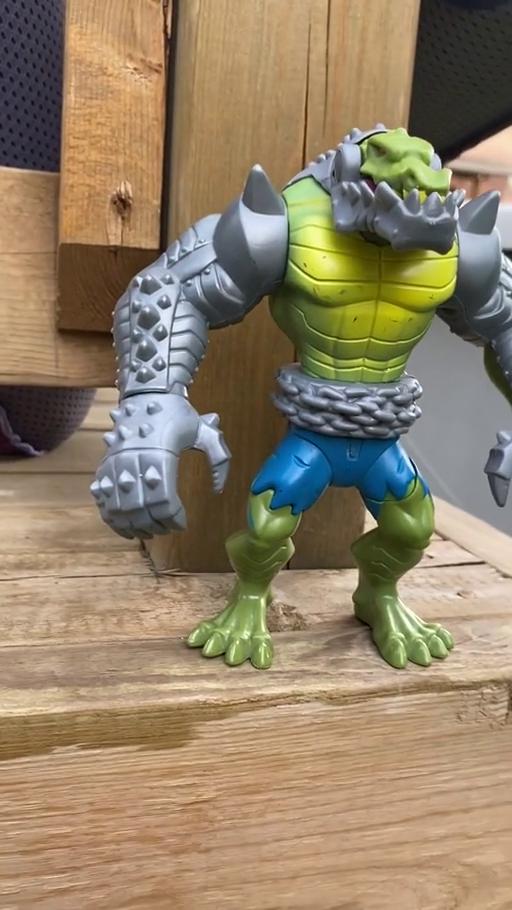}};
        \end{tikzpicture} &

        \begin{tikzpicture}[spy using outlines={}]
            \node {\includegraphics[width=0.15\columnwidth]{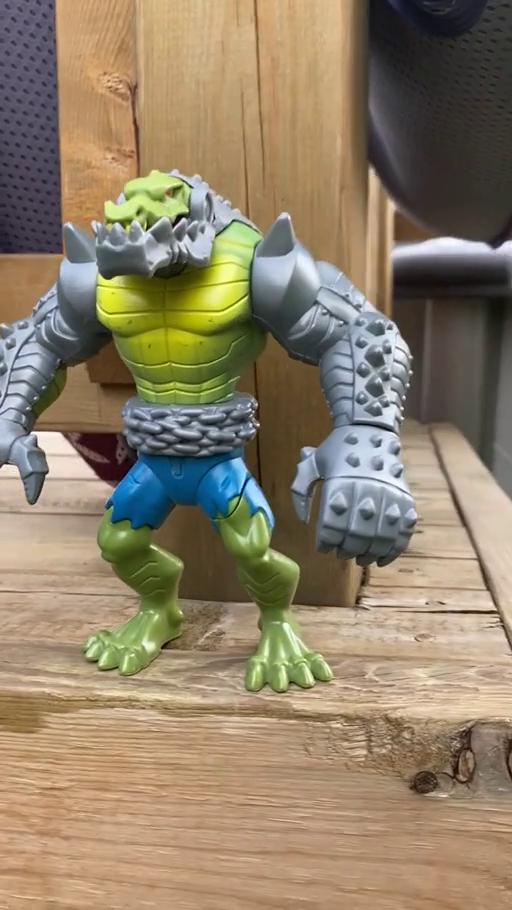}};
        \end{tikzpicture} &

        \begin{tikzpicture}[spy using outlines={}]
            \node {\includegraphics[width=0.15\columnwidth]{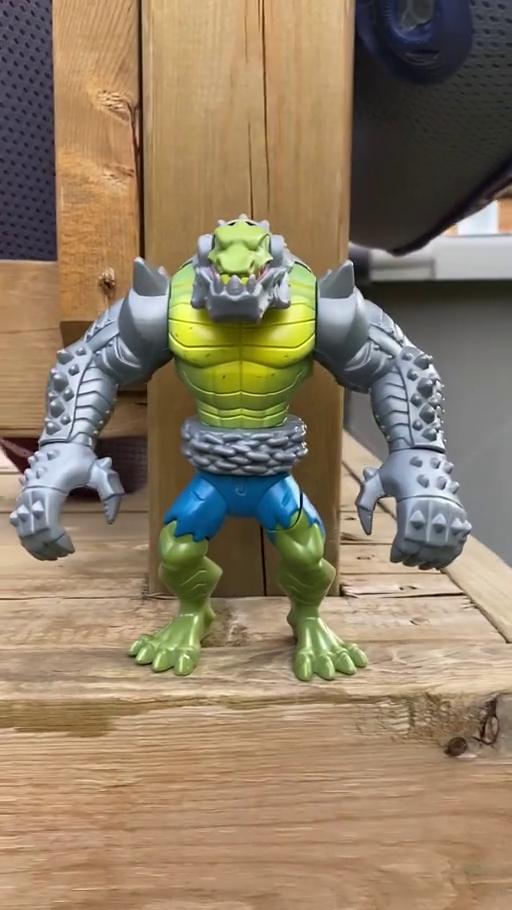}};
        \end{tikzpicture} \\[-8pt]

        \raisebox{30pt}{\rotatebox[origin=t]{90}{MasaCtrl}} \hspace{1pt} &
        \begin{tikzpicture}[spy using outlines={}]
            \node {\includegraphics[width=0.15\columnwidth]{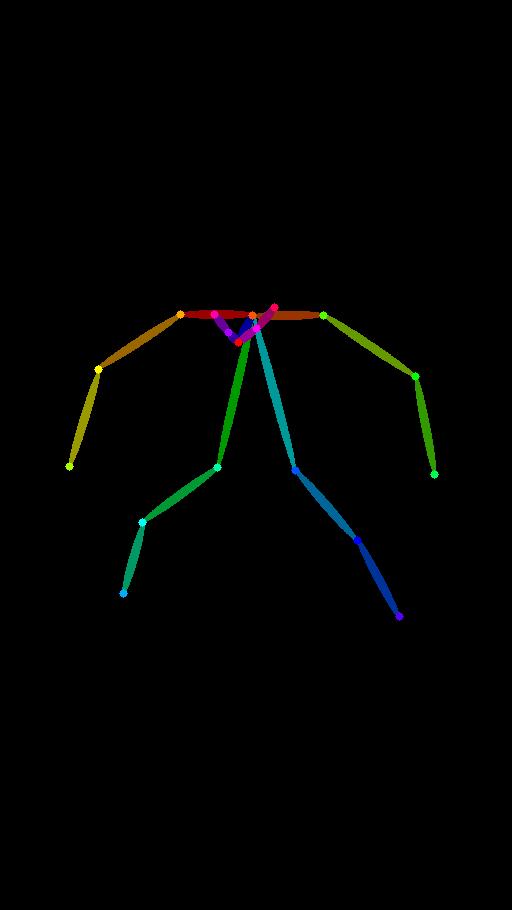}};
        \end{tikzpicture} &
        \begin{tikzpicture}[spy using outlines={circle,yellow,magnification=2,size=0.6cm, connect spies}]
            \node {\includegraphics[width=0.15\columnwidth]{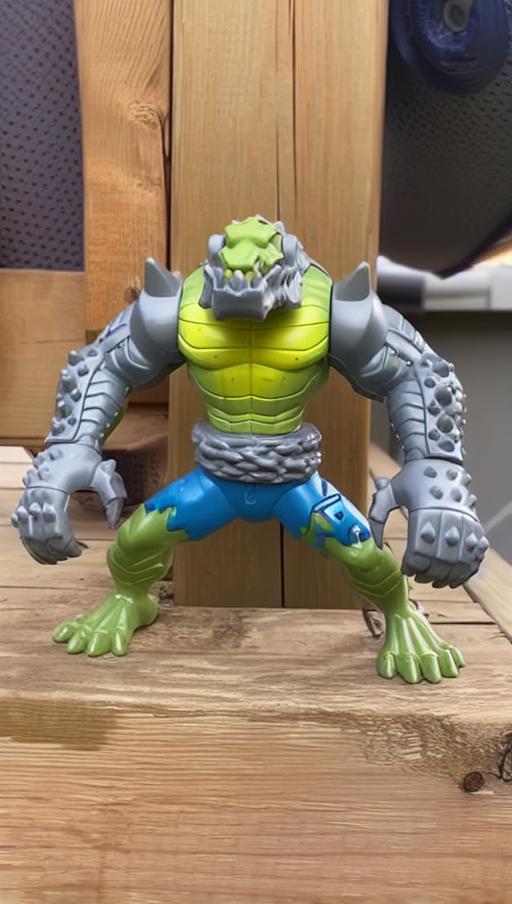}};
            \spy on (0.40,-0.58) in node [left] at (0.12,-0.78);
        \end{tikzpicture} &

        \begin{tikzpicture}[spy using outlines={}]
            \node {\includegraphics[width=0.15\columnwidth]{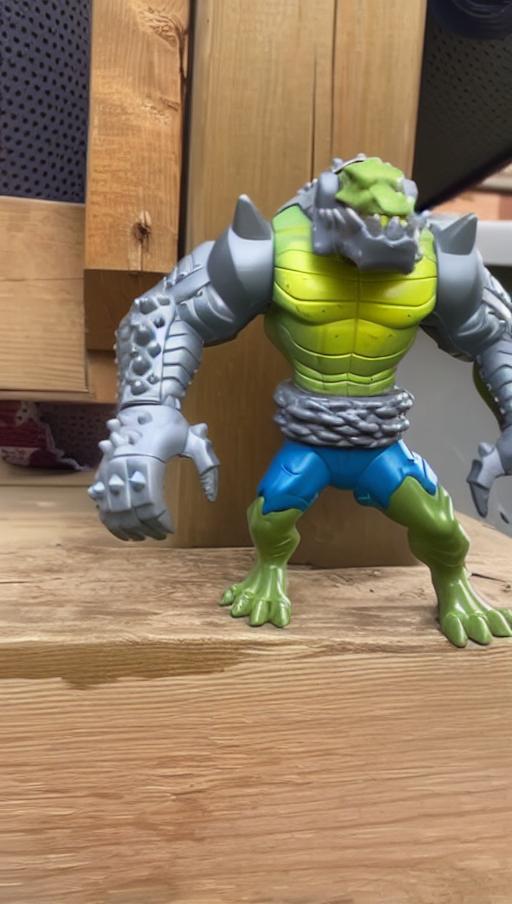}};
        \end{tikzpicture} &

        \begin{tikzpicture}[spy using outlines={circle,yellow,magnification=2,size=0.6cm, connect spies}]
            \node {\includegraphics[width=0.15\columnwidth]{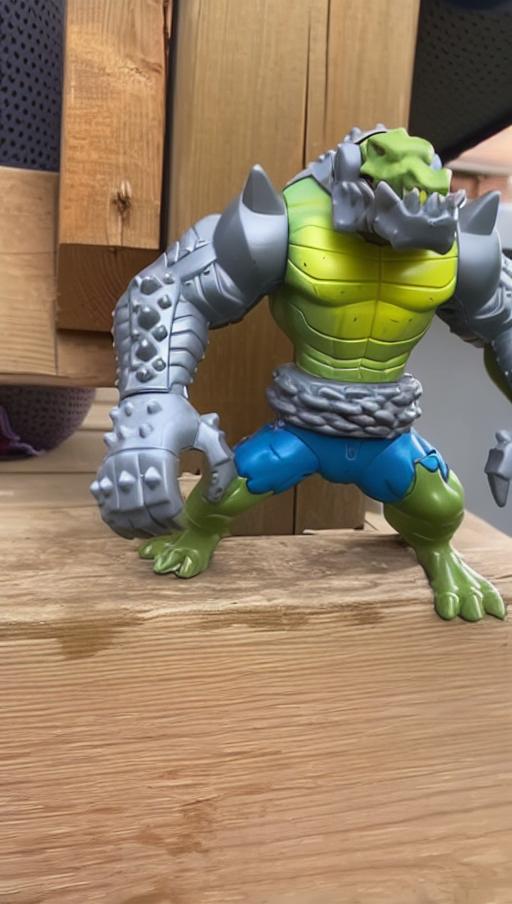}};
            \spy on (0.47,-0.38) in node [left] at (0.12,-0.78);
        \end{tikzpicture} &

        \begin{tikzpicture}[spy using outlines={}]
            \node {\includegraphics[width=0.15\columnwidth]{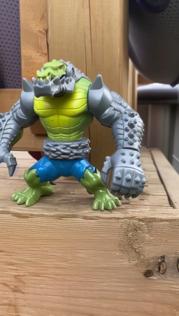}};
        \end{tikzpicture} &

        \begin{tikzpicture}[spy using outlines={circle,yellow,magnification=2,size=0.6cm, connect spies}]
            \node {\includegraphics[width=0.15\columnwidth]{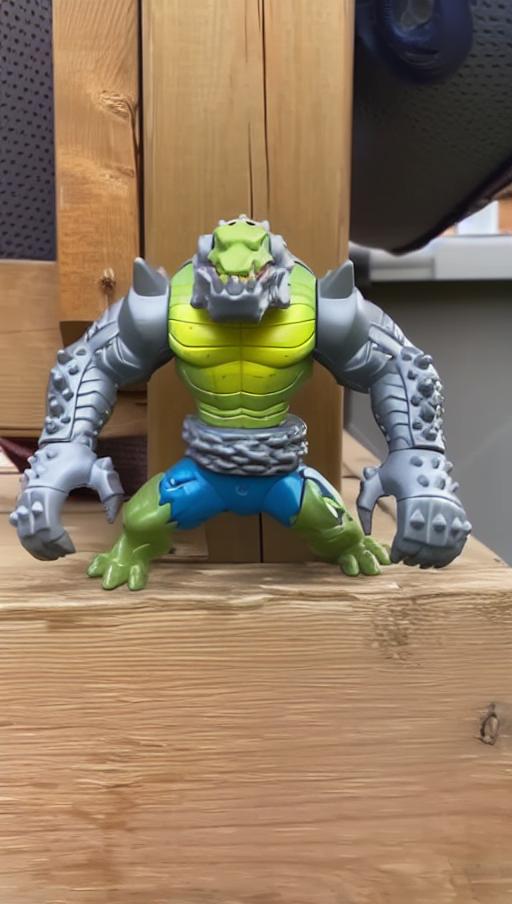}};
            \spy on (0.24,-0.24) in node [left] at (0.12,-0.78);
        \end{tikzpicture} \\[-8pt]
        
        \raisebox{29pt}{\rotatebox[origin=t]{90}{w/o soft-injection}} \hspace{1pt} &
        \begin{tikzpicture}[spy using outlines={}]
            \node {\includegraphics[width=0.15\columnwidth]{images/ablation/skel_01_frame_00011.jpg}};
        \end{tikzpicture} &
        \begin{tikzpicture}[spy using outlines={}]
            \node {\includegraphics[width=0.15\columnwidth]{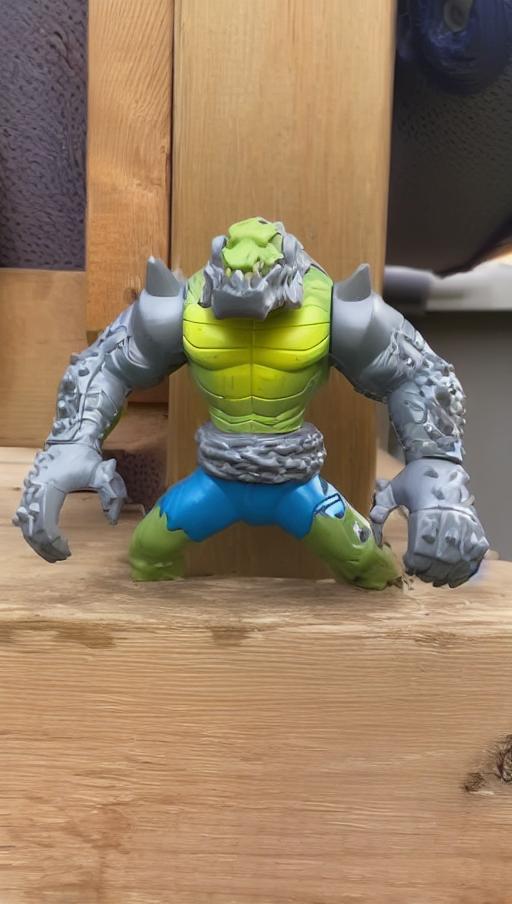}};
        \end{tikzpicture} &

        \begin{tikzpicture}[spy using outlines={}]
            \node {\includegraphics[width=0.15\columnwidth]{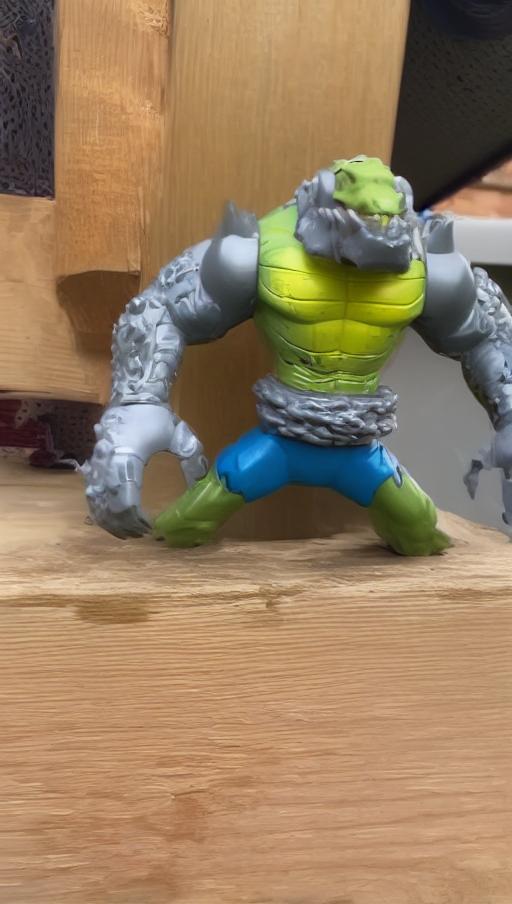}};
        \end{tikzpicture} &

        \begin{tikzpicture}[spy using outlines={circle,yellow,magnification=2,size=0.6cm, connect spies}]
            \node {\includegraphics[width=0.15\columnwidth]{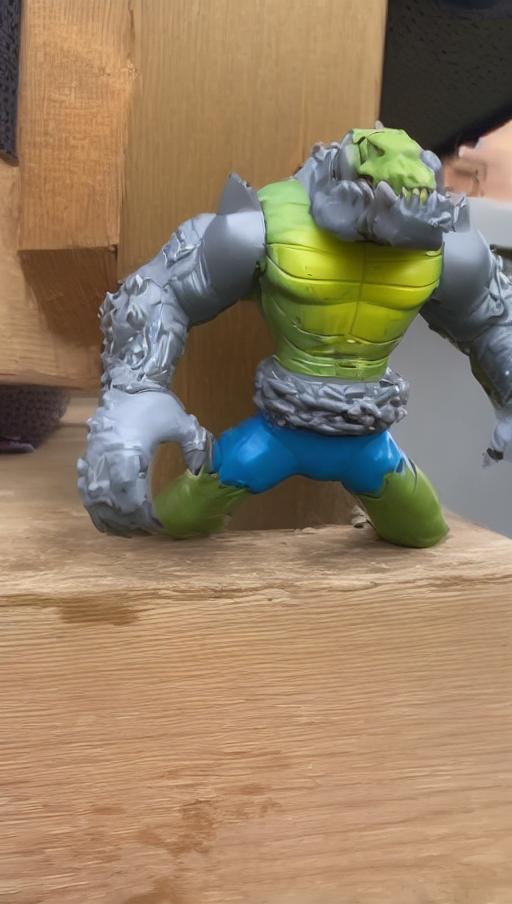}};
            \spy on (0.40,-0.25) in node [left] at (0.12,-0.78);
        \end{tikzpicture} &

        \begin{tikzpicture}[spy using outlines={}]
            \node {\includegraphics[width=0.15\columnwidth]{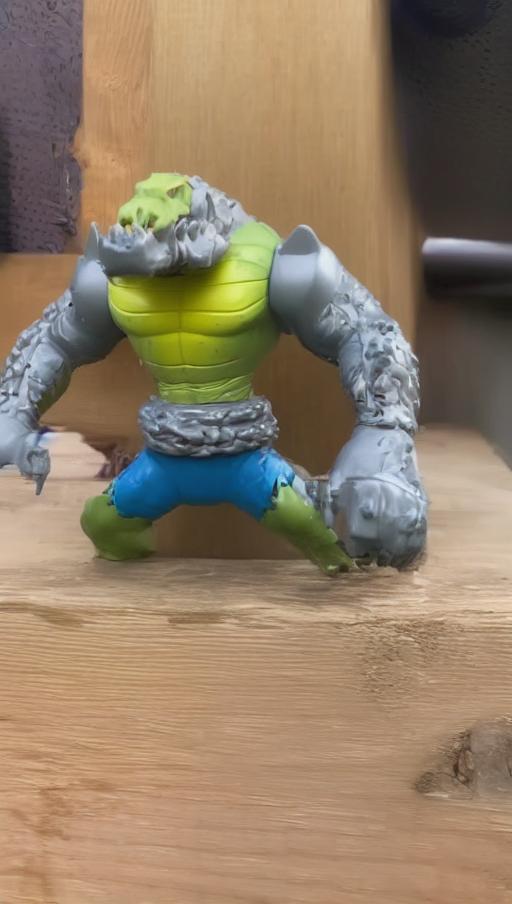}};
        \end{tikzpicture} &

        \begin{tikzpicture}[spy using outlines={}]
            \node {\includegraphics[width=0.15\columnwidth]{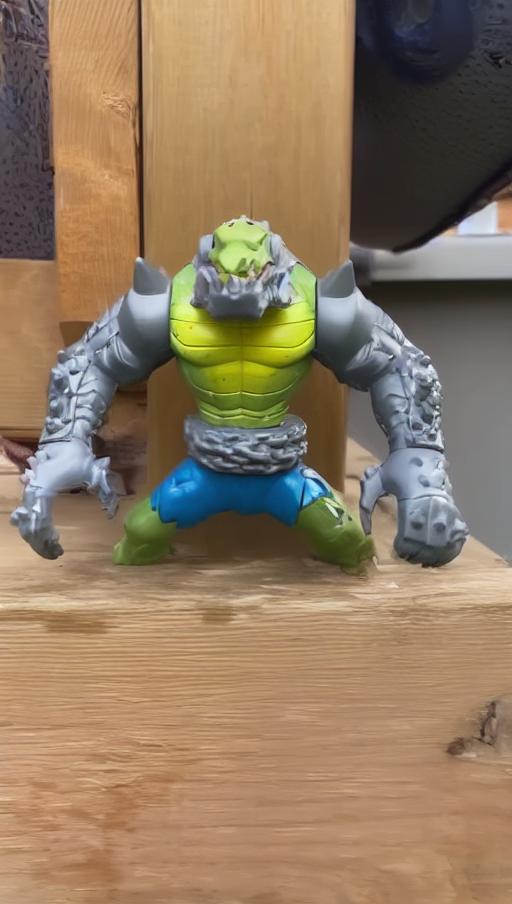}};
        \end{tikzpicture} \\[-8pt]

        \raisebox{29pt}{\rotatebox[origin=t]{90}{w/o progressive}} \hspace{1pt} &
        \begin{tikzpicture}[spy using outlines={}]
            \node {\includegraphics[width=0.15\columnwidth]{images/ablation/skel_01_frame_00011.jpg}};
        \end{tikzpicture} &
        \begin{tikzpicture}[spy using outlines={}]
            \node {\includegraphics[width=0.15\columnwidth]{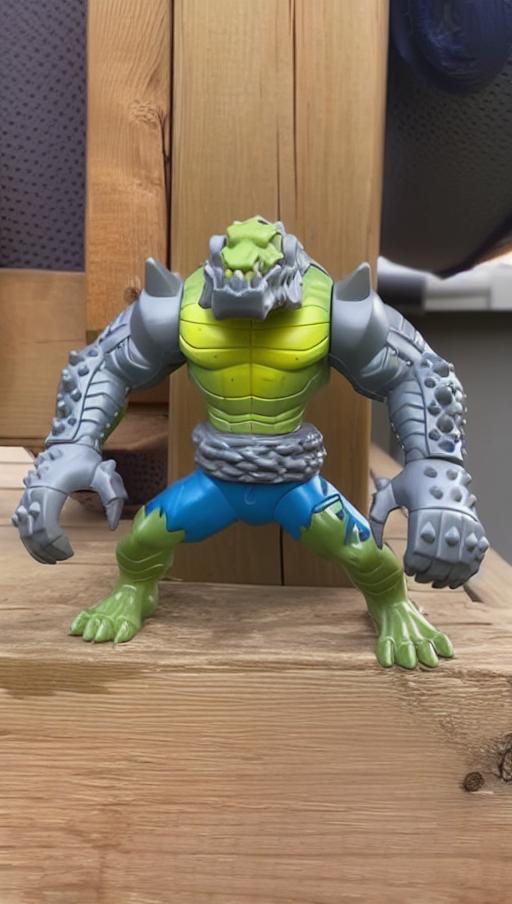}};
        \end{tikzpicture} &

        \begin{tikzpicture}[spy using outlines={}]
            \node {\includegraphics[width=0.15\columnwidth]{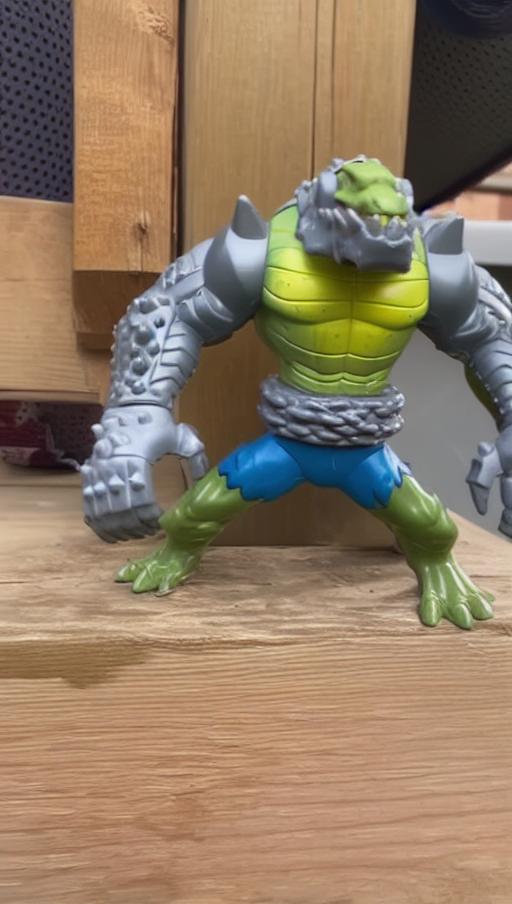}};
        \end{tikzpicture} &

        \begin{tikzpicture}[spy using outlines={circle,yellow,magnification=2,size=0.6cm, connect spies}]
            \node {\includegraphics[width=0.15\columnwidth]{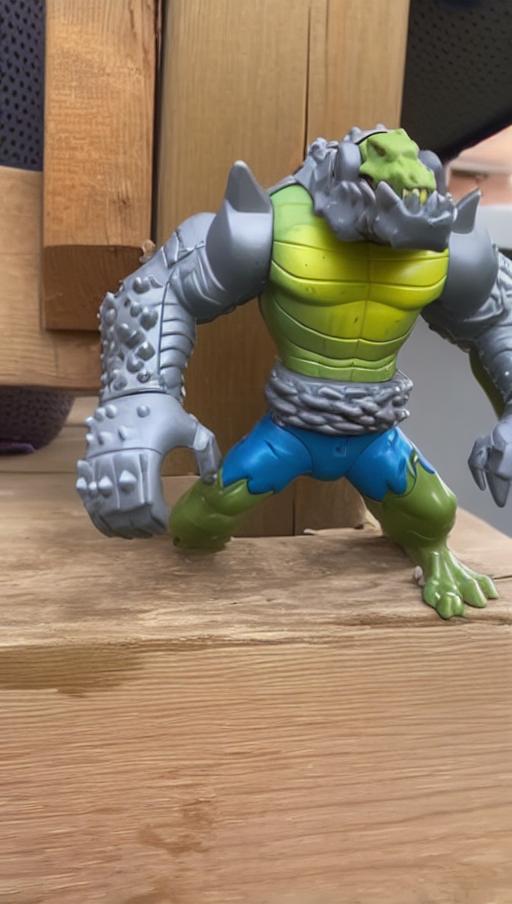}};
            \spy on (-0.10,-0.20) in node [left] at (0.12,-0.78);
        \end{tikzpicture} &

        \begin{tikzpicture}[spy using outlines={}]
            \node {\includegraphics[width=0.15\columnwidth]{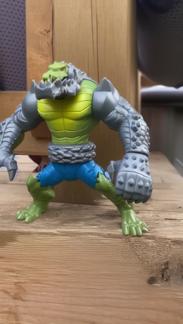}};
        \end{tikzpicture} &

        \begin{tikzpicture}[spy using outlines={}]
            \node {\includegraphics[width=0.15\columnwidth]{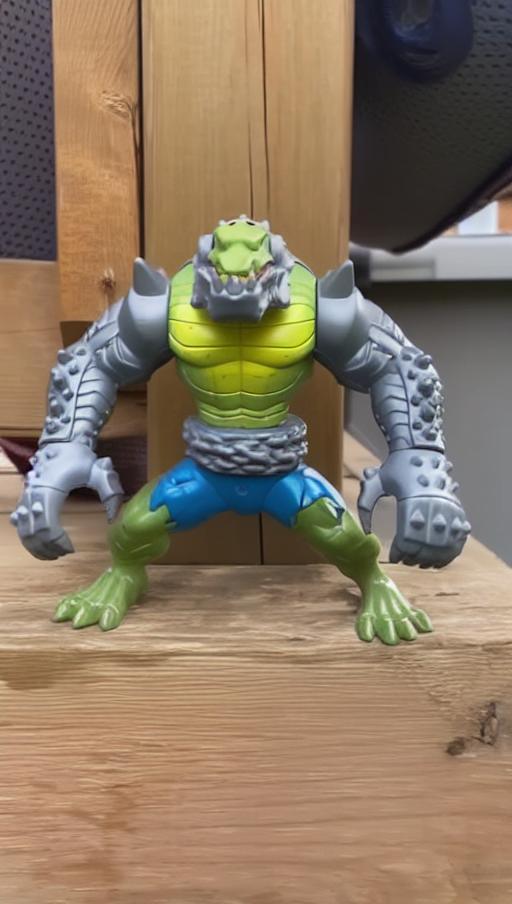}};
        \end{tikzpicture} \\[-8pt]
        
        \raisebox{30pt}{\rotatebox[origin=t]{90}{full method}} \hspace{1pt} &
        \begin{tikzpicture}[spy using outlines={}]
            \node {\includegraphics[width=0.15\columnwidth]{images/ablation/skel_01_frame_00011.jpg}};
        \end{tikzpicture} &
        \begin{tikzpicture}[spy using outlines={circle,yellow,magnification=2,size=0.6cm, connect spies}]
            \node {\includegraphics[width=0.15\columnwidth]{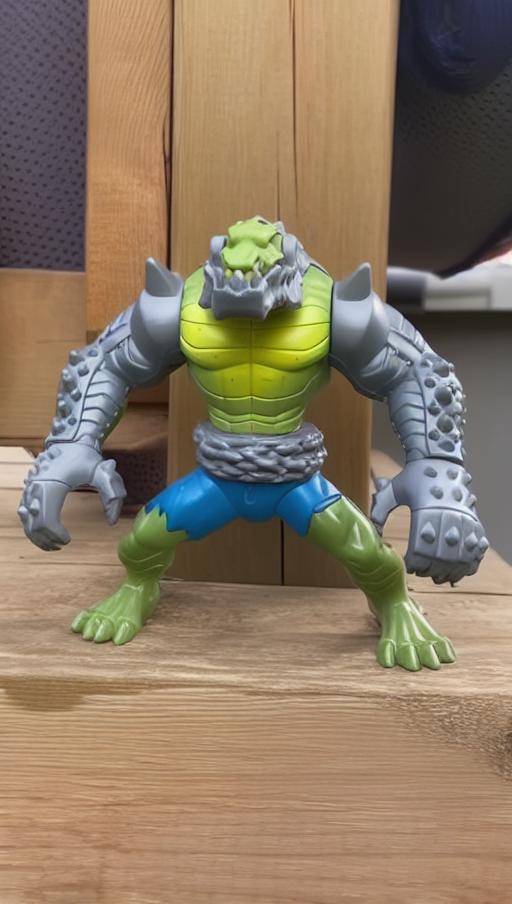}};
        \end{tikzpicture} &

        \begin{tikzpicture}[spy using outlines={}]
            \node {\includegraphics[width=0.15\columnwidth]{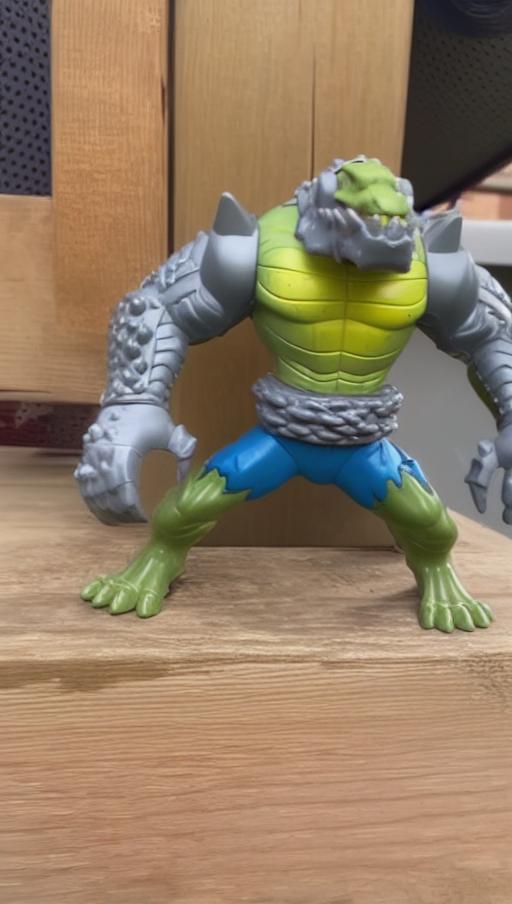}};
        \end{tikzpicture} &

        \begin{tikzpicture}[spy using outlines={circle,yellow,magnification=2,size=0.6cm, connect spies}]
            \node {\includegraphics[width=0.15\columnwidth]{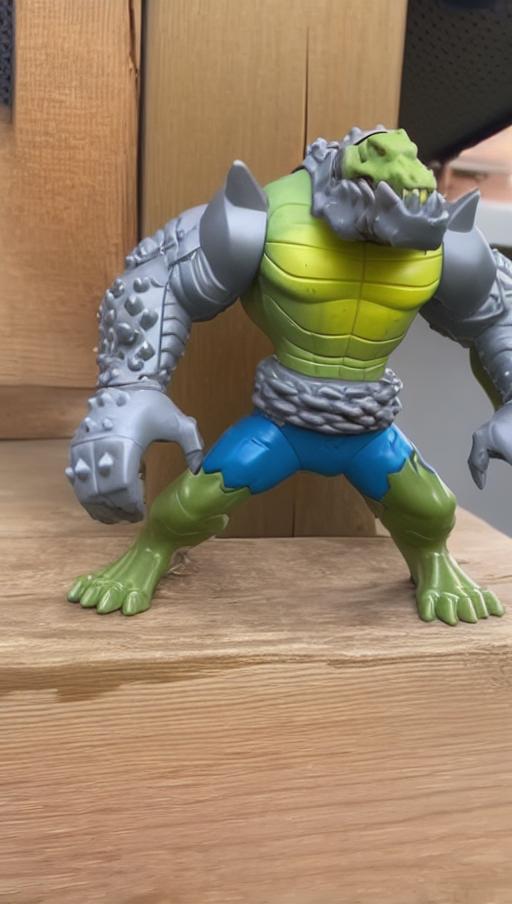}};
            \spy on (0.46,-0.45) in node [left] at (0.12,-0.79);
        \end{tikzpicture} &

        \begin{tikzpicture}[spy using outlines={}]
            \node {\includegraphics[width=0.15\columnwidth]{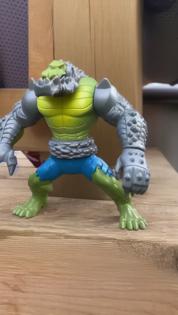}};
        \end{tikzpicture} &

        \begin{tikzpicture}[spy using outlines={}]
            \node {\includegraphics[width=0.15\columnwidth]{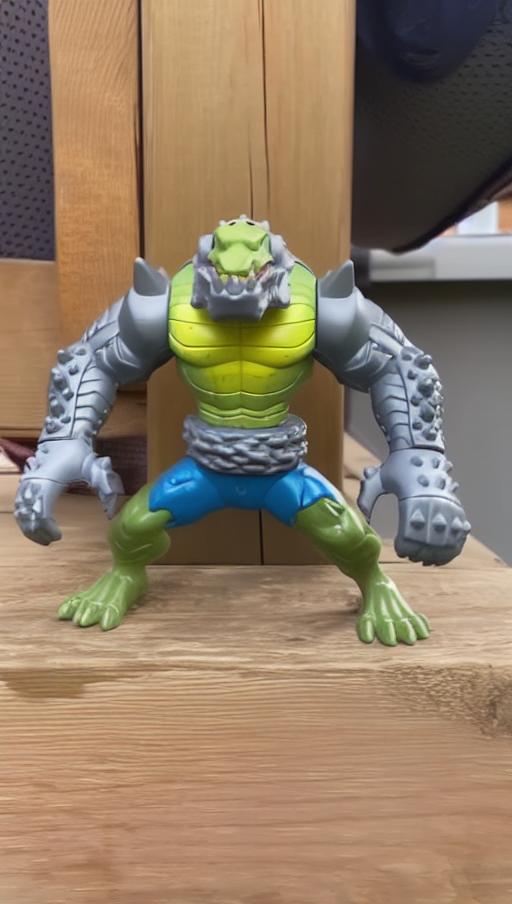}};
        \end{tikzpicture} \\[-6pt]

    \end{tabular}
    }
    \vspace{-6pt}
    \caption{Ablation study results. Our full method creates more consistent images, while accurately preserving the original scene. }
    \vspace{-16pt}
    \label{fig:ablation}
\end{figure}

We begin with a study of the importance of the main components of our method. 
We consider the following configurations: (i) Independently editing the images with MasaCtrl~\cite{cao2023masactrl}, (ii) directly injecting the rendered queries instead of our soft-injection mechanism, and (iii) using a non-progressive consolidation process. There, we first edit all the images independently using MasaCtrl and cache their self-attention queries along different timesteps. We then train QNeRFs on these queries, and finally re-create the edits with soft Q-injections.
Qualitative results are presented in Figure~\ref{fig:ablation}. As can be seen, independent image editing (second row) leads to inconsistent results. For example, the subject's feet are located in different positions with respect to the wooden deck (second column compared to third column). Additionally, the legs differ in their shapes (rightmost column) and the height of the deck varies between the images.
Directly injecting the rendered queries (third row) makes the images consistent, but they diverge too much from the original images, cutting the legs and increasing the height of the deck.
Training the QNeRFs with a non-progressive approach (fourth row) can lead to more artifacts, such as the missing leg in the fourth column.
In the last row, the legs are consistently positioned, and their shape does not vary between different images. Additionally, the shape of the deck remains as in the original images.

\begin{table*}
    \centering
    \begin{tabular}{lcccc}
        \toprule
        Metric & IN2N-CN & CSD-CN & TokenFlow & Ours \\
        \midrule
        KID ($\downarrow$) & 0.280 & 0.090 & 0.440 & \textbf{0.072} \\
        FID ($\downarrow$) & 201 & 87 & 295 & \textbf{73} \\
        \midrule
        User Study Rank ($\downarrow$) & 2.26 & 2.12 & 3.70 & \textbf{1.90} \\
        User Study Win-rate ($\uparrow$) & 14.16\% & 34.16\% & 0.83\% & \textbf{50.83\%} \\
        \bottomrule
    \end{tabular}
    \caption{Quantitative evaluation metrics. Our method outperforms the baselines both in terms of fidelity, and user-preference.}
    \label{tab:quant}
    \vspace{-12pt}
\end{table*}

\subsection{Comparisons}
Next, we compare our method to prior art that tackles the multi-image editing problem. Specifically, we compare to three methods that take different approaches to the multi-image consistency problem:
First, we compare with InstructNeRF2NeRF (IN2N)~\cite{instructnerf2023}, a dataset update technique that first trains a NeRF on the data, then modifies it by iteratively rendering new images to replace existing dataset pictures, editing them, and training on them to better align the NeRF with the edits.
Second, we compare with CSD~\cite{kim2023collaborative} which performs a collaborative score distillation sampling (SDS) process that better aligns the SDS gradients across a large subset of images. We follow the authors and integrate CSD with IN2N, where the image-editing step itself is replaced with a CSD-based multi-view editing step. Finally, we compare with TokenFlow~\cite{tokenflow2023}, a recent text-based video editing method that improves cross-frame consistency through a flow-based approach. Here, we concatenate the initial image set into a single video which we then edit.

The above methods originally rely on text as an interface for editing. We integrate them with ControlNet~\cite{zhang2023adding}, allowing us to specify the target edit with spatial control. Specifically, in IN2N and CSD, we replace the InstructPix2Pix editor with MasaCtrl~\cite{cao2023masactrl}. We denote these versions of the methods by IN2N-CN and CSD-CN, respectively. For TokenFlow, we make use of their ControlNet version. Notably, TokenFlow builds its flow according to patch-similarity in the original video, which would not match the changed geometry in our case. We find that performance significantly improves when the flow is grounded in the inconsistent, frame-by-frame MasaCtrl images (rather than unedited images), and so compare with this variation of the model.

Finally, the outputs of CSD and IN2N are NeRFs. Hence, for a fair comparison with TokenFlow and our method, we simply train a NeRF on the outputs of TokenFlow and our method. Training a NeRF on their outputs further allows us to compare the geometry between the different methods, by observing the depth of the NeRF.

\begin{figure*}
\centering
\small{
\setlength{\tabcolsep}{0pt}
\begin{tabular}{c c c c c c c c c c c c c c c c c}

    \includegraphics[width=0.084\textwidth]{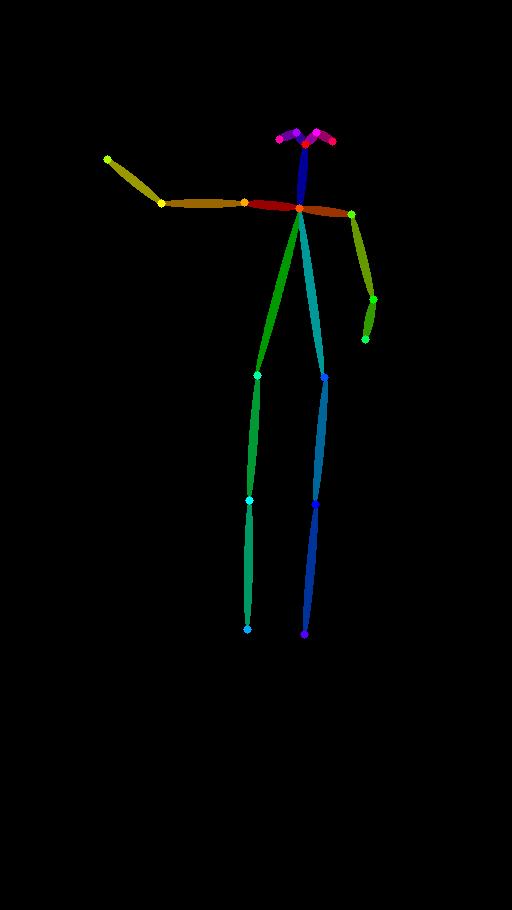} &
    { } &
    \includegraphics[width=0.084\textwidth]{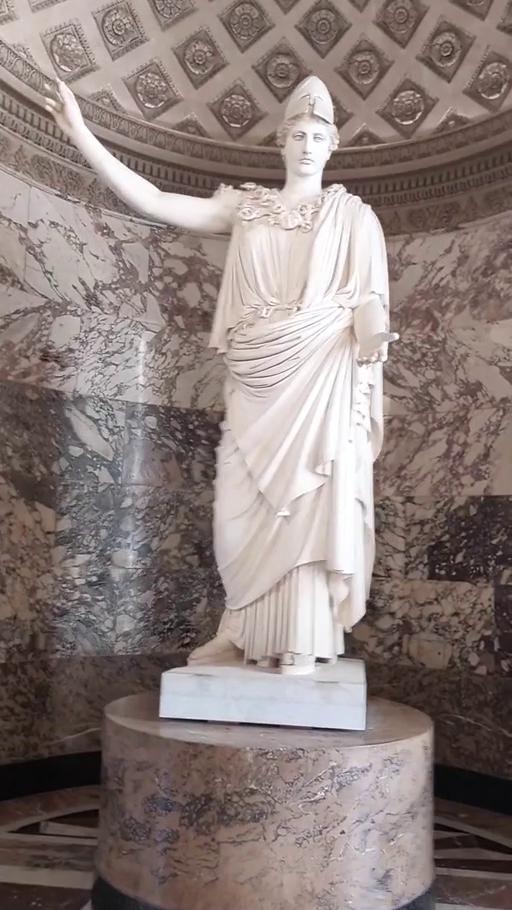} &
    { } &
    \includegraphics[width=0.084\textwidth]{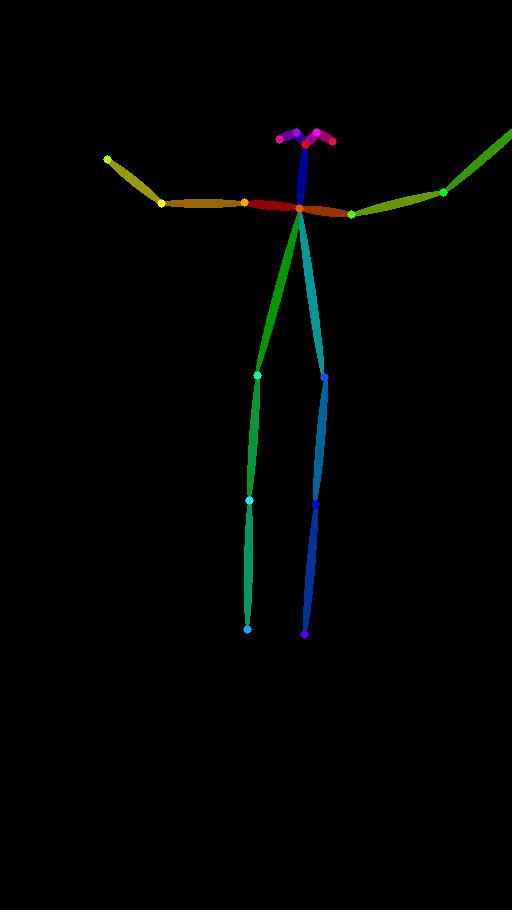} &
    { } &
    \includegraphics[width=0.084\textwidth]{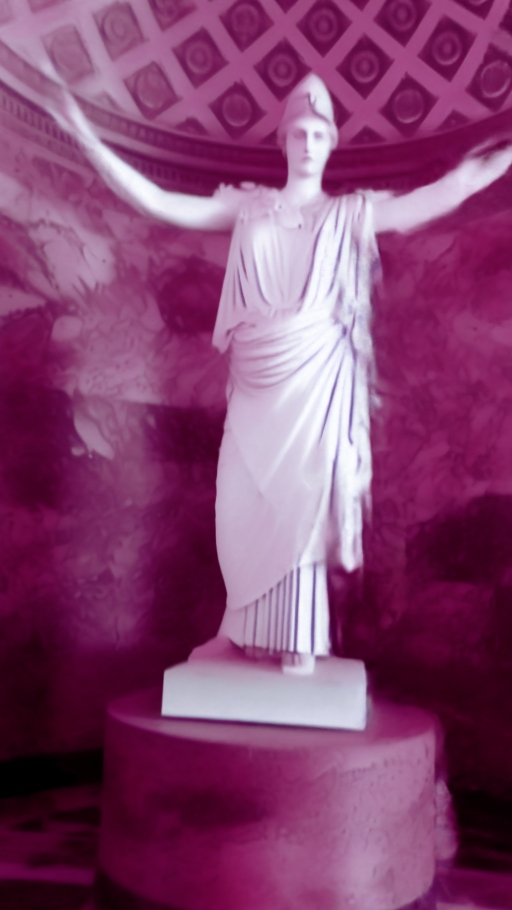} &
    \includegraphics[width=0.084\textwidth]{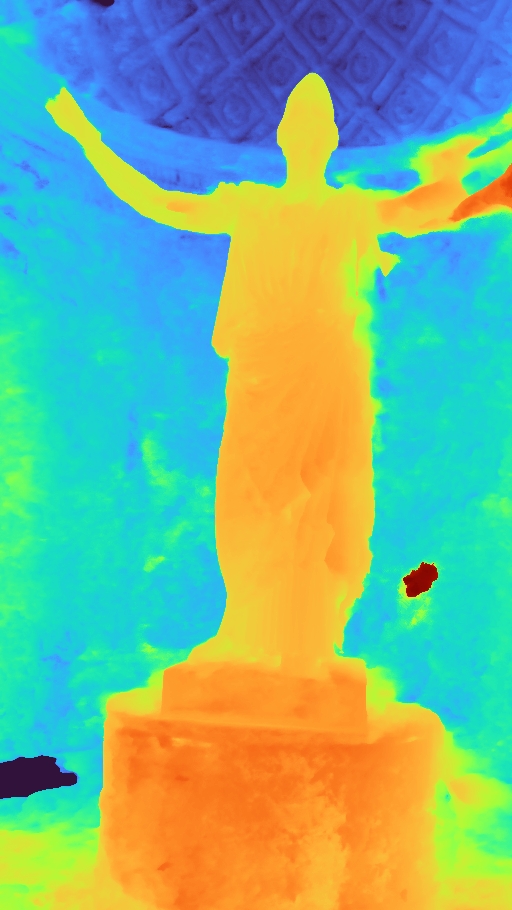} &
    { } &
    \includegraphics[width=0.084\textwidth]{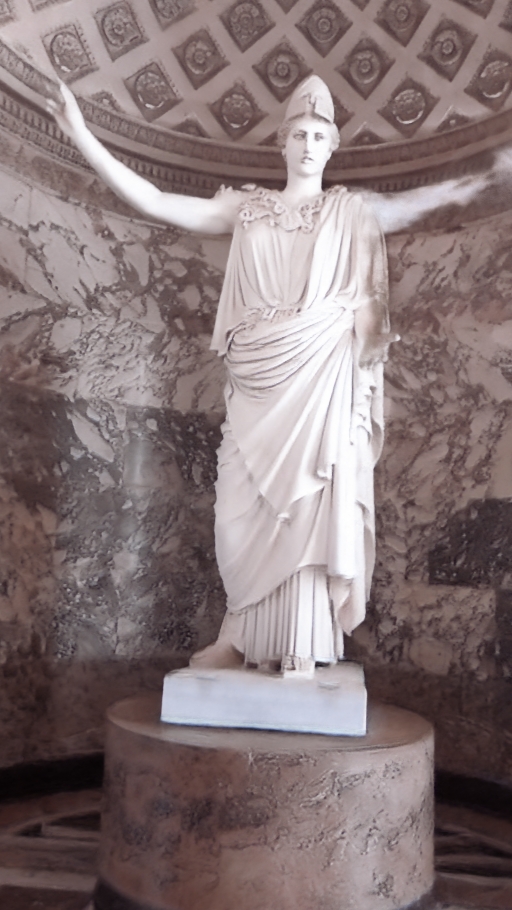} &
    \includegraphics[width=0.084\textwidth]{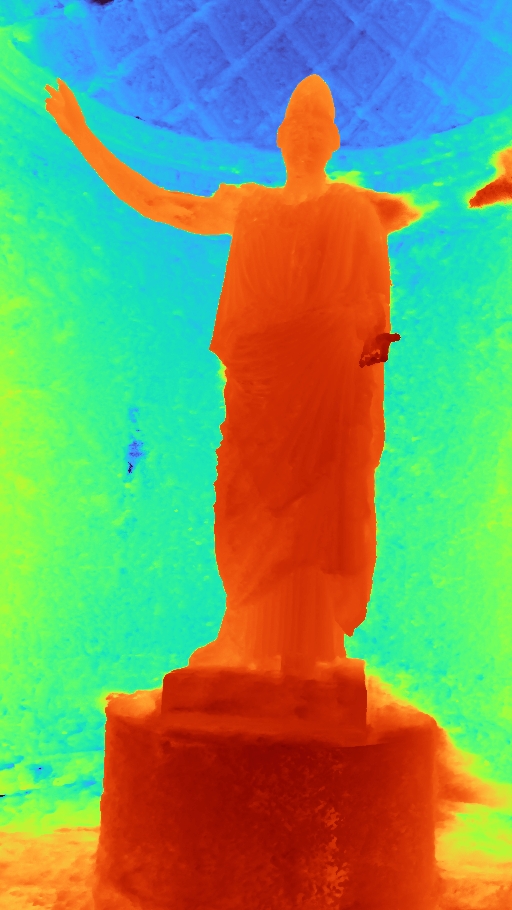} &
    { } &
    \includegraphics[width=0.084\textwidth]{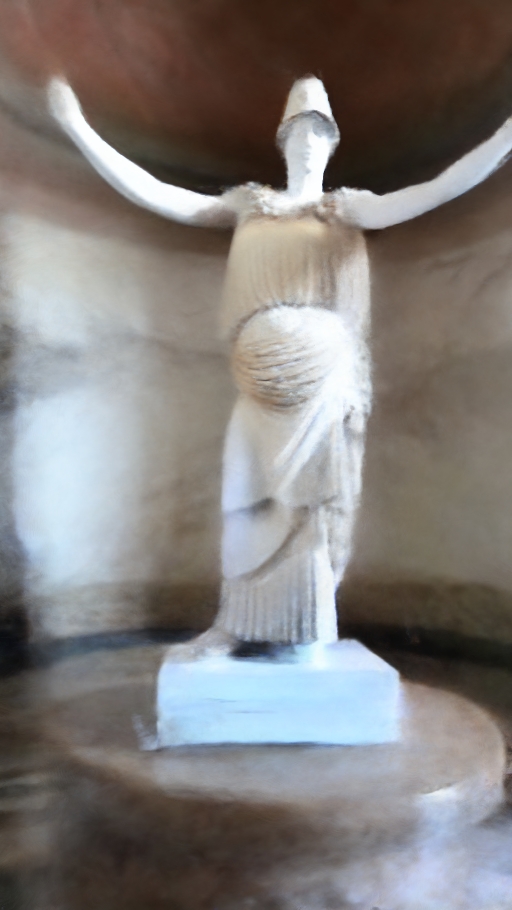} &
    \includegraphics[width=0.084\textwidth]{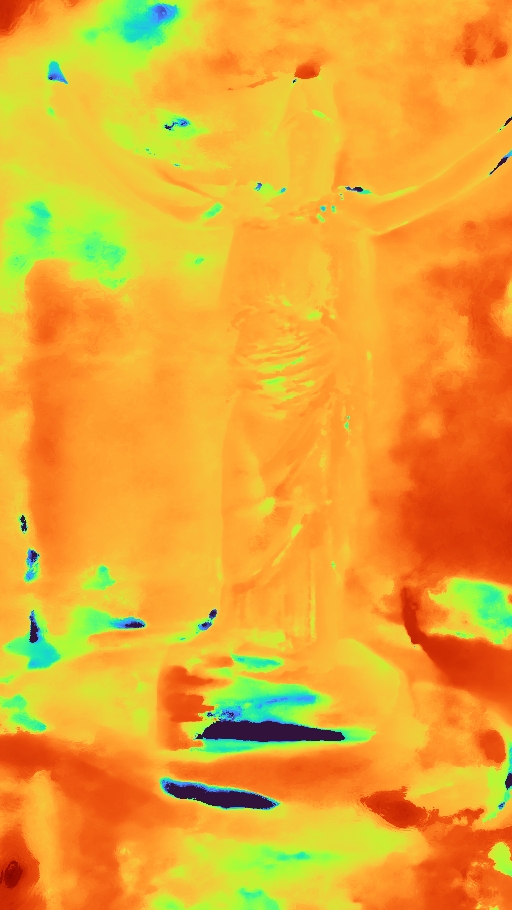} &
    { } &
    \includegraphics[width=0.084\textwidth]{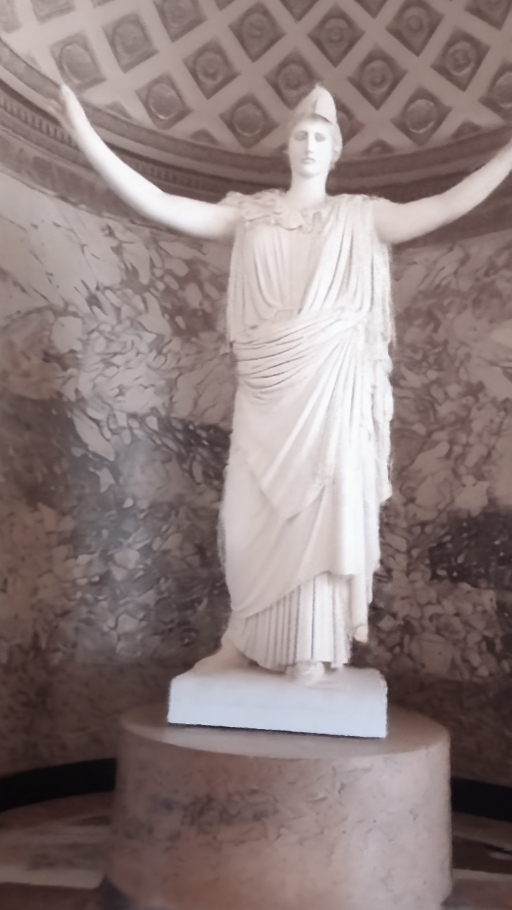} &
    \includegraphics[width=0.084\textwidth]{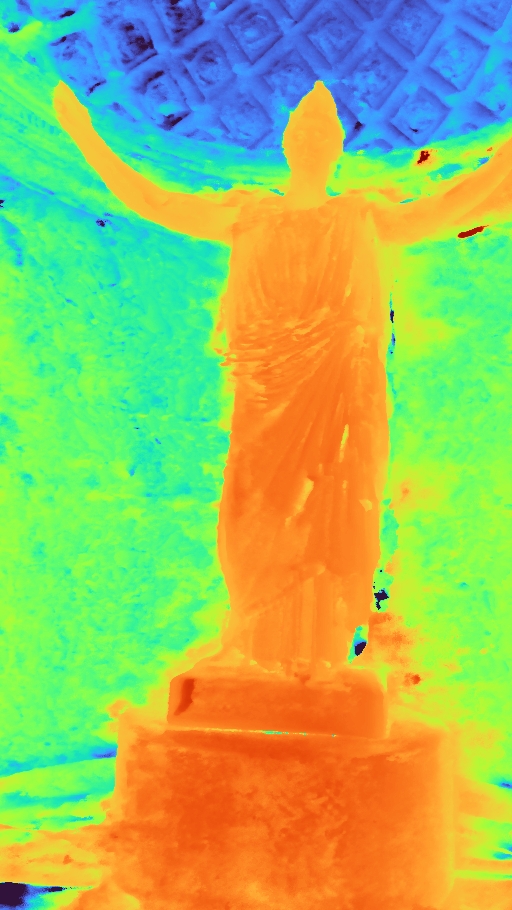} 
    \\
    \includegraphics[width=0.084\textwidth]{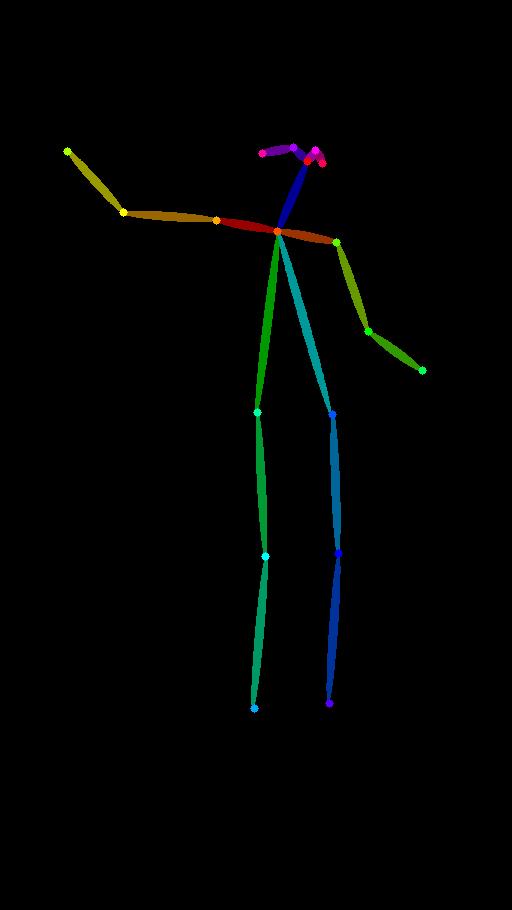} &
    { } &
    \includegraphics[width=0.084\textwidth]{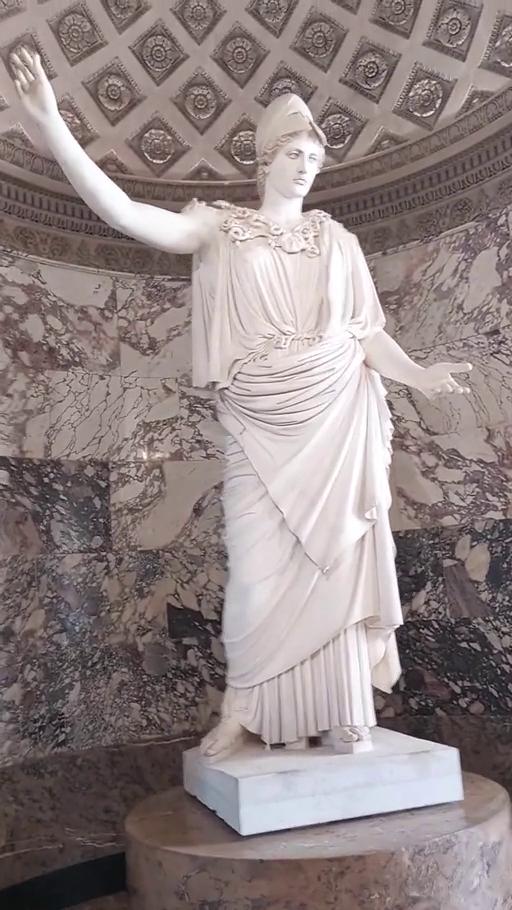} &
    { } &
    \includegraphics[width=0.084\textwidth]{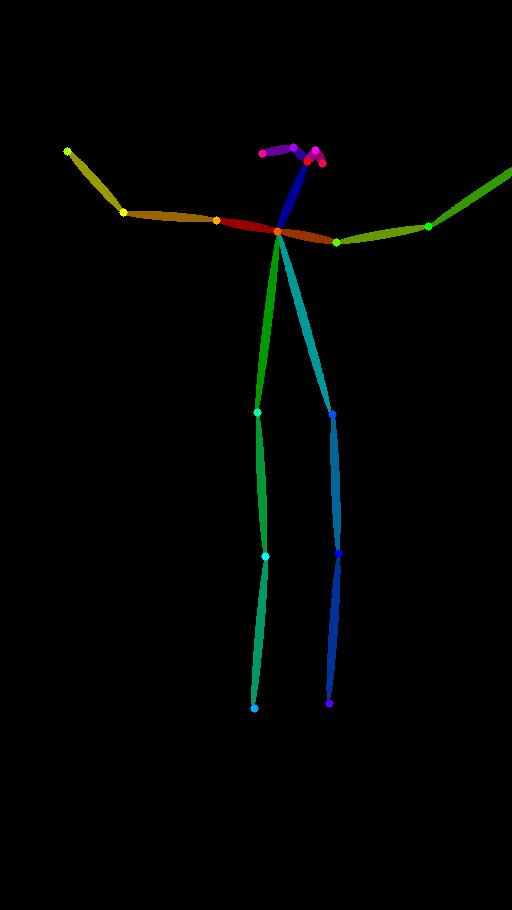} &
    { } &
    \includegraphics[width=0.084\textwidth]{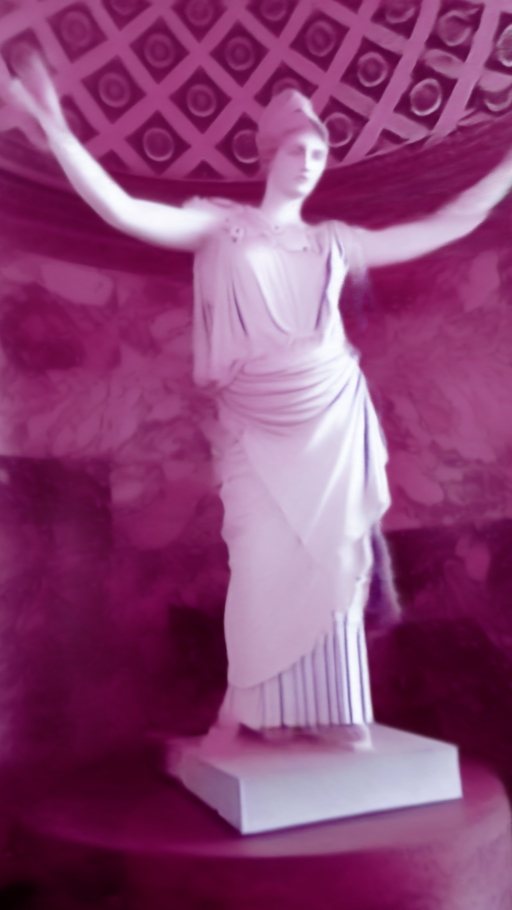} &
    \includegraphics[width=0.084\textwidth]{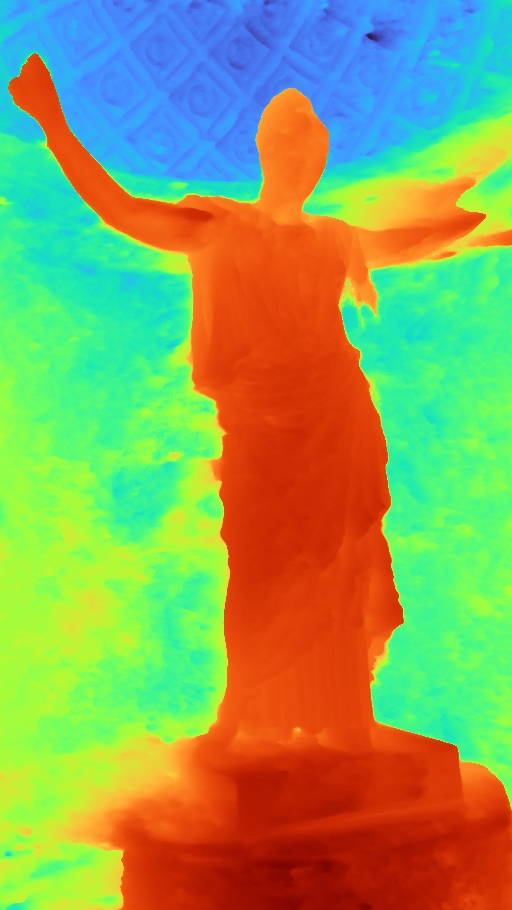} &
    { } &
    \includegraphics[width=0.084\textwidth]{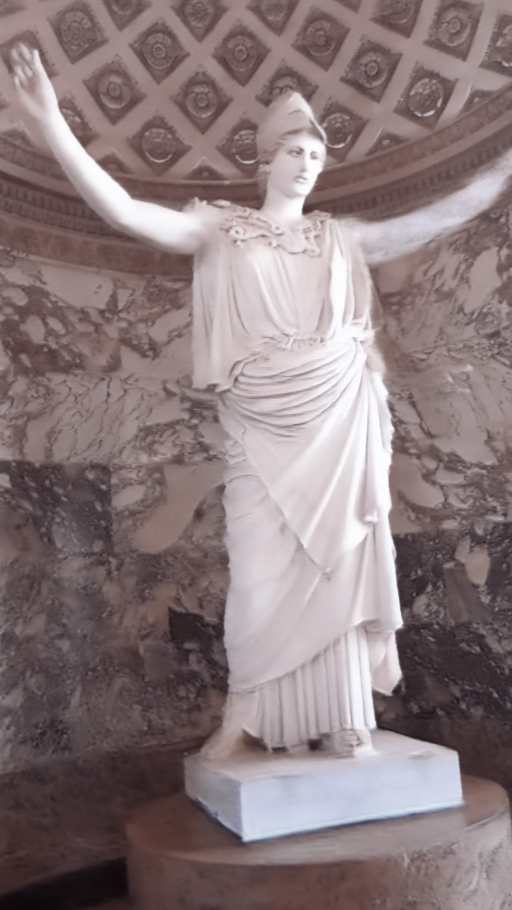} &
    \includegraphics[width=0.084\textwidth]{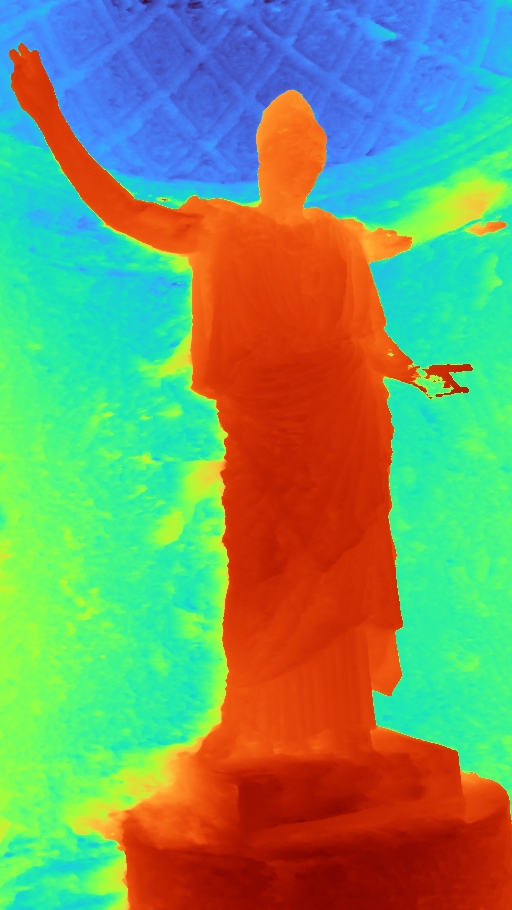} &
    { } &
    \includegraphics[width=0.084\textwidth]{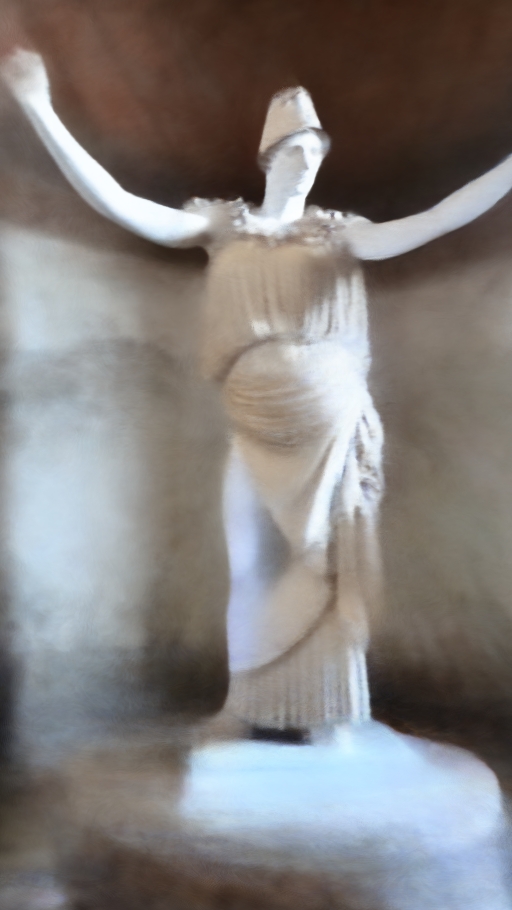} &
    \includegraphics[width=0.084\textwidth]{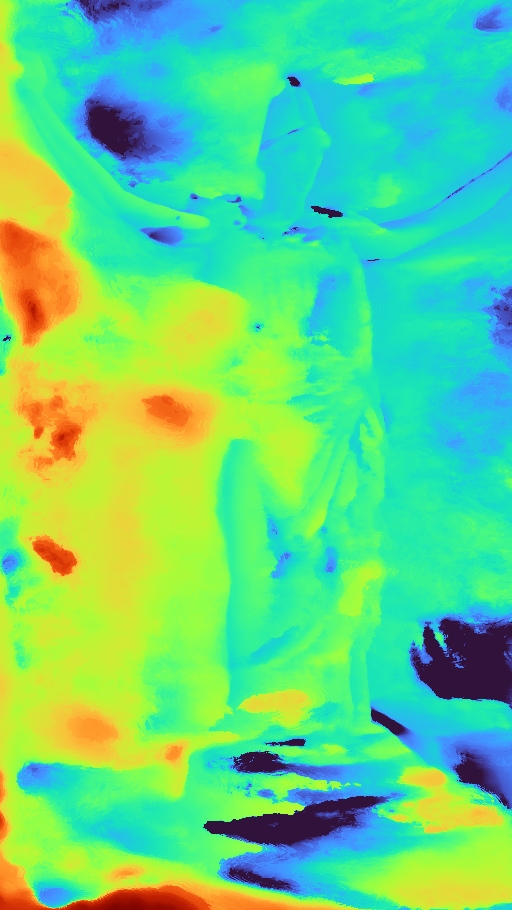} &
    { } &
    \includegraphics[width=0.084\textwidth]{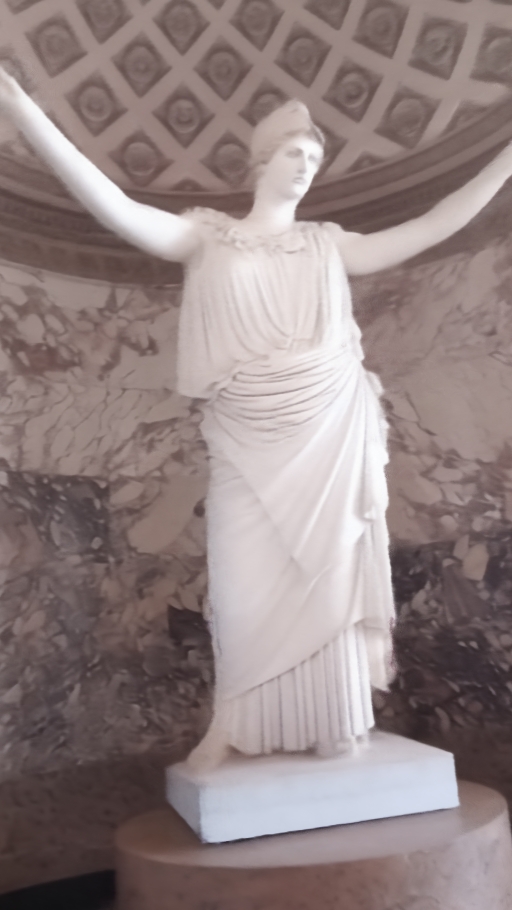} &
    \includegraphics[width=0.084\textwidth]{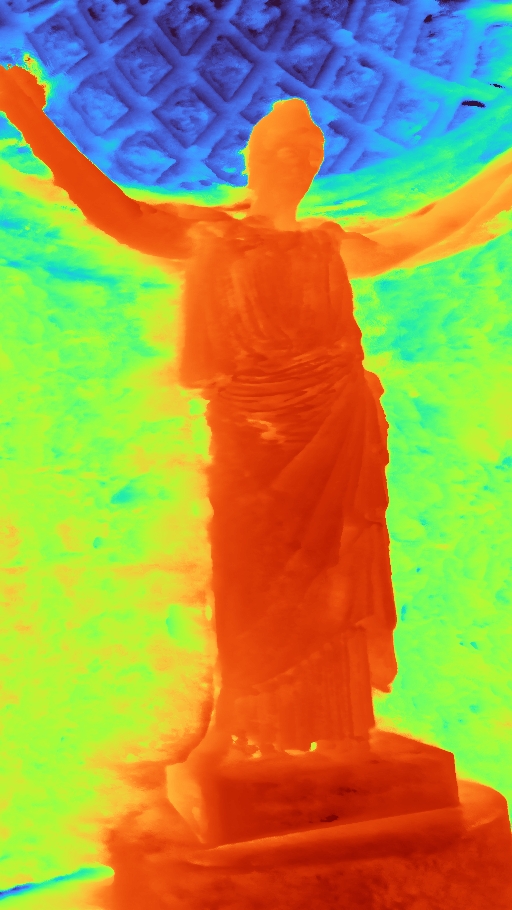} 
    \\
    \includegraphics[width=0.084\textwidth]{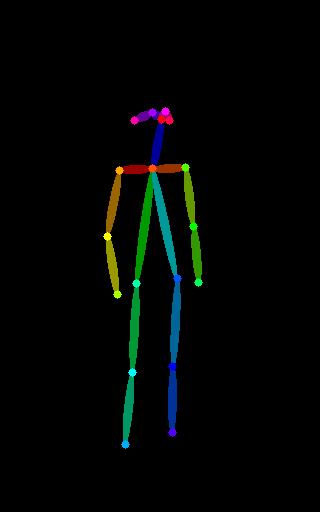} &
    { } &
    \includegraphics[width=0.084\textwidth]{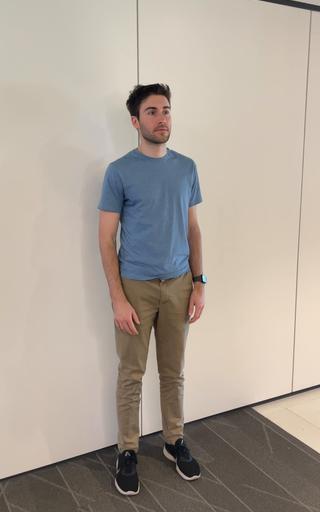} &
    { } &
    \includegraphics[width=0.084\textwidth]{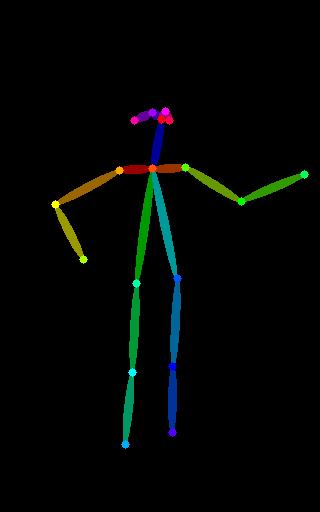} &
    { } &
    \includegraphics[width=0.084\textwidth]{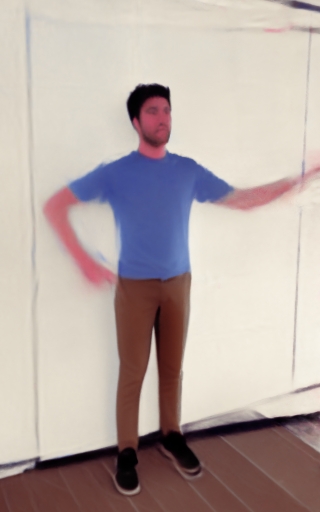} &
    \includegraphics[width=0.084\textwidth]{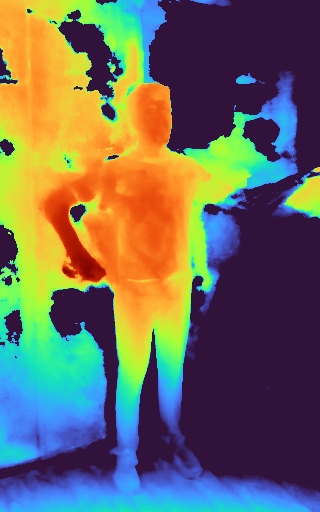} &
    { } &
    \includegraphics[width=0.084\textwidth]{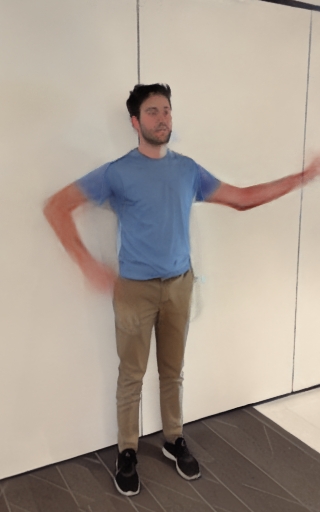} &
    \includegraphics[width=0.084\textwidth]{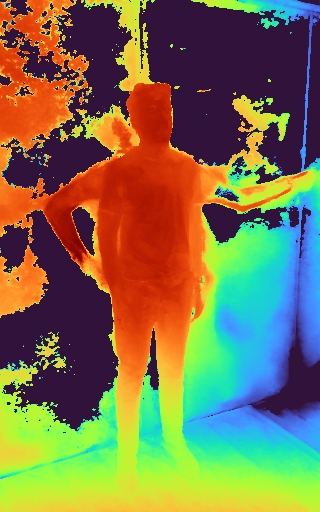} &
    { } &
    \includegraphics[width=0.084\textwidth]{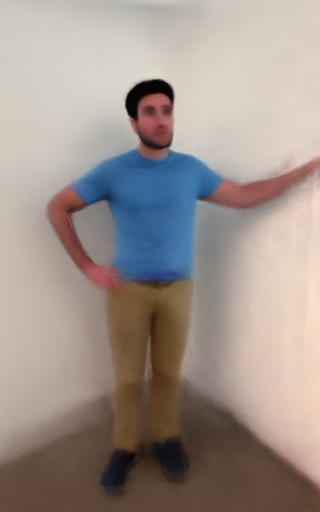} &
    \includegraphics[width=0.084\textwidth]{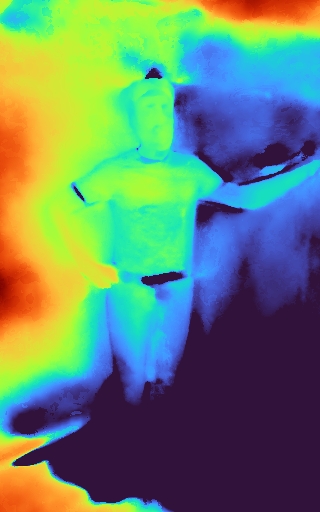} &
    { } &
    \includegraphics[width=0.084\textwidth]{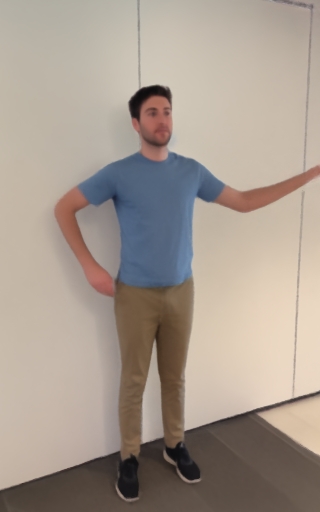} &
    \includegraphics[width=0.084\textwidth]{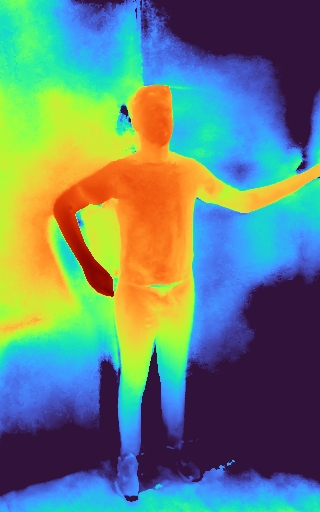} 
    \\
    \includegraphics[width=0.084\textwidth]{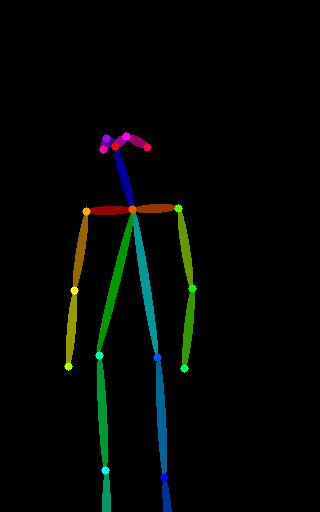} &
    { } &
    \includegraphics[width=0.084\textwidth]{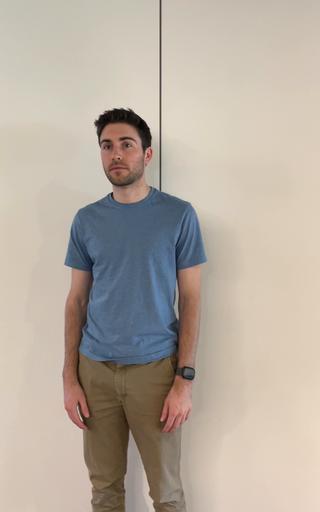} &
    { } &
    \includegraphics[width=0.084\textwidth]{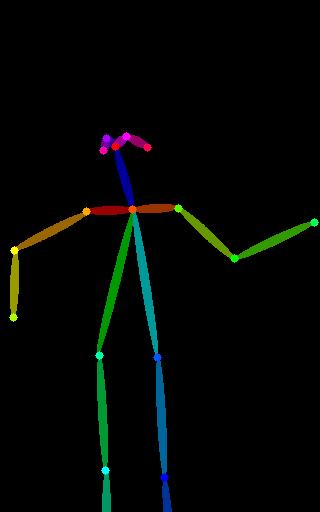 } &
    { } &
    \includegraphics[width=0.084\textwidth]{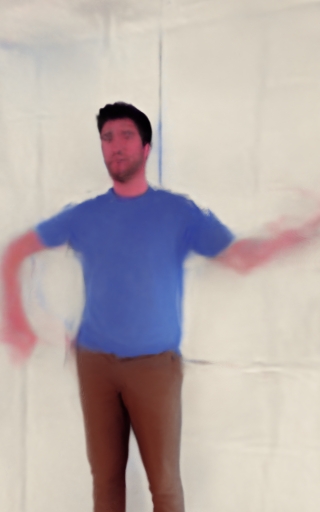} &
    \includegraphics[width=0.084\textwidth]{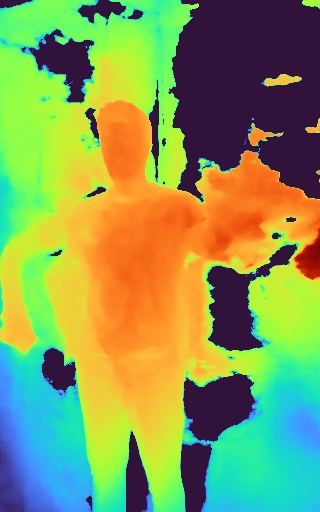} &
    { } &
    \includegraphics[width=0.084\textwidth]{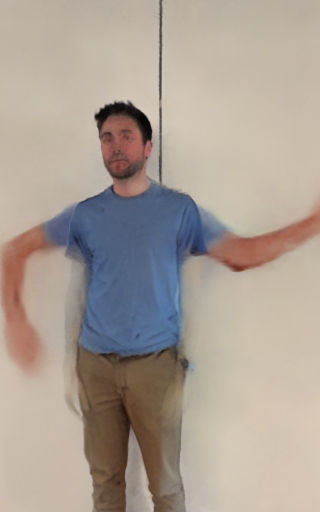} &
    \includegraphics[width=0.084\textwidth]{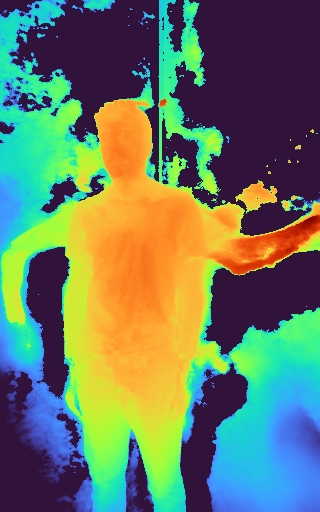} &
    { } &
    \includegraphics[width=0.084\textwidth]{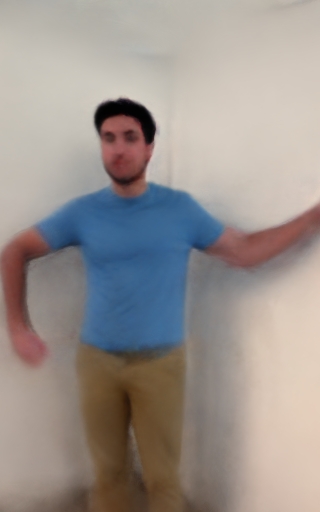} &
    \includegraphics[width=0.084\textwidth]{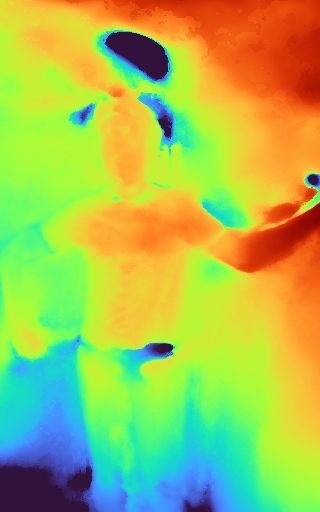} &
    { } &
    \includegraphics[width=0.084\textwidth]{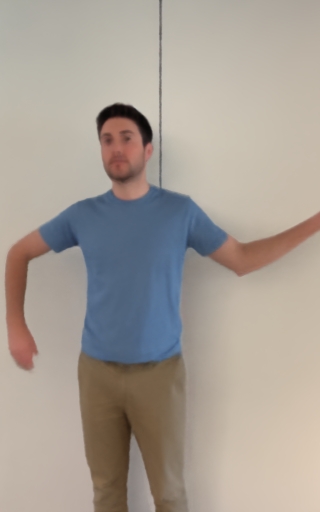} &
    \includegraphics[width=0.084\textwidth]{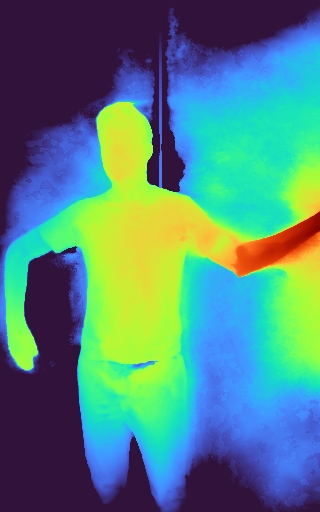} 
    \\
    Input && Input && Target && \multicolumn{2}{c}{IN2N-CN} && \multicolumn{2}{c}{CSD-CN} && \multicolumn{2}{c}{TokenFlow} && \multicolumn{2}{c}{Ours}
    \\
    Control && Image && Control 
     
\end{tabular}
}
\vspace{-8pt}
\caption{
Qualitative comparison of our approach with baseline methods. 
Techniques relying on ``dataset update'', such as IN2N-CN and CSD-CN, struggle to alter the geometry.
This can be seen in the noisy depth of the statue's right arm when using IN2N-CN, and the ghostly right arm of the statue with CSD-CN. TokenFlow struggles to preserve the appearance of the original image, and tends to produce noisy geometry, suggesting a lack of consistency between the edited frames. Our method preserves the appearance of the original images while changing the geometry consistently.
}
\vspace{-14pt}
\label{fig:comparison}
\end{figure*}

Rendered images of all methods are presented in Figure~\ref{fig:comparison}. In both IN2N and CSD, the model relies on partial dataset updates -- \ie, there exists an implicit assumption that the target edit can be performed gradually, and that averaging over the partially-edited views will lead to a consistent NeRF. While this assumption holds for appearance edits, it does not hold for shape changes such as articulations. For example, moving the arms of a person in part of the dataset, eventually leads to a NeRF where partial arms are located in both the source and target locations. Now, the subject has $4$ arms, and the baseline editing method struggles to recover. Indeed, such ghostly-limb artifacts can be seen in both methods.
For example, in the last row in Figure~\ref{fig:comparison}, the original arms can be seen in both the RGB and depth images of CSD. For IN2N, the original arms can be seen in the depth image, and the new right arm has noisy geometry.
Coupling CSD with MasaCtrl also leads to significant increases in computation times, as it requires running thousands of DDIM~\cite{song2021denoising} inversions for every dataset update step. In one extreme case (the alligator-toy scene, containing $491$ images), the method failed to update enough images to change the NeRF, even after a full week of training on an H100 GPU. We omit this instance from the automated evaluations.
TokenFlow edits all images without a-priori averaging through a NeRF. Hence, it better aligns with the desired shape and avoids the ghostly-limb artifacts. However, geometric edits still violate its flow-consistency assumptions, leading to visual artifacts. Moreover, its feature injection approach struggles to match MasaCtrl's in faithfulness to the original frames.

Our approach successfully overcomes the artifacts that arise with gradual dataset updates over an existing NeRF, while still retaining high similarity to the original frame.

\begin{figure*}
\centering
\setlength{\tabcolsep}{1pt}
\begin{tabular}{c c c c}

    \includegraphics[width=0.30\textwidth]{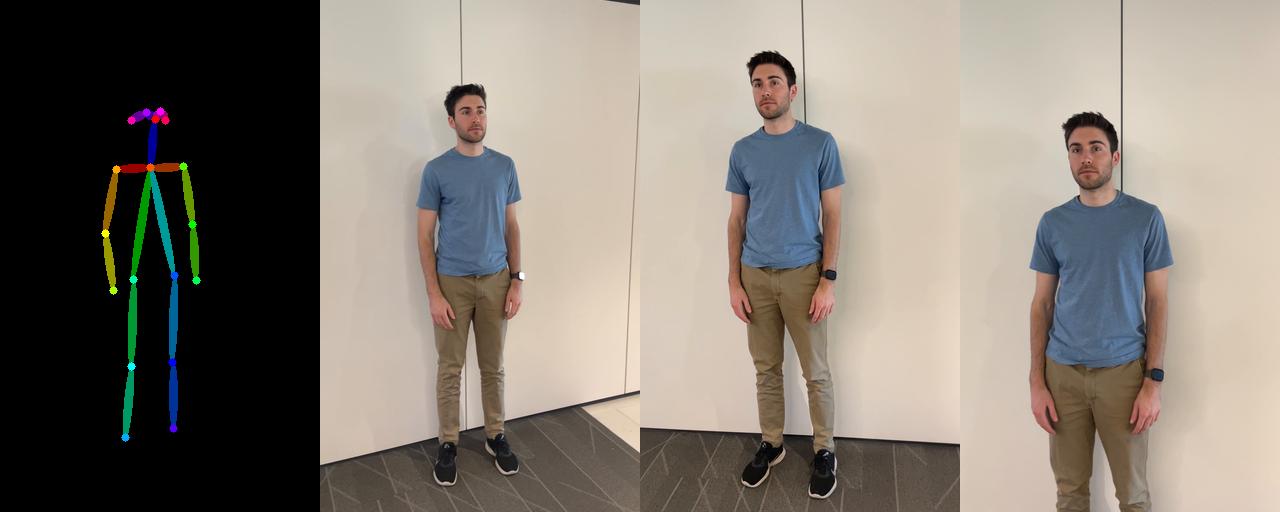} &
    \includegraphics[width=0.30\textwidth]{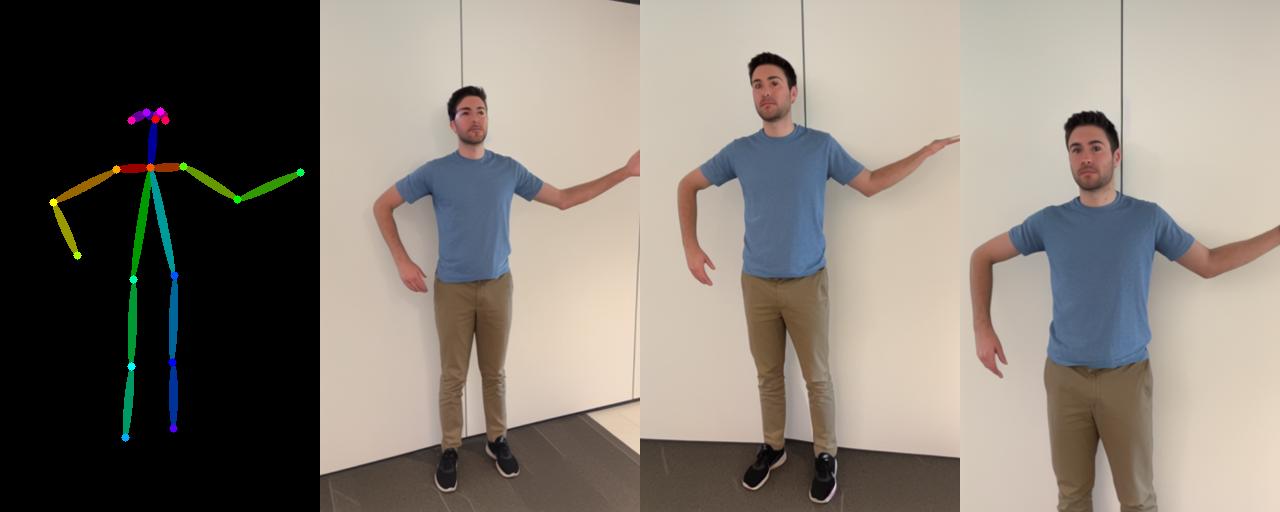} &
    \includegraphics[width=0.30\textwidth]{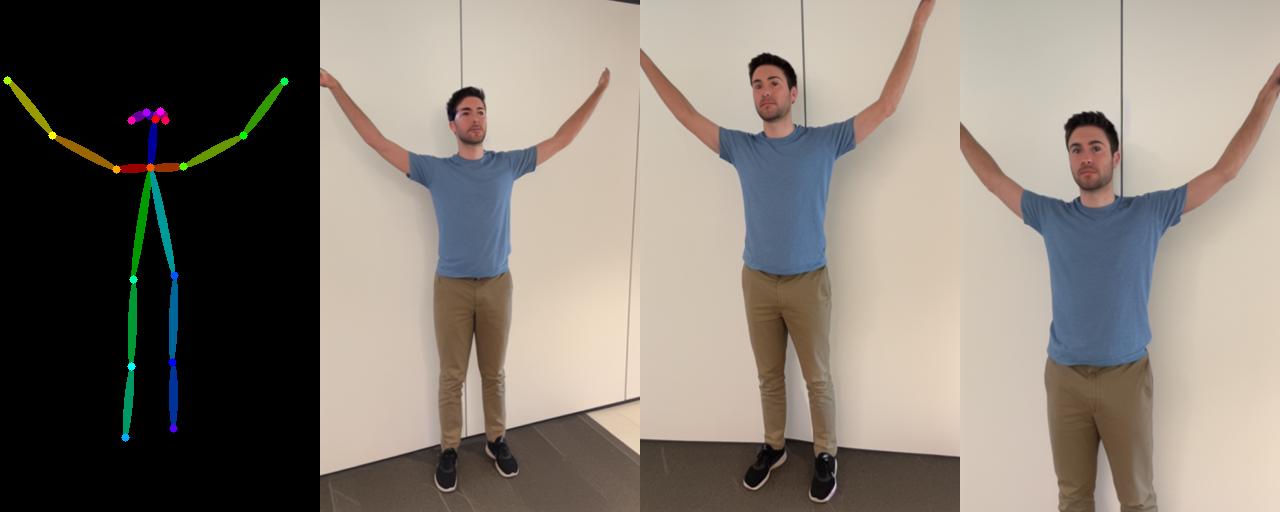} 
    \\
    Original \\
    \includegraphics[width=0.30\textwidth]{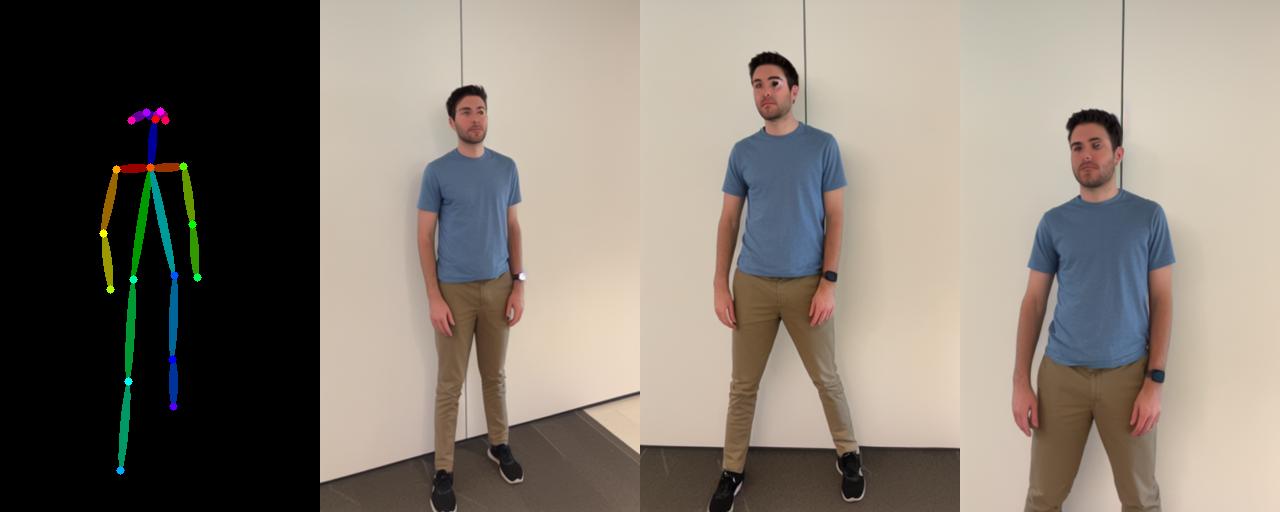} &
    \includegraphics[width=0.30\textwidth]{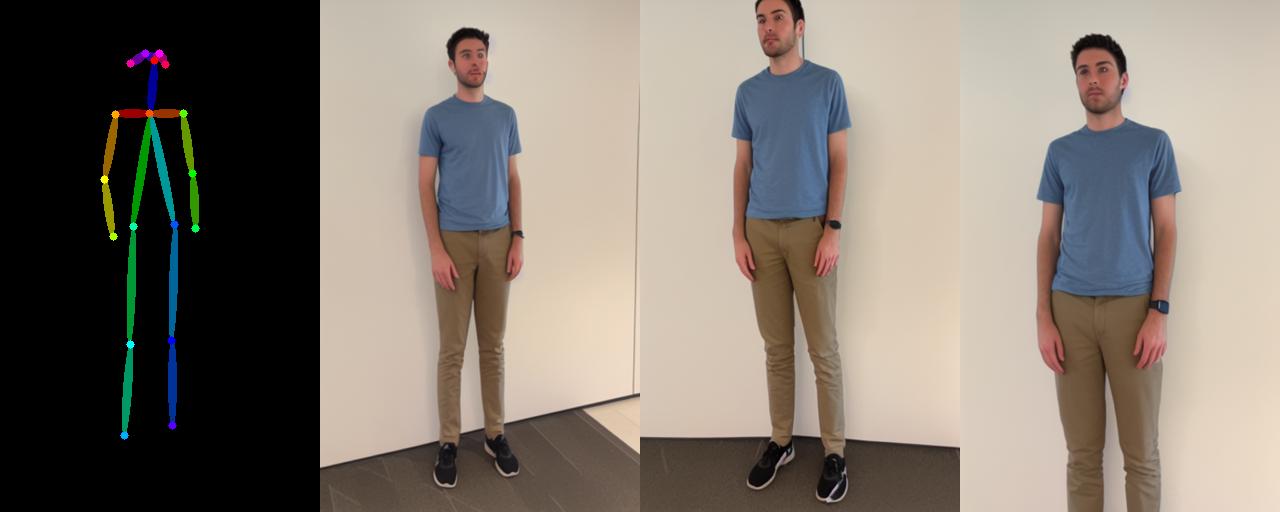} &
    \includegraphics[width=0.30\textwidth]{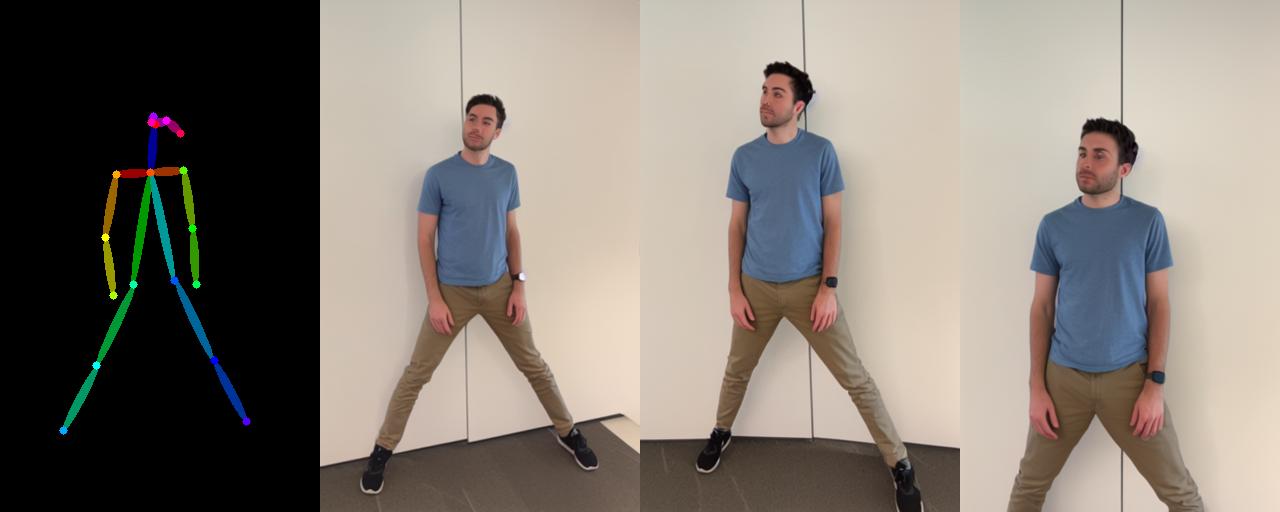} 
    \\

    \includegraphics[width=0.30\textwidth]{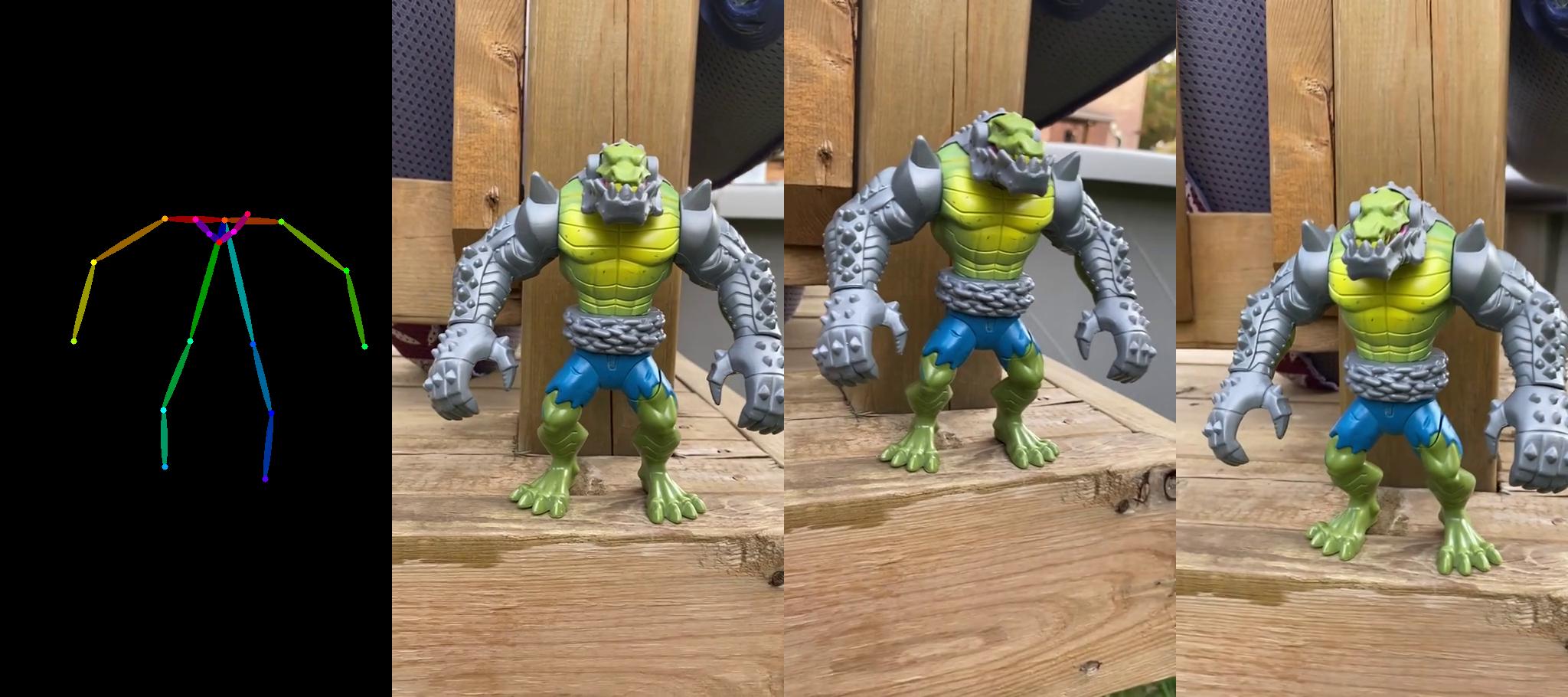} &
    \includegraphics[width=0.30\textwidth]{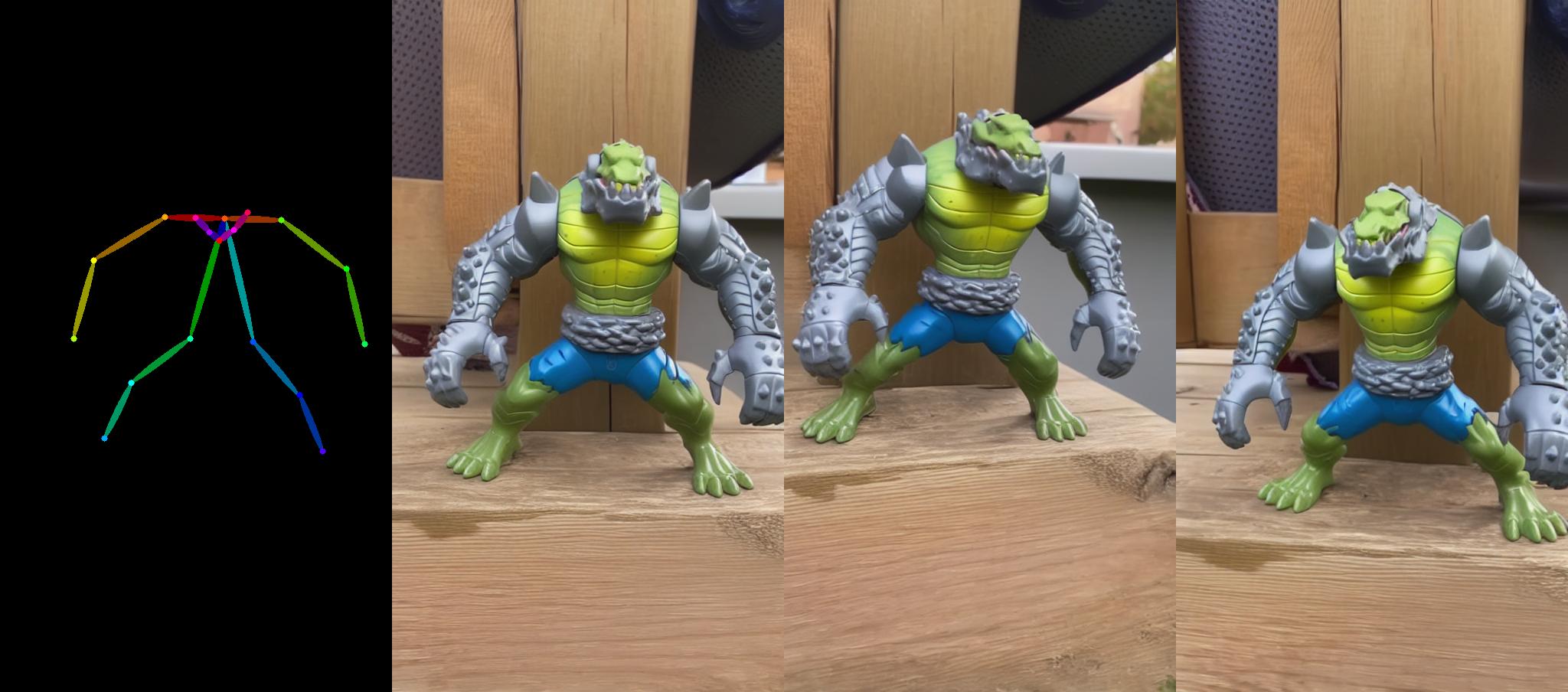} &
    \includegraphics[width=0.30\textwidth]{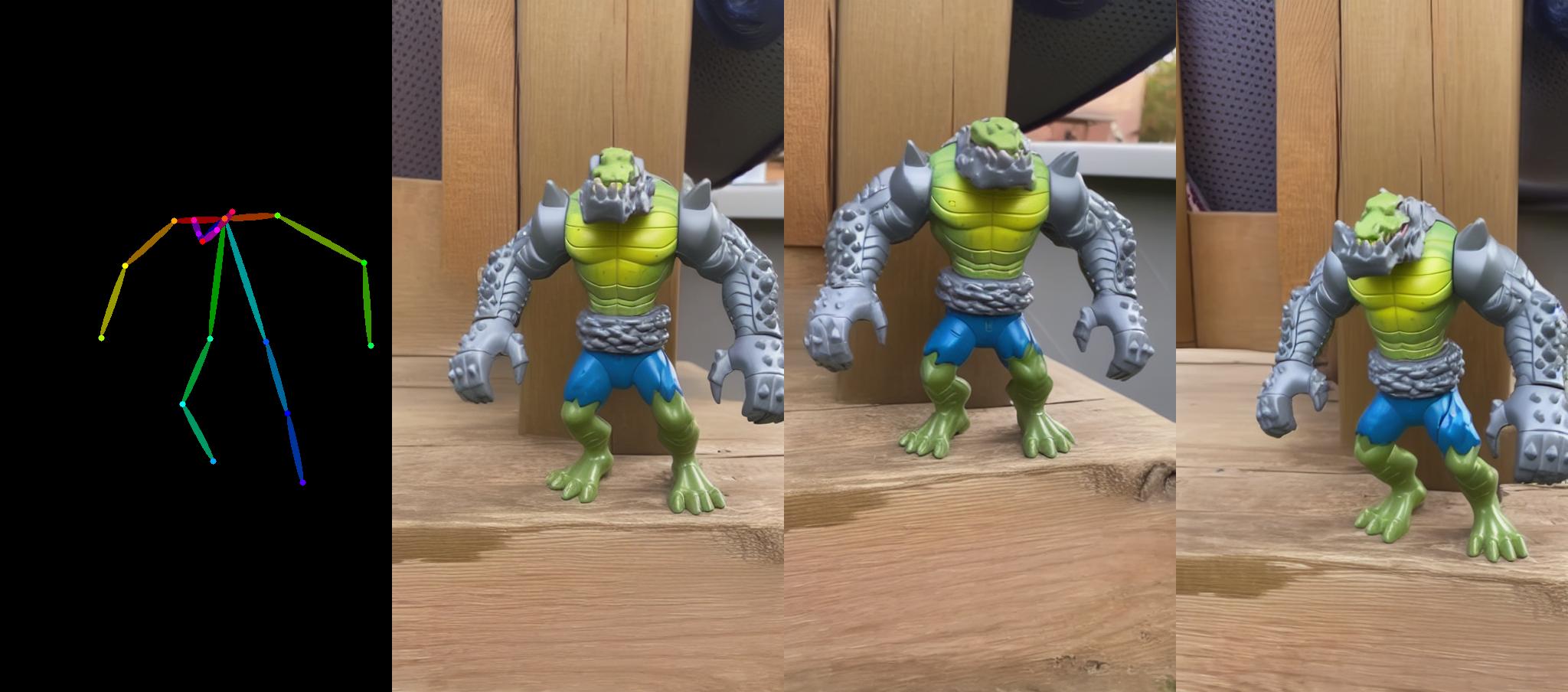} 
    \\
    Original \\
    \includegraphics[width=0.30\textwidth]{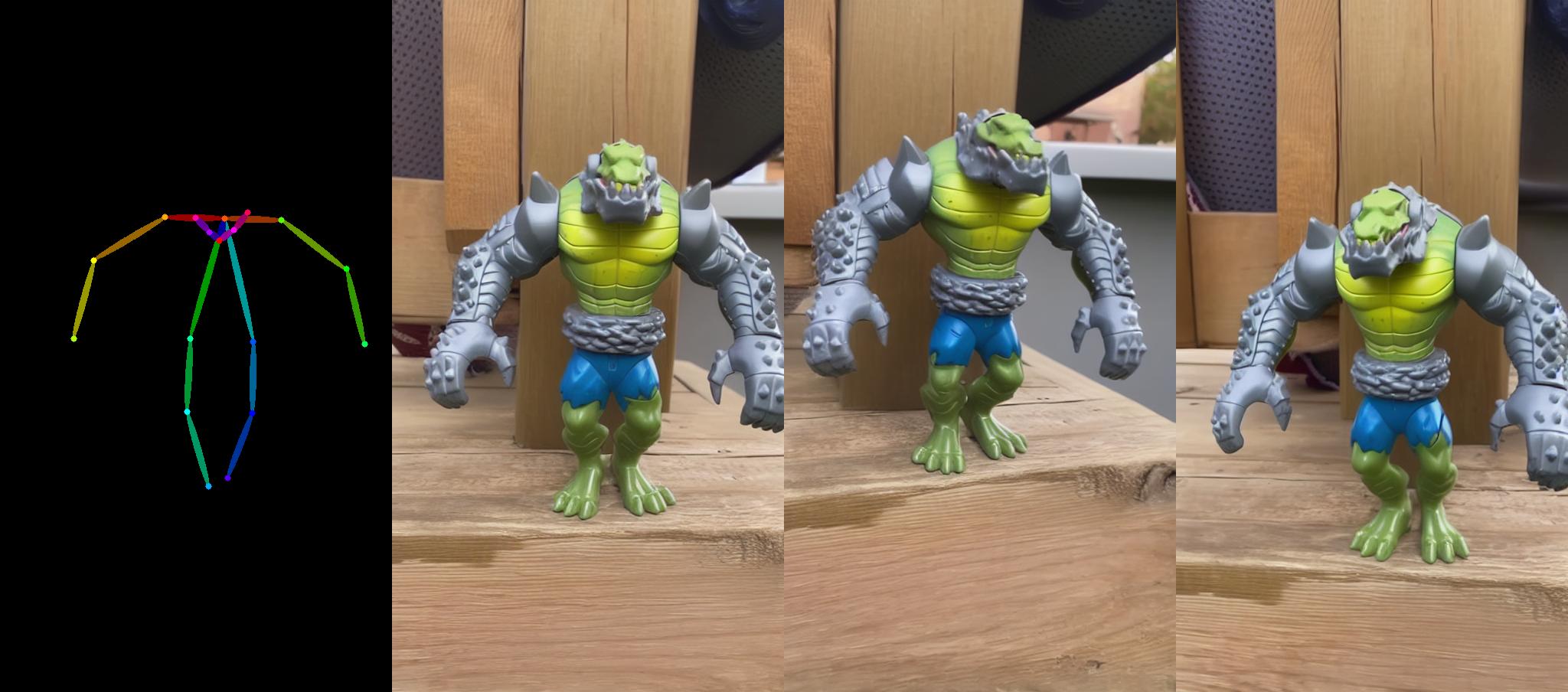} &
    \includegraphics[width=0.30\textwidth]{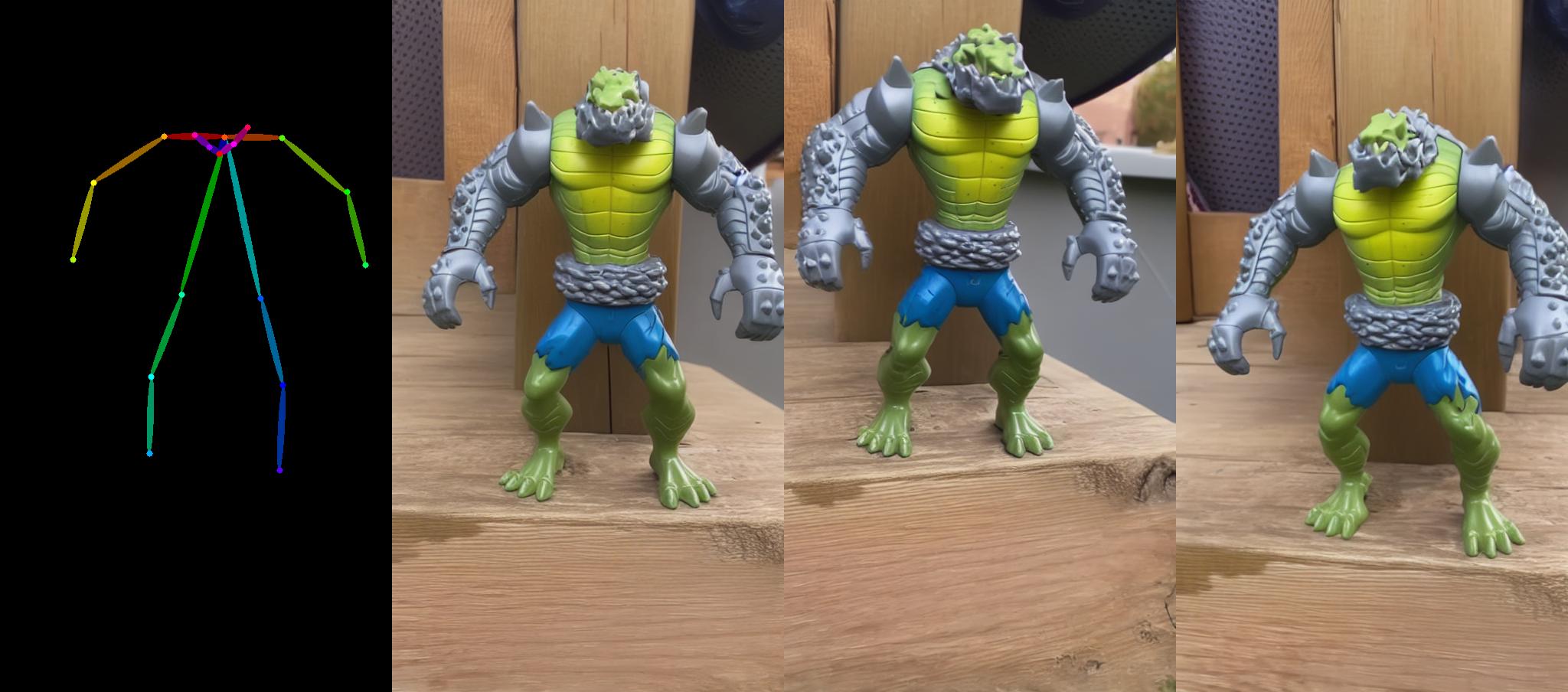} &
    \includegraphics[width=0.30\textwidth]{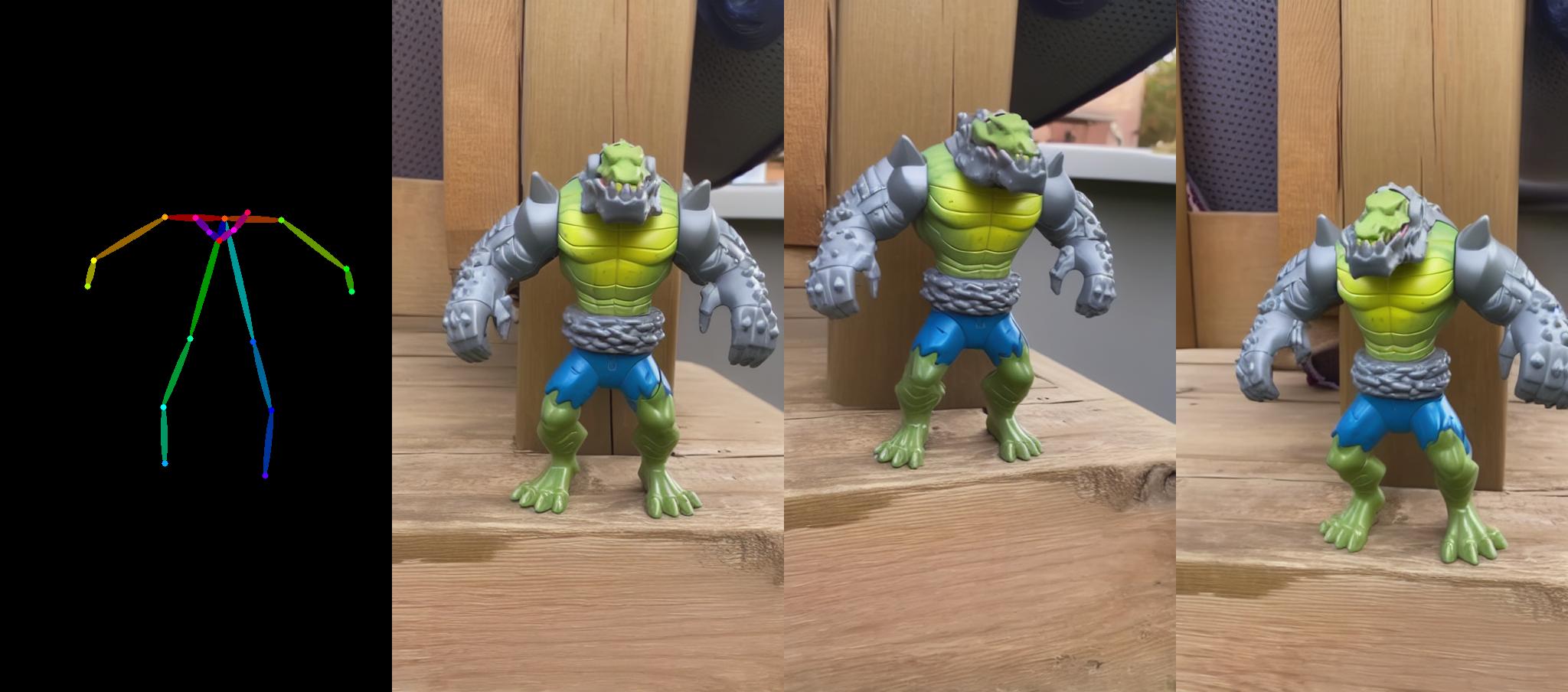} 
    \\

    \includegraphics[width=0.30\textwidth]{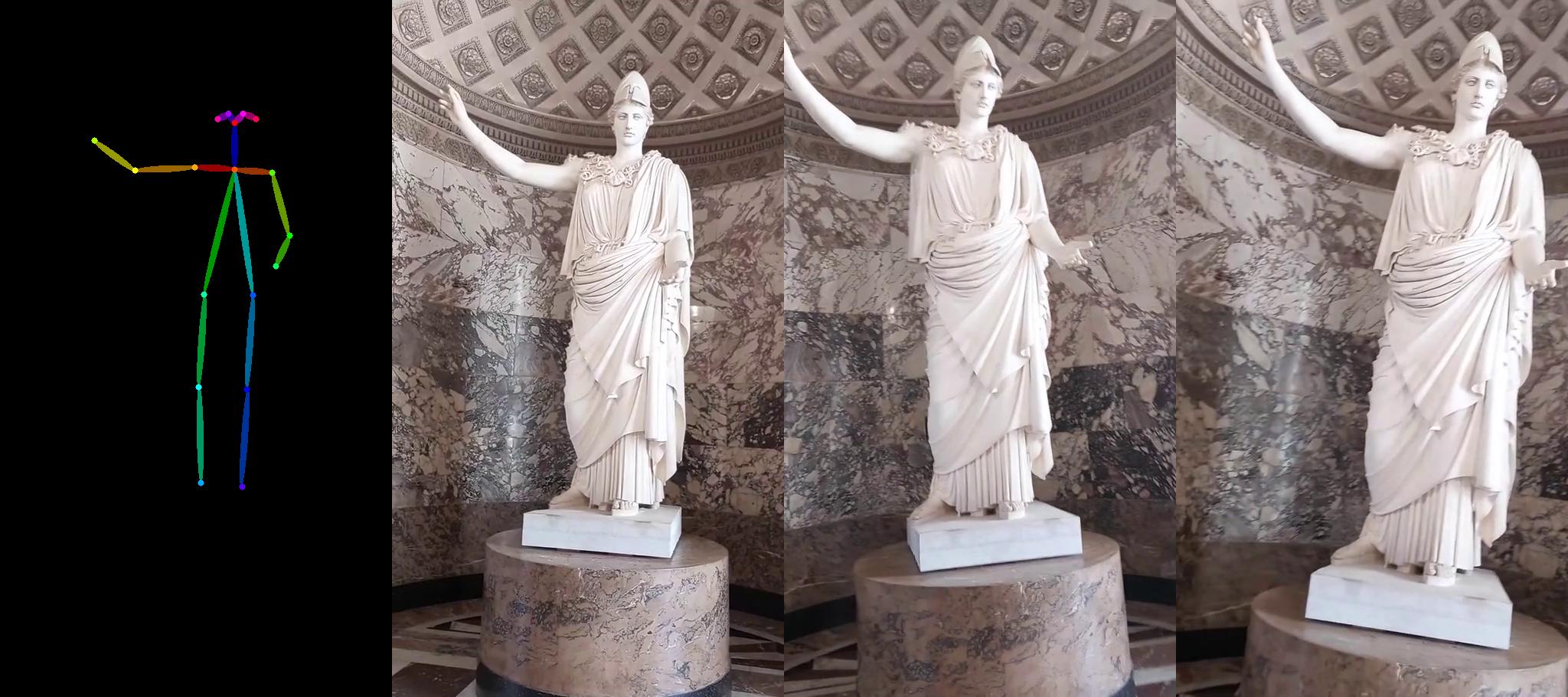} &
    \includegraphics[width=0.30\textwidth]{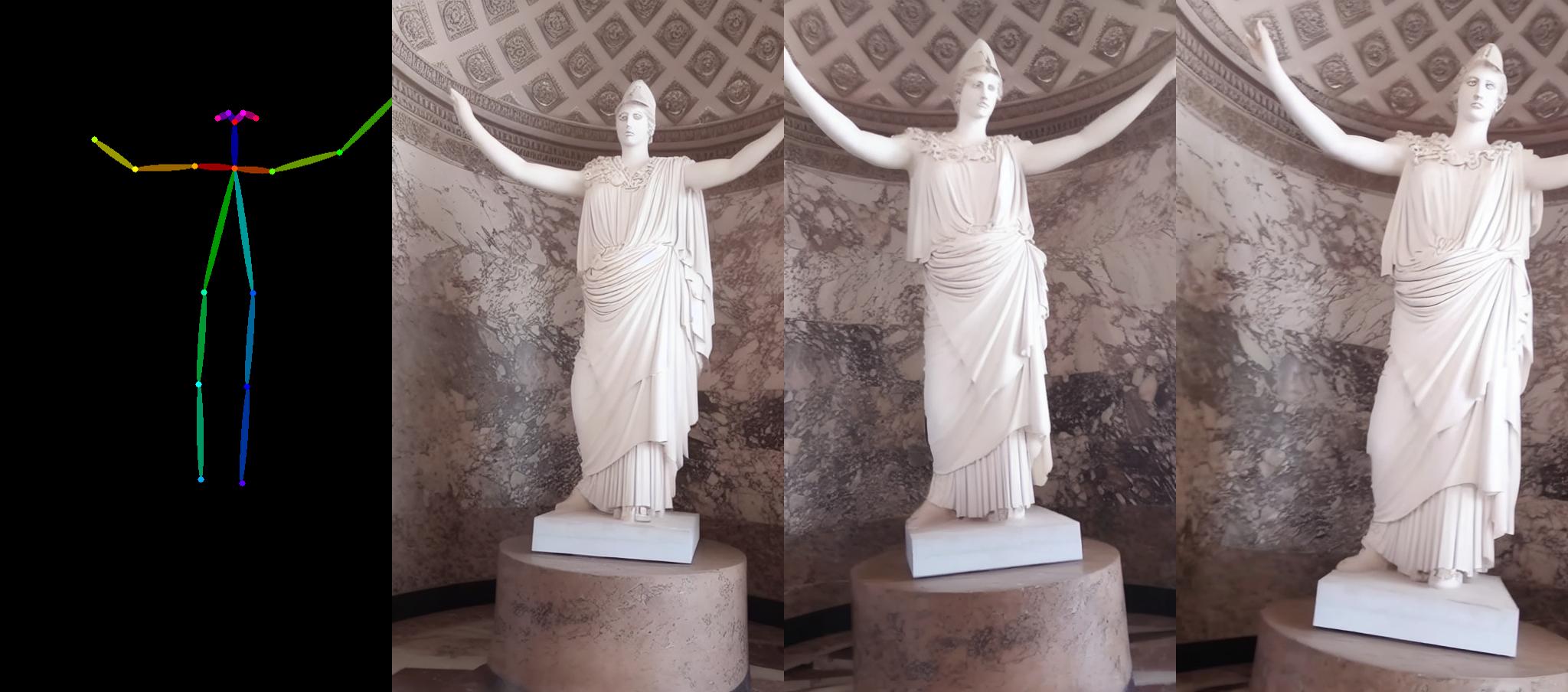} &
    \includegraphics[width=0.30\textwidth]{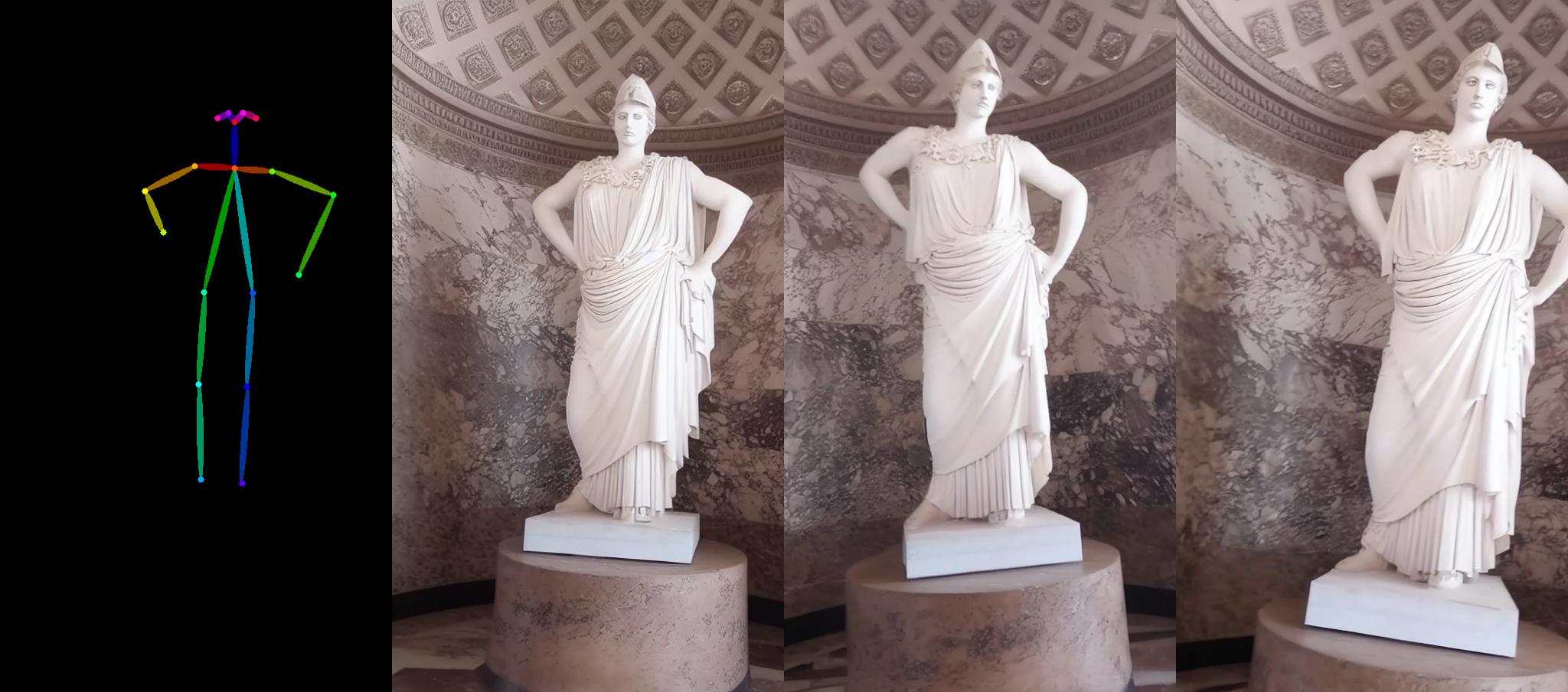} 
    \\
    Original \\
    \includegraphics[width=0.30\textwidth]{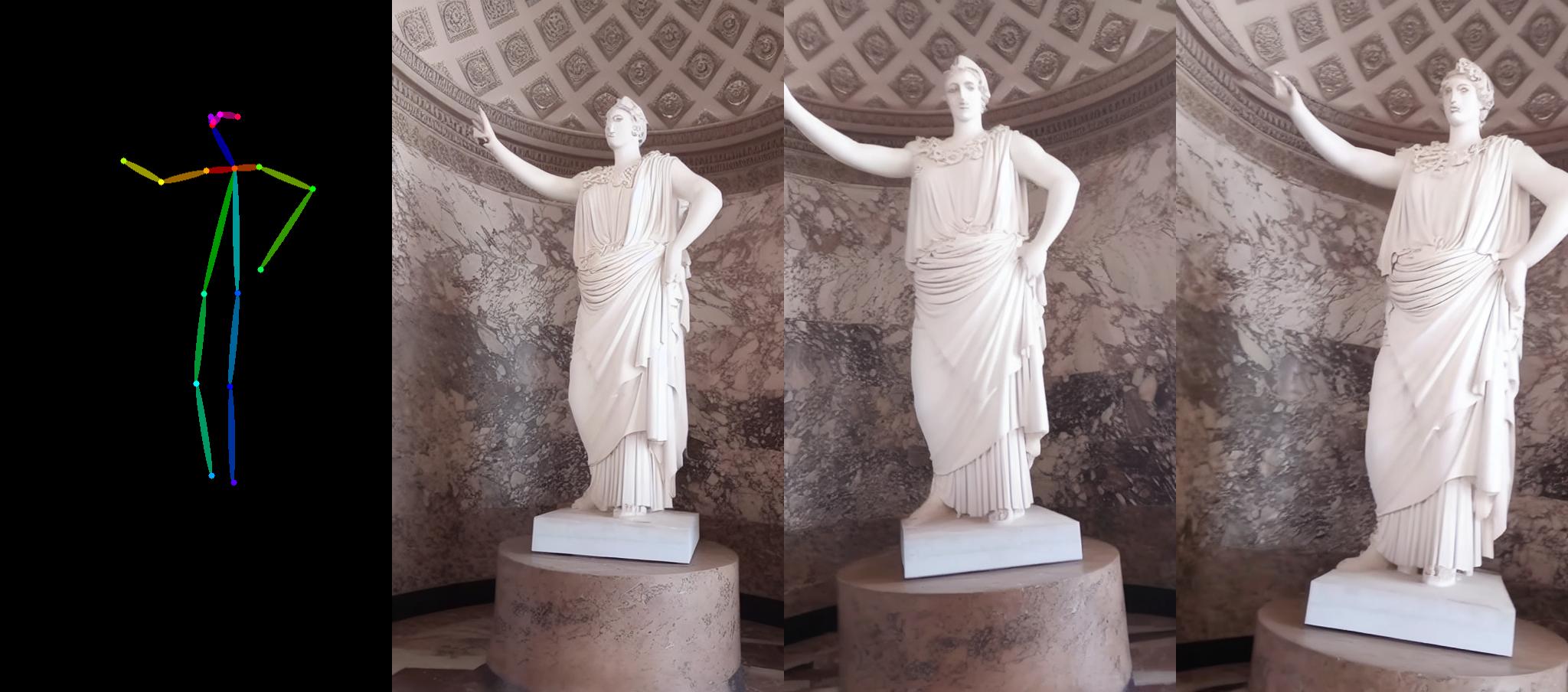} &
    \includegraphics[width=0.30\textwidth]{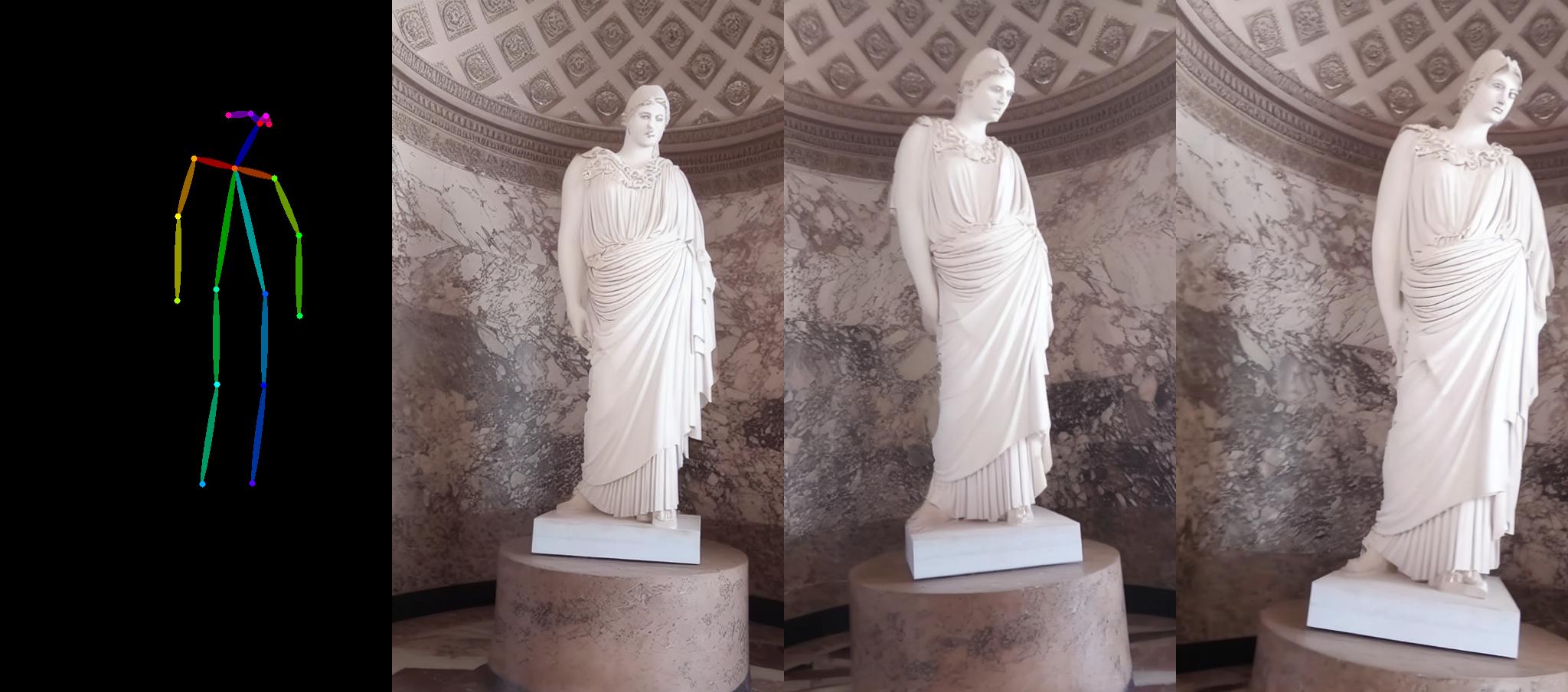} &
    \includegraphics[width=0.30\textwidth]{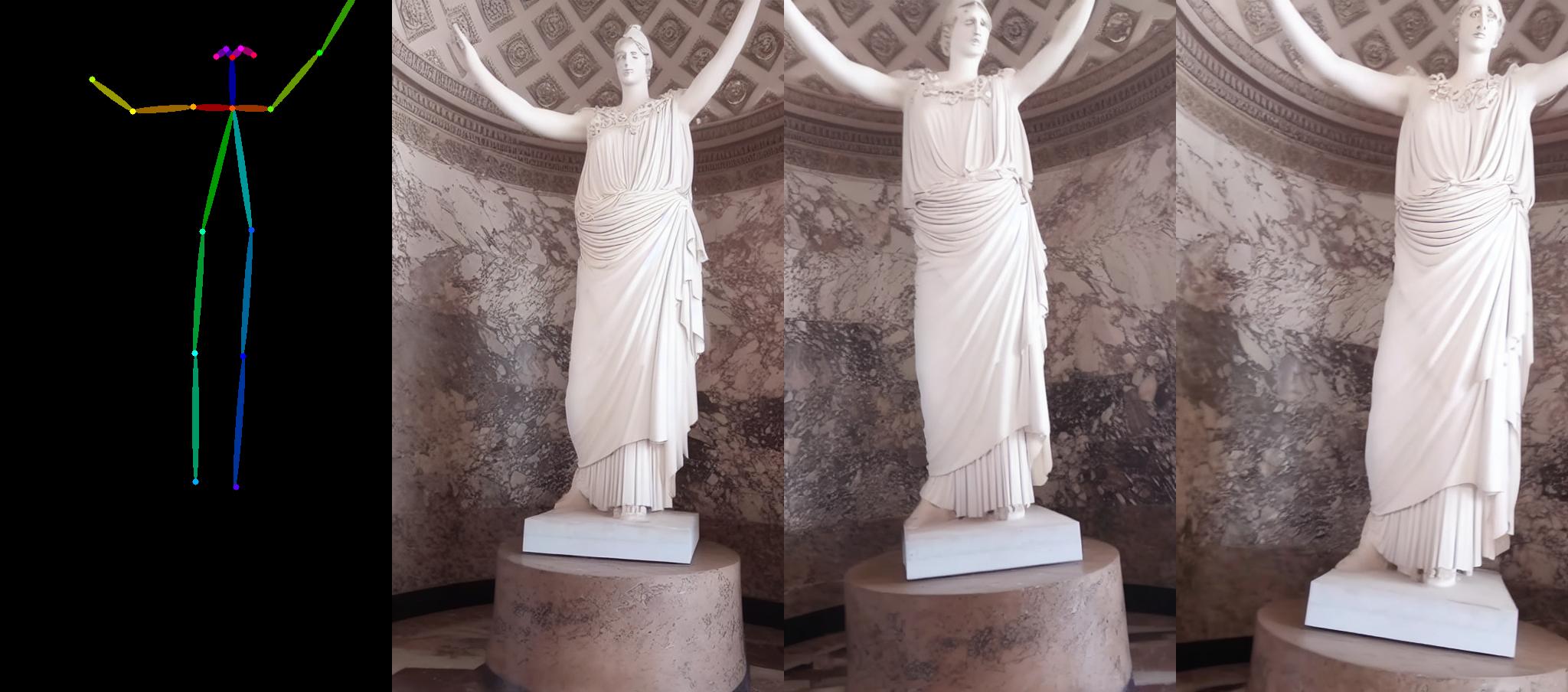} 
    \\

    \includegraphics[width=0.30\textwidth]{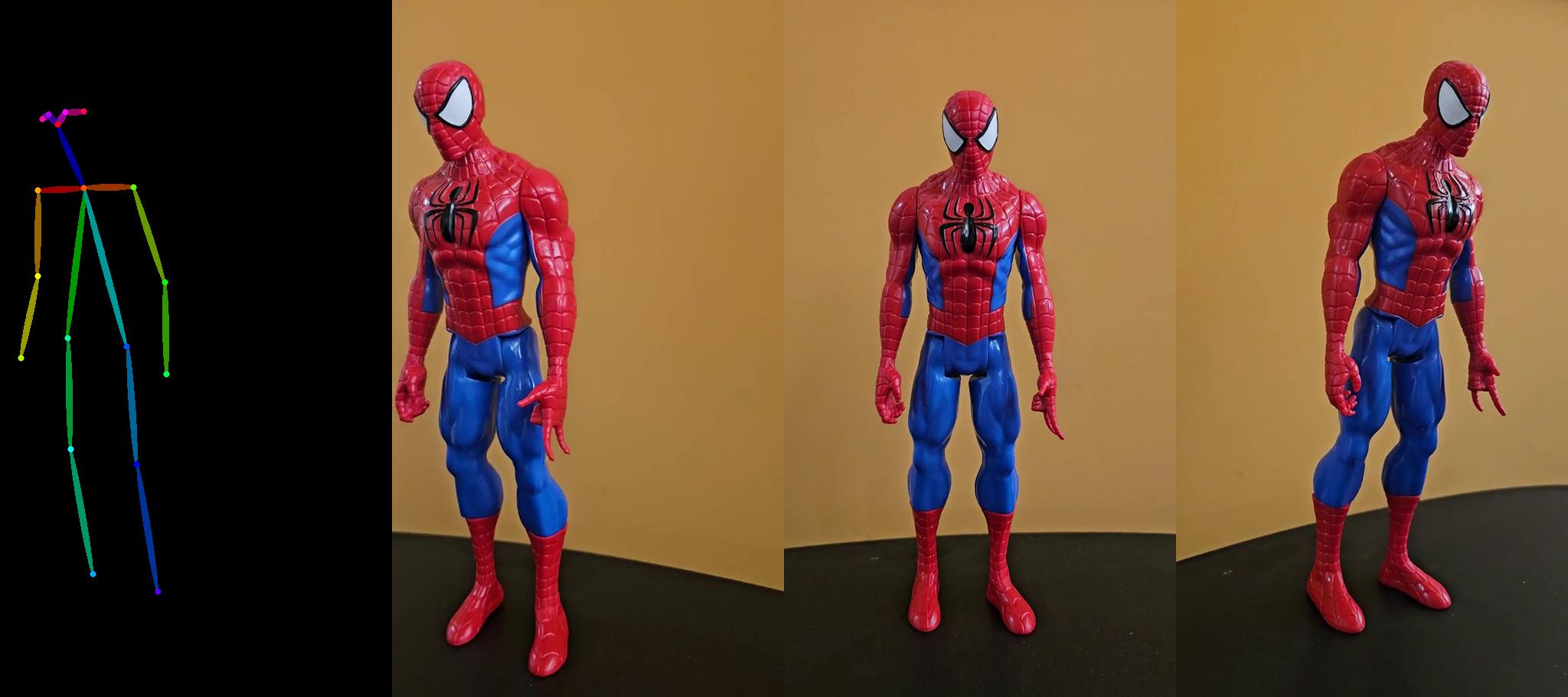} &
    \includegraphics[width=0.30\textwidth]{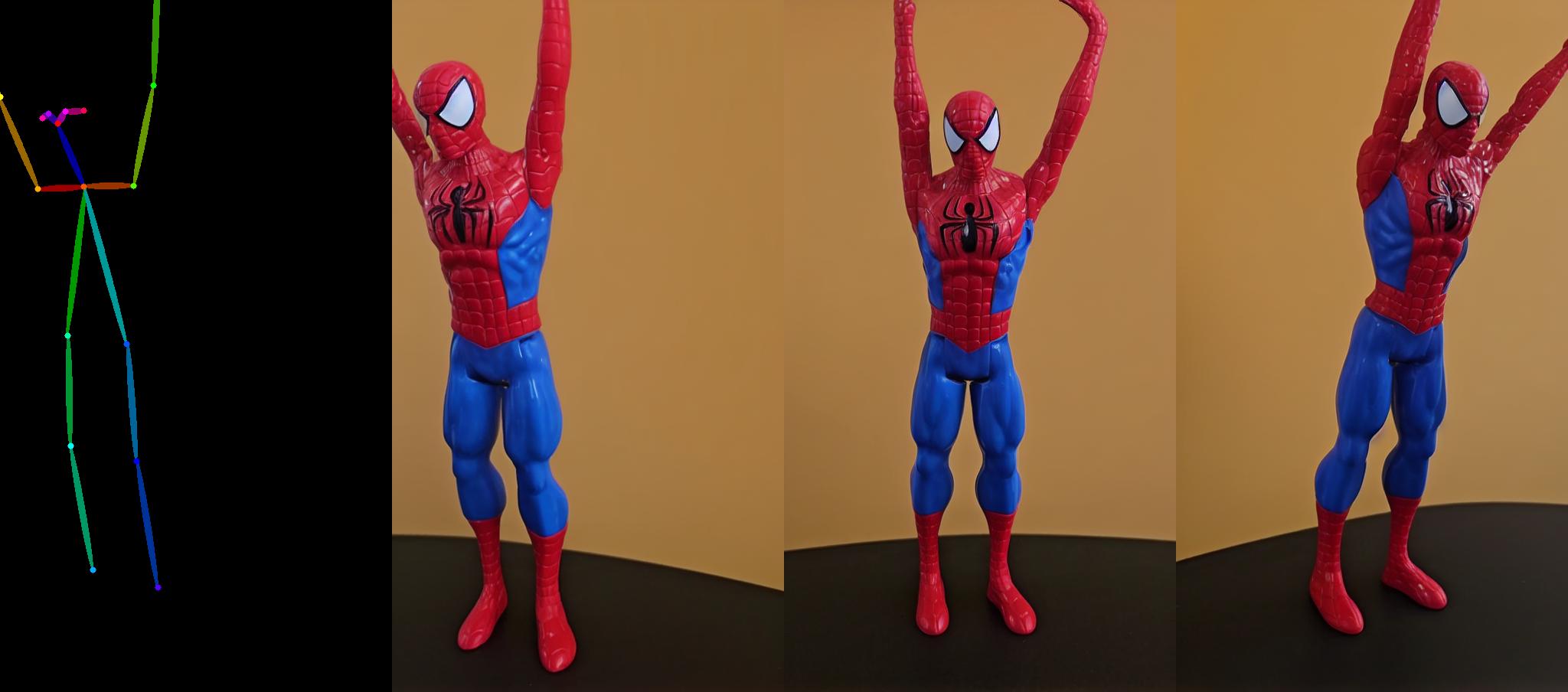} &
    \includegraphics[width=0.30\textwidth]{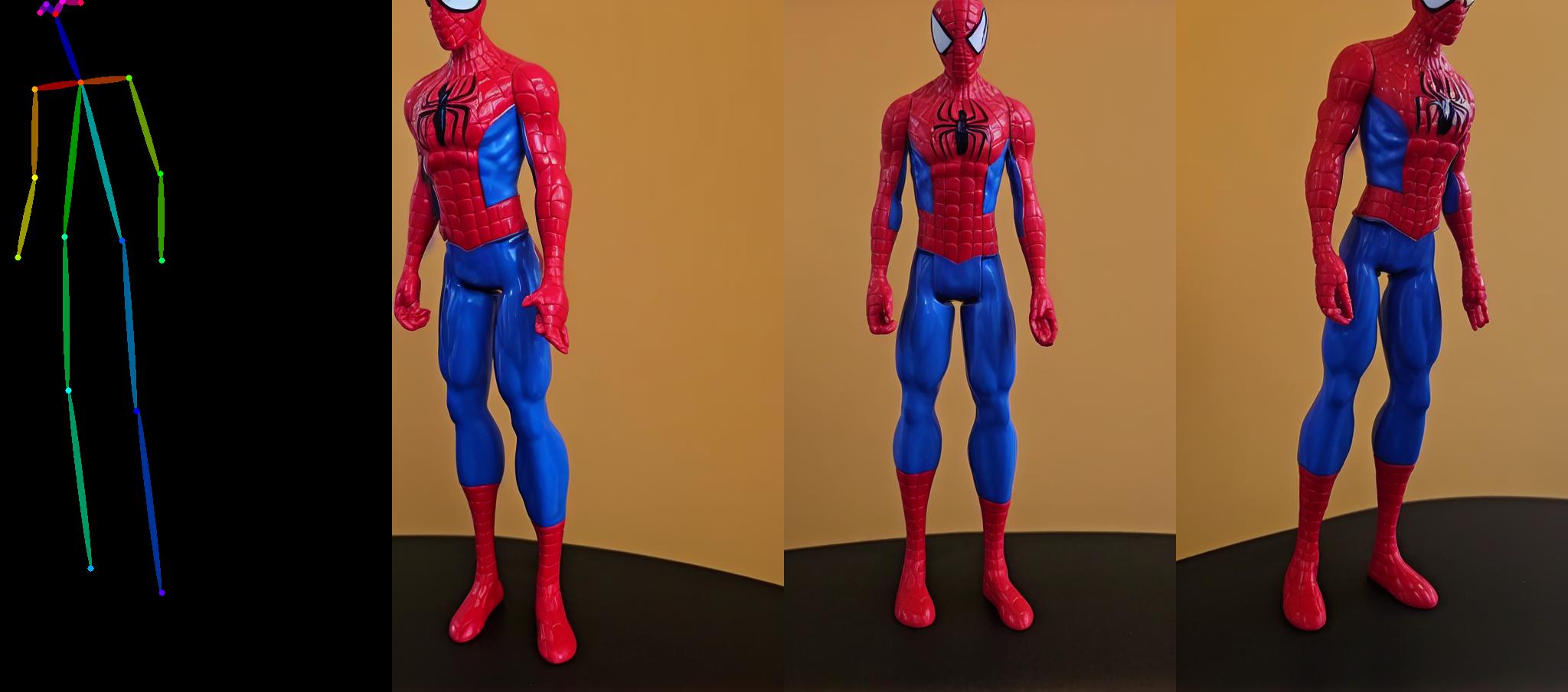} 
    \\
    Original \\
    \includegraphics[width=0.30\textwidth]{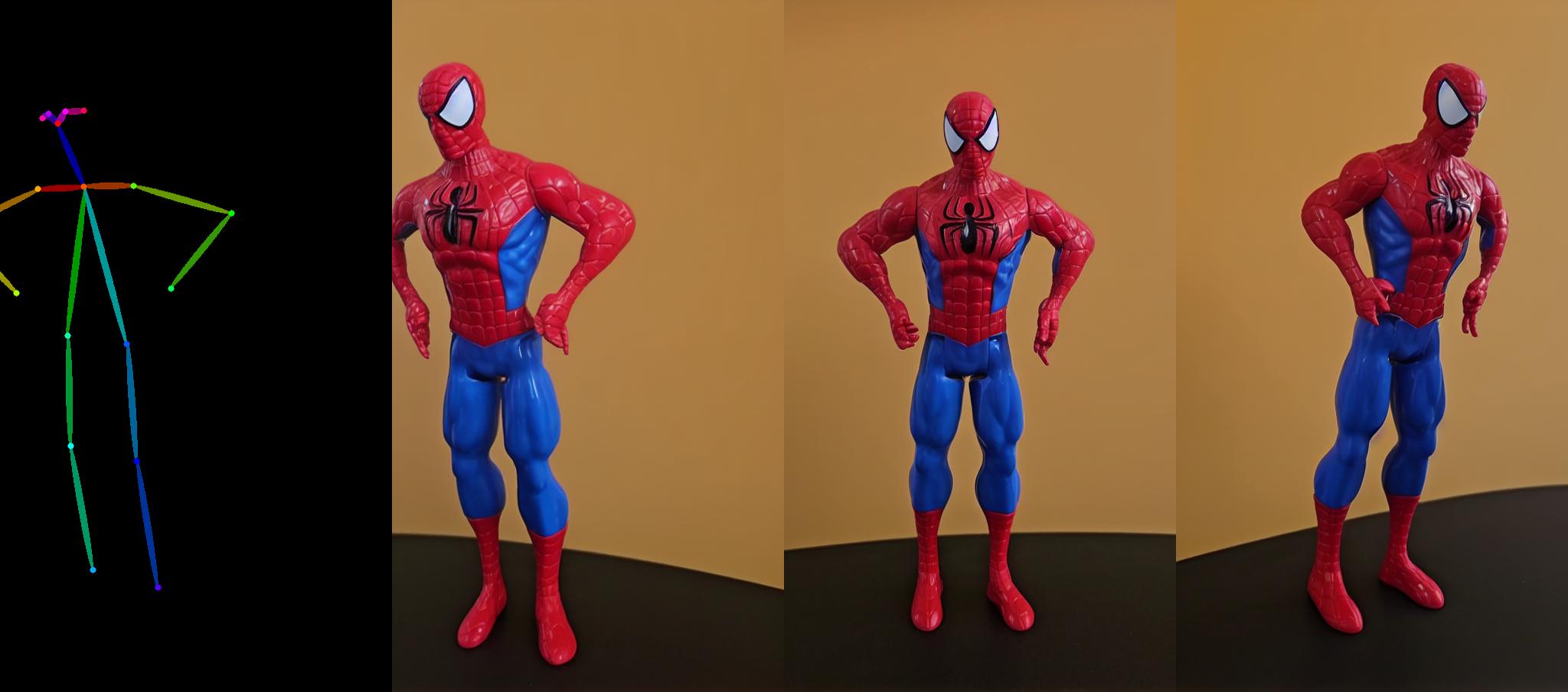} &
    \includegraphics[width=0.30\textwidth]{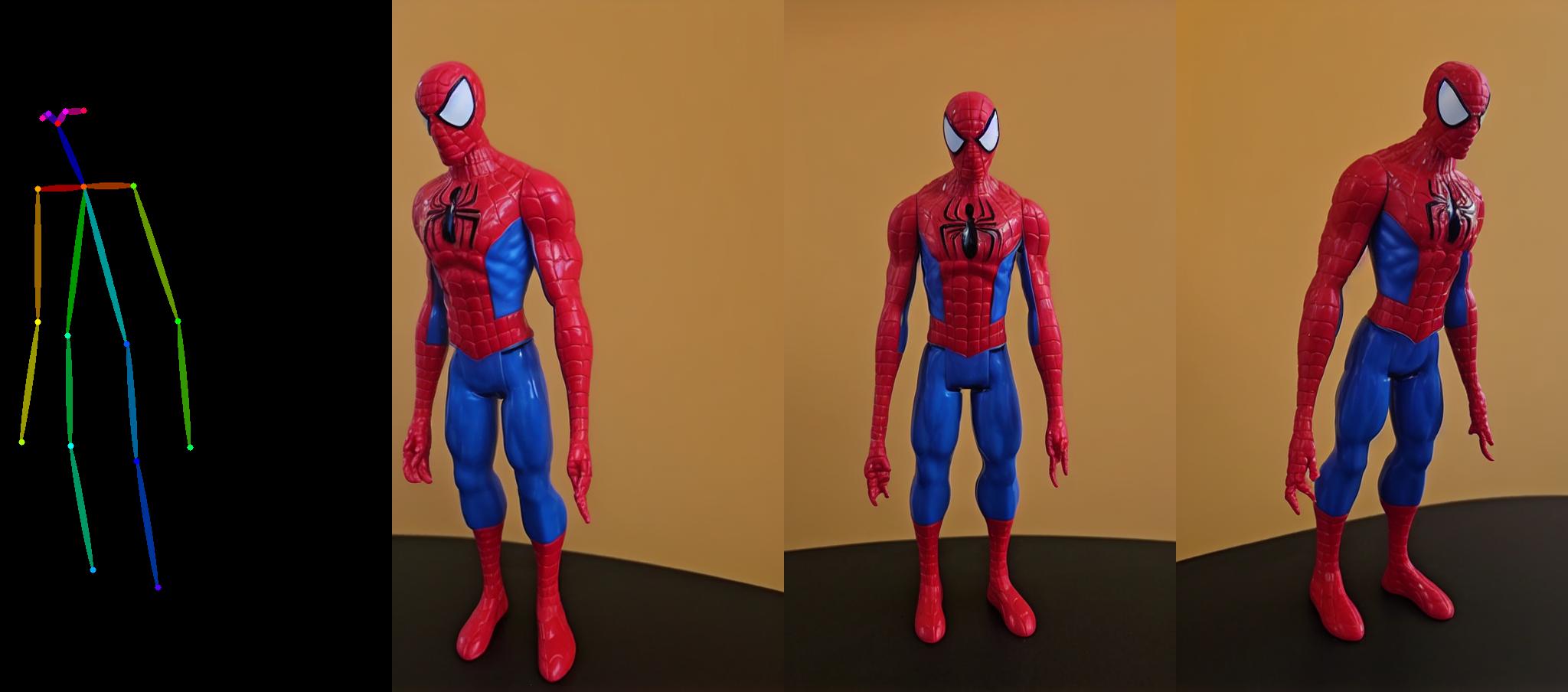} &
    \includegraphics[width=0.30\textwidth]{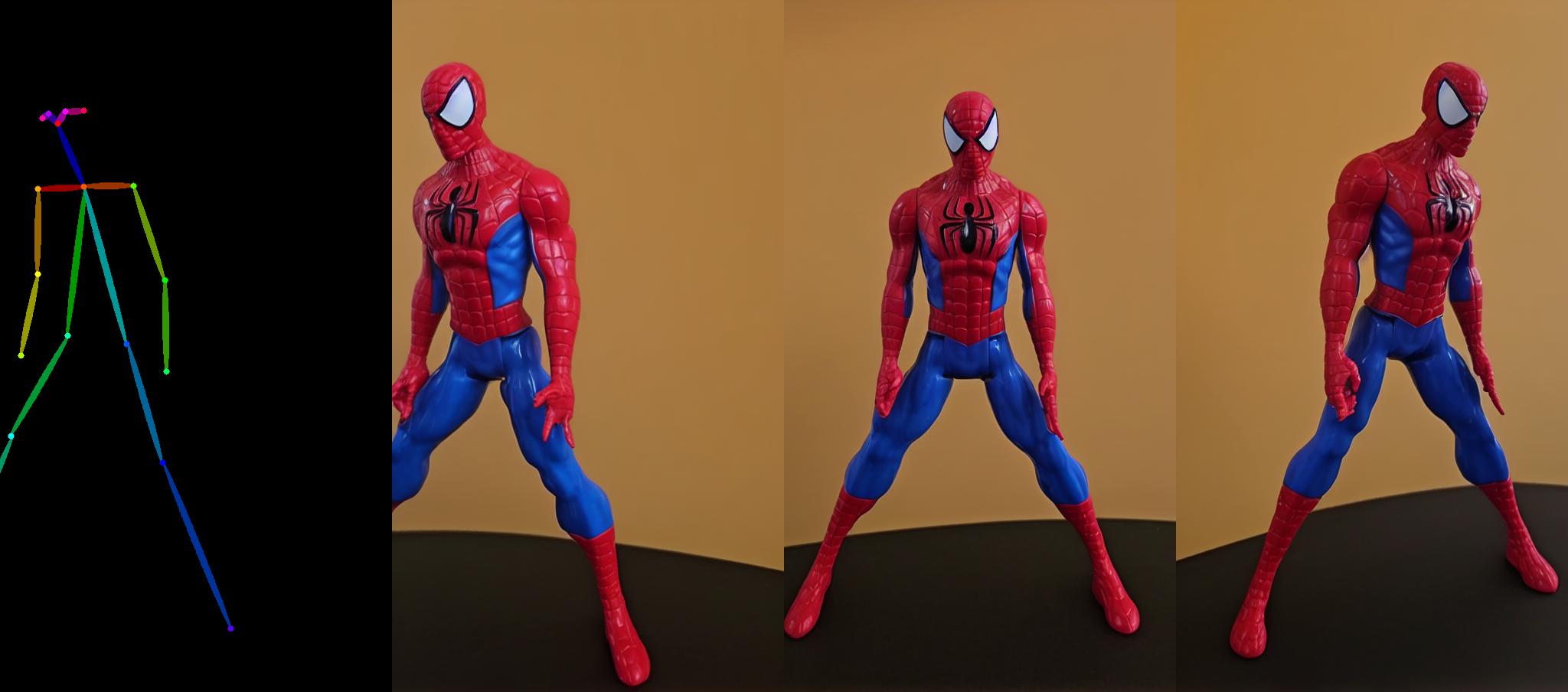} 
    \\

\end{tabular}
\caption{
Qualitative results of our method. Here we edit the images with a skeleton and show a sample of three different views for each example.
}
\label{fig:additional-page-1}
\end{figure*}

\begin{figure*}
\centering
\setlength{\tabcolsep}{1pt}
\begin{tabular}{c c}

    \raisebox{25pt}{\rotatebox[origin=t]{90}{Original}} \hspace{1pt} &
    \includegraphics[width=0.78\textwidth]{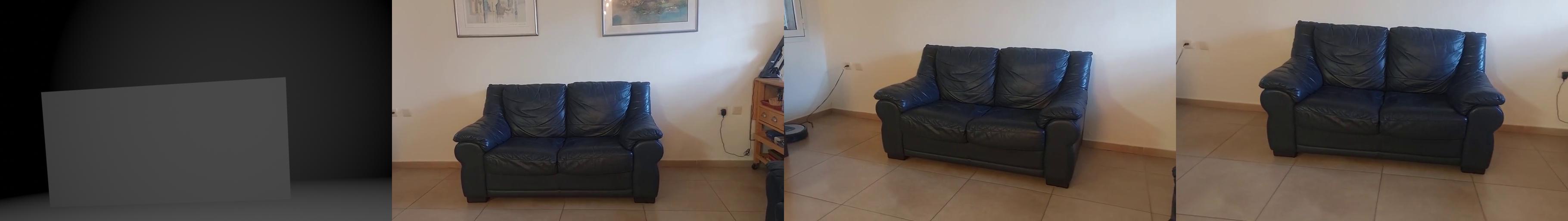} \\
    & \includegraphics[width=0.78\textwidth]{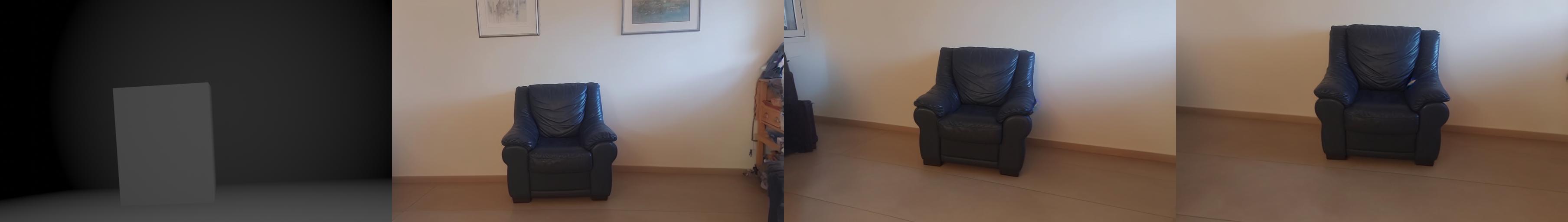} \\
    & \includegraphics[width=0.78\textwidth]{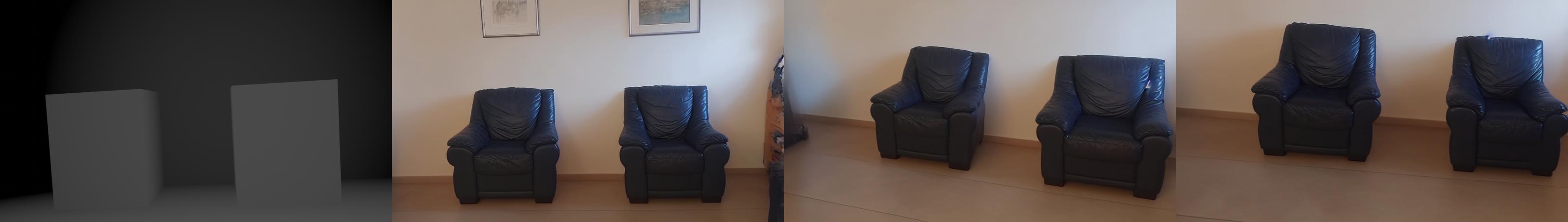} 
    \\

    \raisebox{25pt}{\rotatebox[origin=t]{90}{Original}} \hspace{1pt} &
    \includegraphics[width=0.78\textwidth]{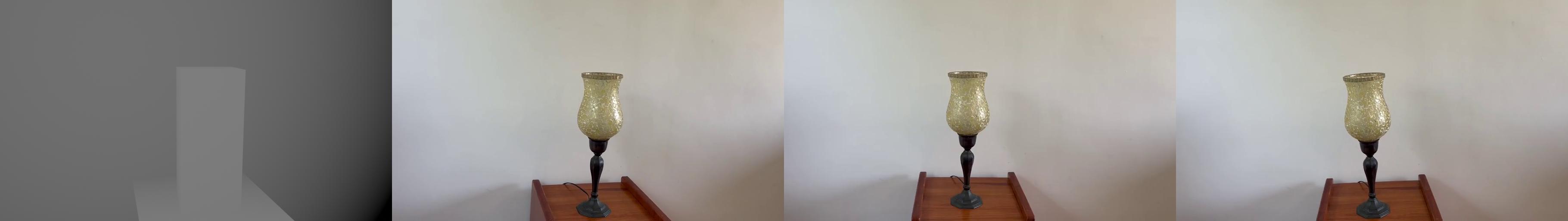} \\
    & \includegraphics[width=0.78\textwidth]{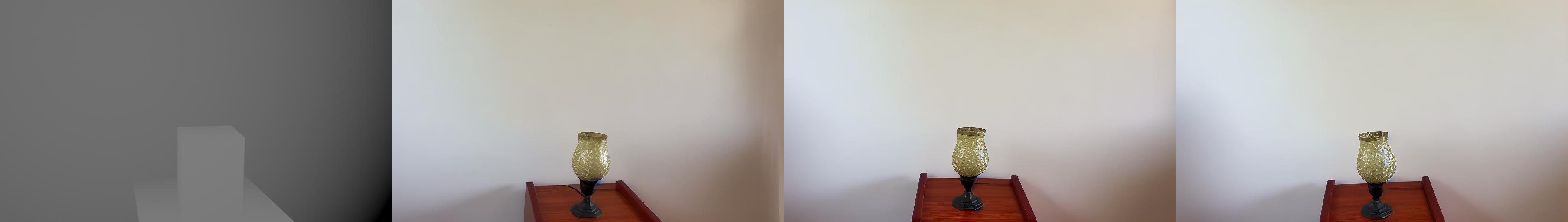} \\
    & \includegraphics[width=0.78\textwidth]{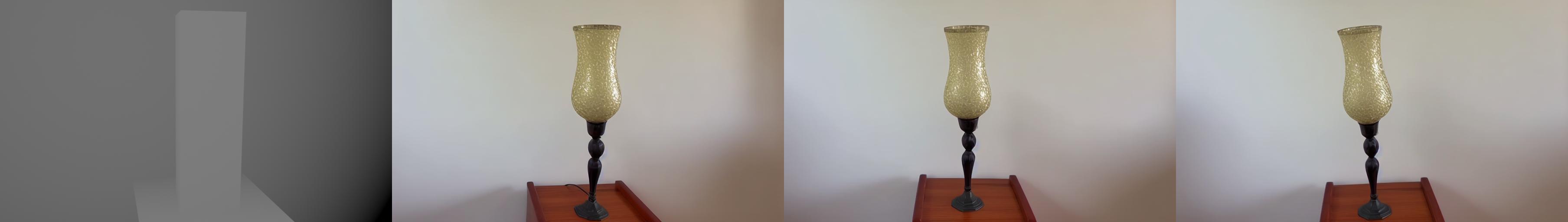} \\
    & \includegraphics[width=0.78\textwidth]{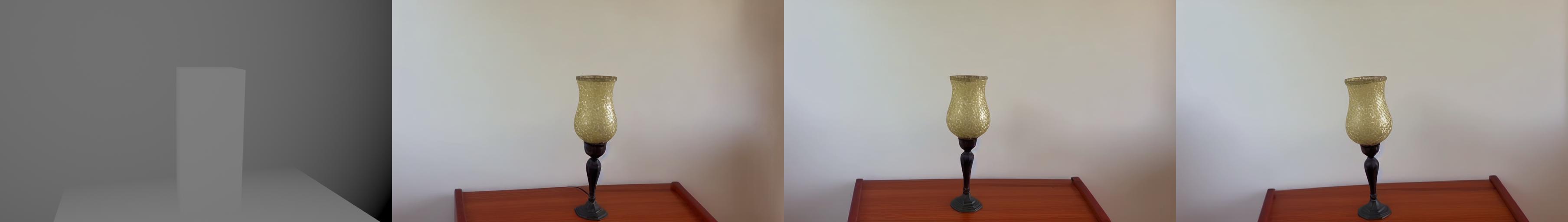}
    \\

    \raisebox{25pt}{\rotatebox[origin=t]{90}{Original}} \hspace{1pt} &
    \includegraphics[width=0.78\textwidth]{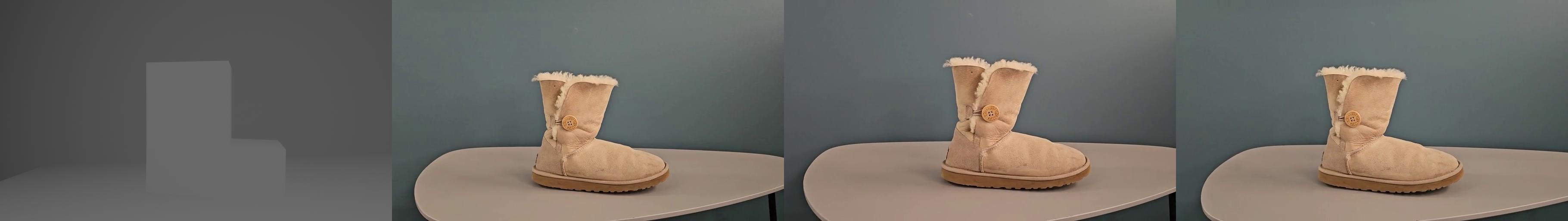} \\
    & \includegraphics[width=0.78\textwidth]{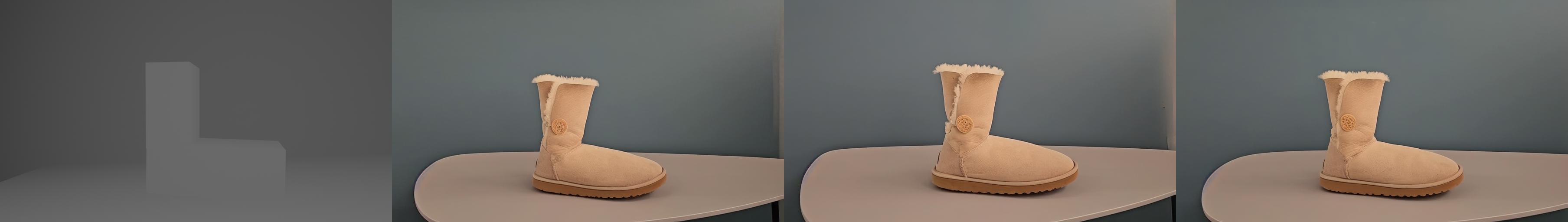} \\
    & \includegraphics[width=0.78\textwidth]{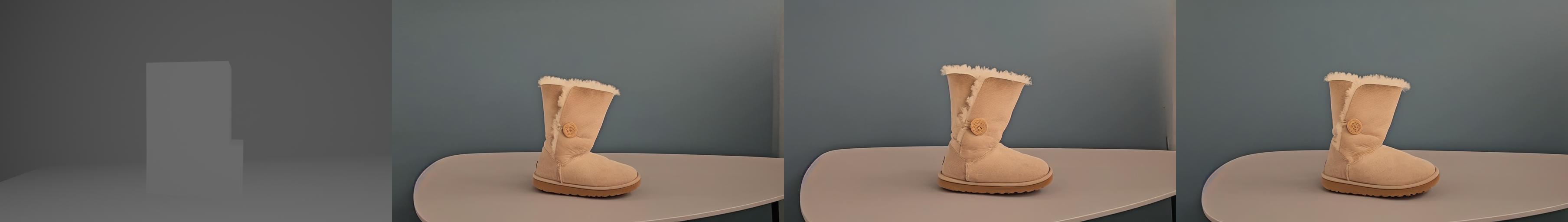}
    \\

\end{tabular}
\caption{
Qualitative results of our method. Here we edit the images with a loose depth map~\cite{bhat2023loosecontrol}, and show a sample of three different views for each example.
}
\label{fig:additional-page-2}
\end{figure*}

More qualitative results of our method are shown in Figures~\ref{fig:additional-page-1}, \ref{fig:additional-page-2}. Additionally, we provide a supplemental video where we show the NeRFs trained on our edited images.

Next, we evaluate our method quantitatively. Our evaluations were conducted over the ``statue", ``person" and ``alligator-toy" scenes (shown in Figures \ref{fig:ablation}, \ref{fig:comparison}). Since we do not have access to ground-truth images or detailed geometry matching the edits, we opt for measuring quality in two ways: (1) Output image quality, as measured by the Kernel Inception Distance (KID)~\cite{binkowski2018demystifying} between edited results and the original images. (2) Quality of final 3D representation, as measured through a user study.

To evaluate image quality, for each scene and edit, we calculate the KID between the outputs of each method and the original scene images. We report the average score across all edits. KID is related to the Fr\'echet Inception Distance (FID)~\cite{heusel2017gans}, but is designed to be more indicative with small (fewer than $50,000$ image) datasets. For completeness, we also report FID. The results are provided in Table~\ref{tab:quant}. In all scenarios, our method achieves improved visual fidelity, indicating that it remains more faithful to the original scene, with fewer visual artifacts when compared to the baselines.

To evaluate the quality of the 3D representation, we conducted a user study where we showed technical users a video of the depth maps extracted from the NeRFs trained for each method. We then asked the users to rank the videos by their alignment with the target pose, and by their quality. The depth maps of competing methods commonly contain holes and clouds created by the NeRF to circumvent inconsistent geometry across the views, hence we expect them to score lower among users. 
We collected $120$ responses from $20$ unique users. In Table \ref{tab:quant} we report the average rank and winrate for each method, averaged over all scenes and edits. Our method was preferred to the baselines in a majority of cases, indicating that the 3D representations extracted from our images were better aligned with the desired edit and of higher visual quality.

\vspace{-4pt}
\section{Conclusions, Limitations, and Future work}
\vspace{-4pt}

We presented a technique to consolidate the results of multi-view editing.
We introduce QNeRF as a means to progressively consolidate the attention features of the images throughout the editing process.
Our approach is generic, making it applicable to various diffusion-based editing techniques where the image layouts are modified. Here, we demonstrated our approach with two types of controls: articulations, and rough bounding boxes. These conditions are intentionally lenient, offering ease of control.

Our work is based on the generative power of text-to-image models. However, it also inherits their common weaknesses. For example, the model struggles to generate human hands. 
Similarly, the model may still hallucinate fine details. In our work, we focused on consolidating the shape, however, in highly detailed objects, these fine details are not consistent despite the shared underlying shape. Similar hallucinations can be noticed in detailed background regions which are dis-occluded by the geometric manipulation.
These inconsistencies can lead to blurry regions when training a NeRF on the edited multi-view images. 
See the supplementary for examples.

In our work, we optimize the QNeRF with a black-box optimizer. This may cause it to ``average'' over outliers, even when it could be more beneficial to filter them out by applying robust statistics techniques. Additionally, we envision exploring alternative means for consolidating features, including the utilization of other three-dimensional representations like Gaussian Splats~\cite{kerbl3Dgaussians}.

\vspace{-4pt}
\section*{Acknowledgement}
\vspace{-4pt}

We thank Gaurav Parmar, Maxwell Jones, Guy Tevet, Kangle Deng, and Or Perel for their early feedback and fruitful discussion. 
We also thank Kfir Aberman, Yuval Alaluf, Ruihan Gao, Songwei Ge, Oren Katzir, Sean Liu, Sigal Raab, and Guy Tevet for proofreading our manuscript and for their useful suggestions. This project is partly supported by Packard Fellowship, the Sony Corporation, and Cisco Research.

{
    \small
    \bibliographystyle{ieeenat_fullname}
    \bibliography{sample-base}
}

\clearpage
\appendix
\section*{\LARGE Appendix}
\section{Implementation Details} \label{sec:implementation-details}

\paragraph{\textbf{Skeleton Editing}}
Given a set of multi-view images of a scene, we begin by extracting the 2D pose using OpenPose~\cite{openpose}. Then, we triangulate the keypoints to obtain a 3D skeleton depicting the subject in the scene which can then be edited (\eg, in Blender~\cite{blender}). We obtain the edited 2D skeletons by projecting the edited 3D skeleton using the camera viewpoints of the original images.
\vspace{-8pt}

\paragraph{\textbf{Loose Depth Editing}}
For the examples shown in the paper, we manually modeled the scene using boxes and planes in Blender~\cite{blender}, which took a couple of minutes per scene. This process can be automated with the data processing pipeline of LooseControl~\cite{bhat2023loosecontrol}. After editing the modeled scene, we rendered the its depth using Blender~\cite{blender} with the camera viewpoints of the original images.
\vspace{-8pt}

\paragraph{\textbf{Original Images Preservation}}
As mentioned earlier, we preserve the appearance of the original images by injecting keys and values of self-attention layers, as done in MasaCtrl~\cite{cao2023masactrl}. Specifically, following MasaCtrl, we perform this injection from the fourth denoising step, in the decoder layers of resolutions $32$ and $64$. 
In addition to this injection, we perform a slight LoRA fine-tune~\cite{hu2022lora} of the model on the input images. The model is fine-tuned with a batch size of 2, for 300 steps. For fairness, we use the same fine-tuned model for all baselines.
\vspace{-8pt}

\paragraph{\textbf{Depth Supervision For QNeRF}}
Since we edit only the main subject in the image, the depth around this object should remain as in the original images. We assume that the depth of the queries is the same as the corresponding image. Therefore, we train a NeRF on the original set of multi-view images and render its depth. Then, we mask these depth images using a rough mask of the union of the object before and after the edit~\cite{kirillov2023segany}. We use the depth loss term introduced in DS-NeRF~\cite{kangle2021dsnerf}.
\vspace{-8pt}

\paragraph{\textbf{Hyperparameters}}
We use Stable Diffusion v1.5~\cite{rombach2021highresolution} and perform 50 steps of denoising. At the end of each interval, the QNeRF is trained for 10,000 steps with a depth-loss coefficent 1. For the query guidance, we set $\alpha = 60$ in Equation~\ref{eq:q-guidance} in the main paper.
\vspace{-8pt}

\section{Datasets Details} \label{sec:dataset-details}
The person dataset is from Instruct-NeRF2NeRF. Besides it, all the other datasets were collected by us, using a simple phone camera. Below are the sizes of each of the datasets.

\begin{table}[h]
    \centering
    \scriptsize
    \begin{tabular}{ccccccc}
        \toprule
          Person & Spiderman & Statue & Alligator-toy & Sofa & Lamp & Boot \\
        \midrule
         350 & 348 & 291 & 493 & 333 & 305 & 281 \\
        \bottomrule
    \end{tabular}
    \label{tab:datasets}
\end{table}

\section{Experiments}

\subsection{Comparisons}
\begin{figure}[h]
\centering
\setlength{\tabcolsep}{1pt}
\begin{tabular}{c c c c c}

    \includegraphics[width=0.19\linewidth]{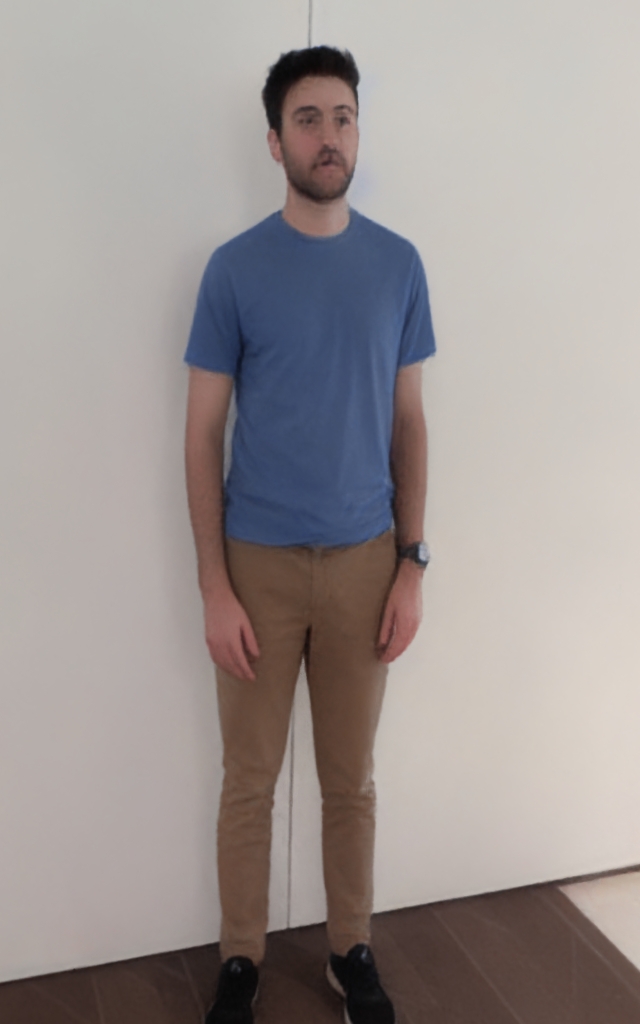} &
    \includegraphics[width=0.19\linewidth]{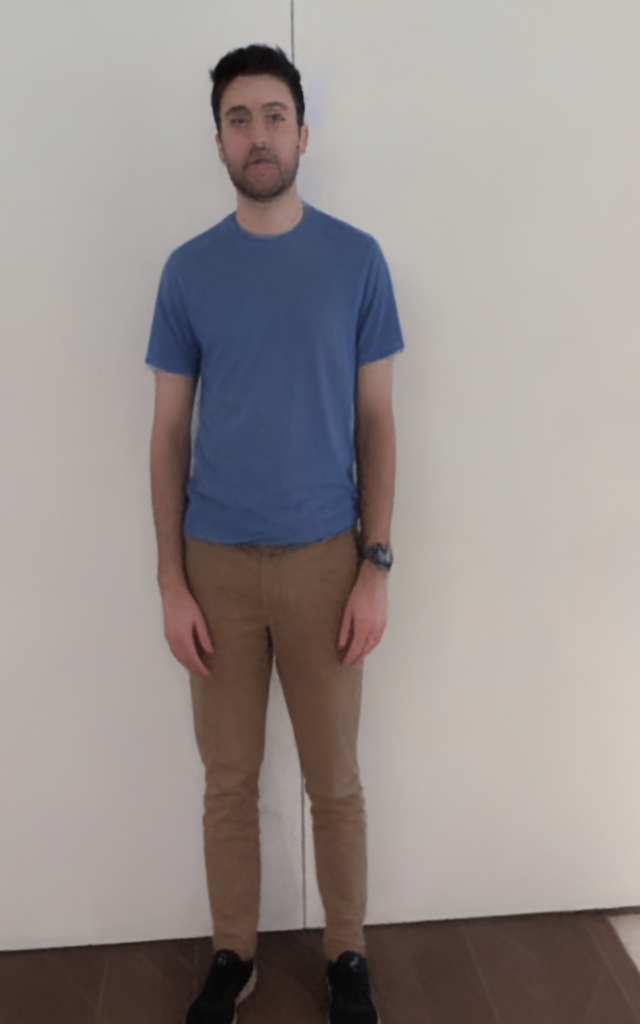} &
    \includegraphics[width=0.19\linewidth]{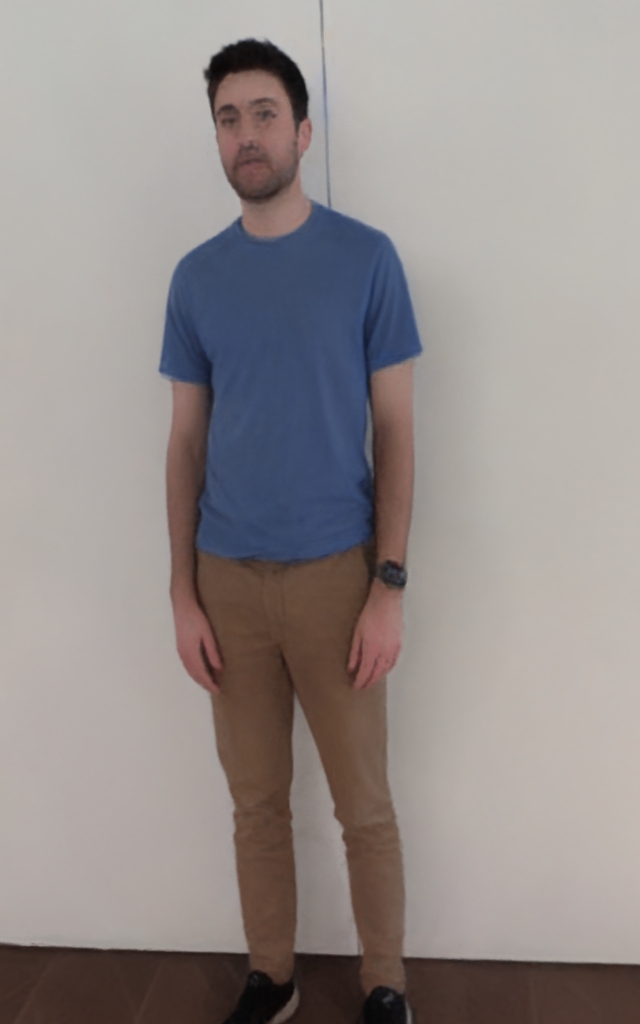} &
    \includegraphics[width=0.19\linewidth]{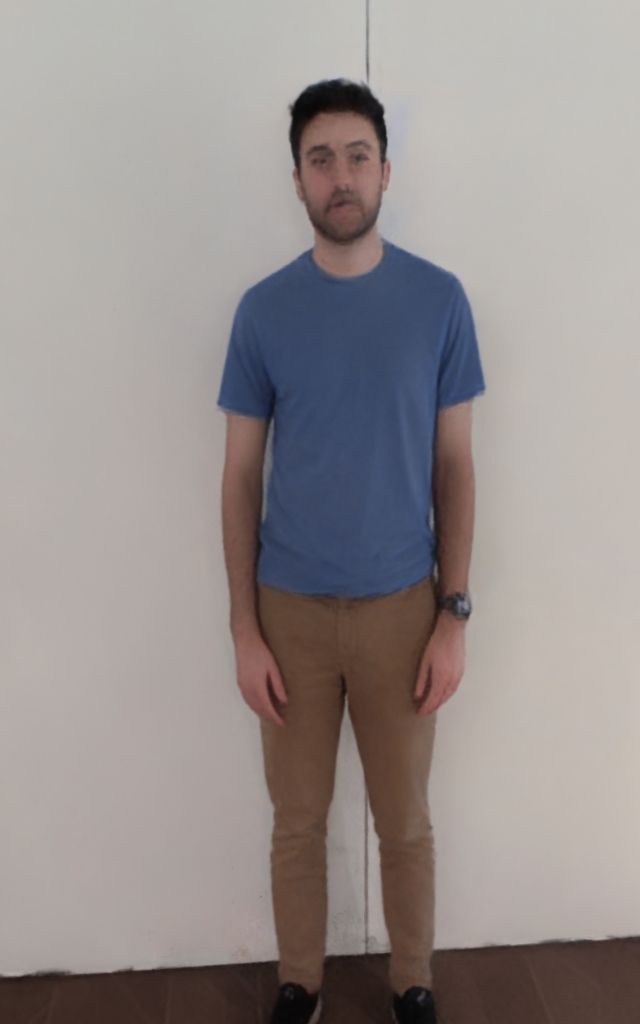} &
    \includegraphics[width=0.19\linewidth]{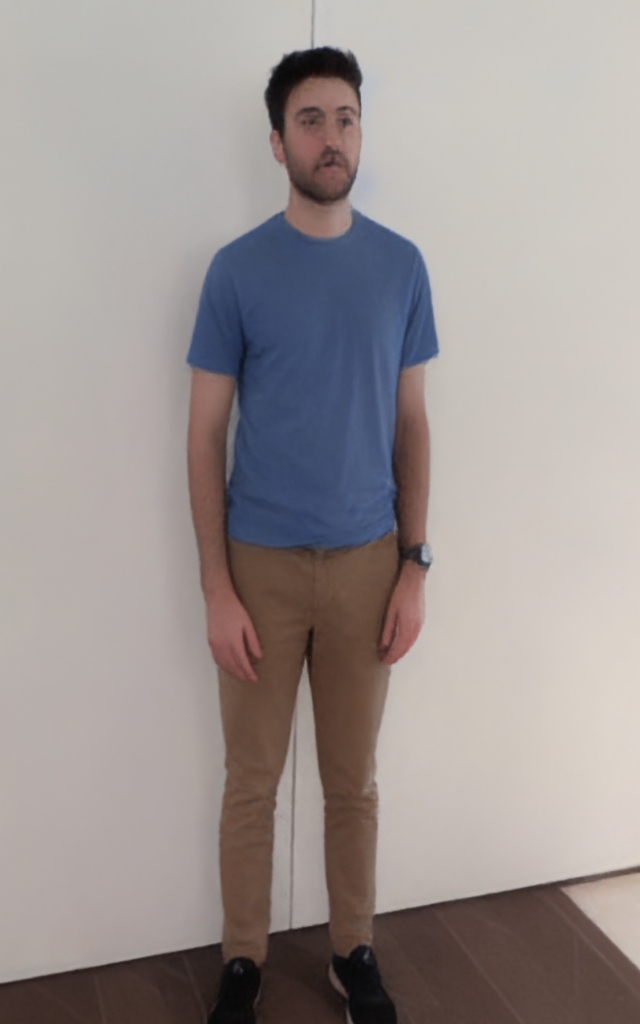} 

\end{tabular}
\caption{
Results of Instruct-NeRF2NeRF for the prompt ``raise his hands''.
}
\label{fig:supp-compare}
\end{figure}

For completeness, we compare our method with the vanilla Instruct-NeRF2NeRF method, using text as an interface for editing. We observe that the NeRF remains as the original one, even when we try different prompts and hyperparameters. 
Results for the prompt ``raise his hands'' are shown in Figure~\ref{fig:supp-compare}.

\subsection{Dataset Size Analysis}
Our consolidation relies on the success of training the QNeRF. Hence, our method requires the same dataset size as the required size to train a NeRF. NeRFs are typically improved with the increase of number of images, and therefore the same applies for our method. It should be noted, that if the multi-view set size is enough to train a NeRF, then it is possible to increase the size of our input set by training a NeRF and rendering novel views. 

\begin{figure*}[tbh]
\centering
\setlength{\tabcolsep}{1pt}
\begin{tabular}{c c c c c c}

    \raisebox{20pt}{\rotatebox[origin=t]{90}{MasaCtrl}} &
    \includegraphics[width=0.19\linewidth]{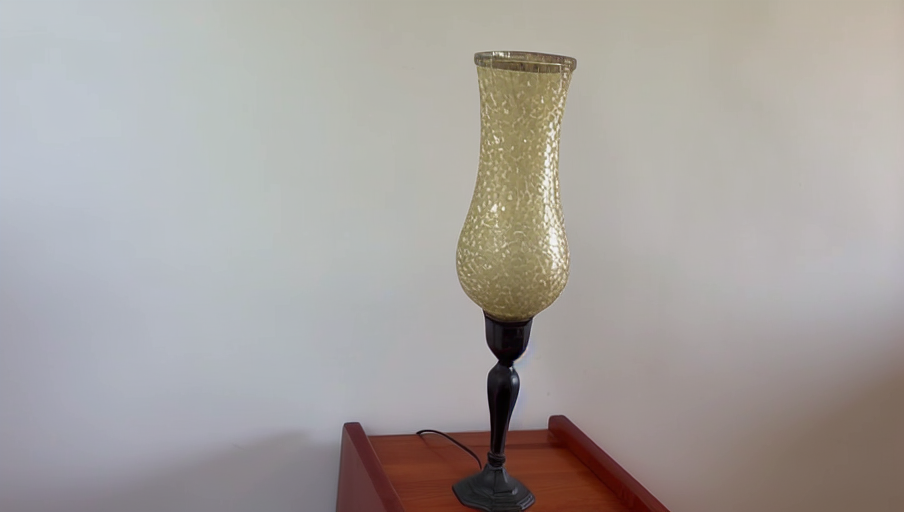} &
    \includegraphics[width=0.19\linewidth]{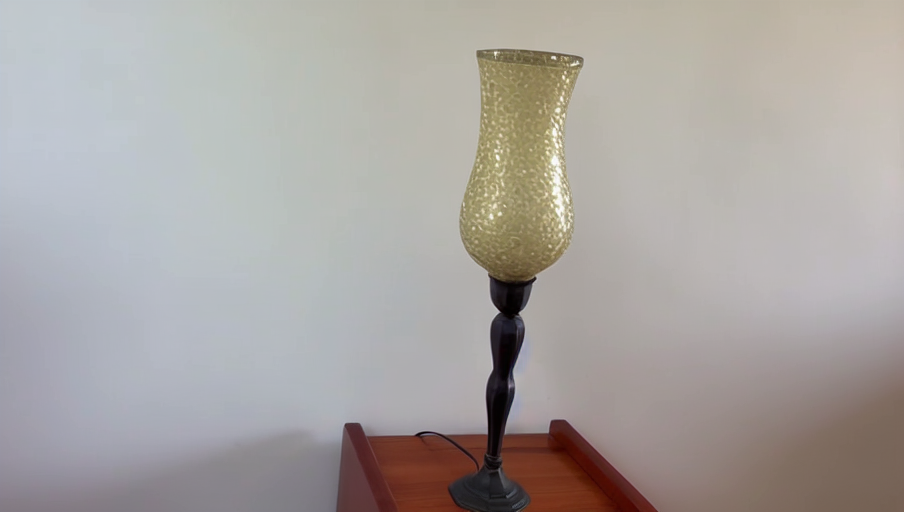} &
    \includegraphics[width=0.19\linewidth]{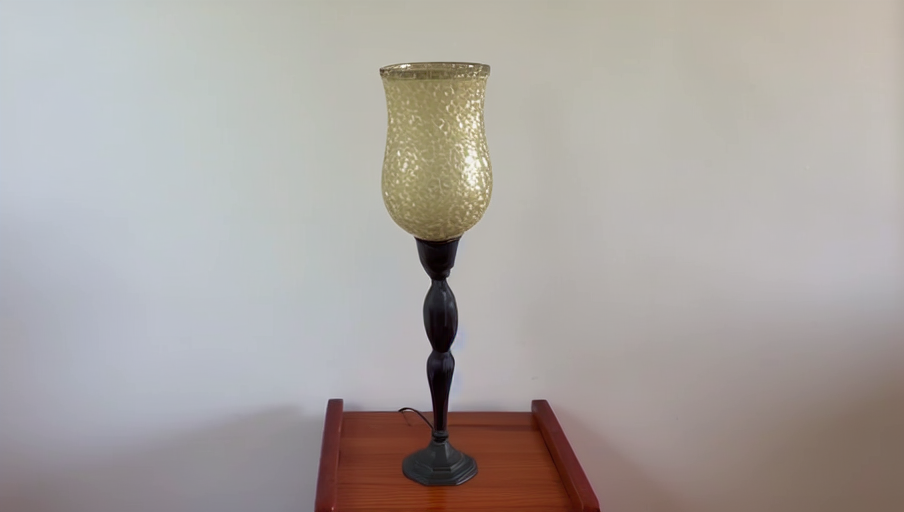} &
    \includegraphics[width=0.19\linewidth]{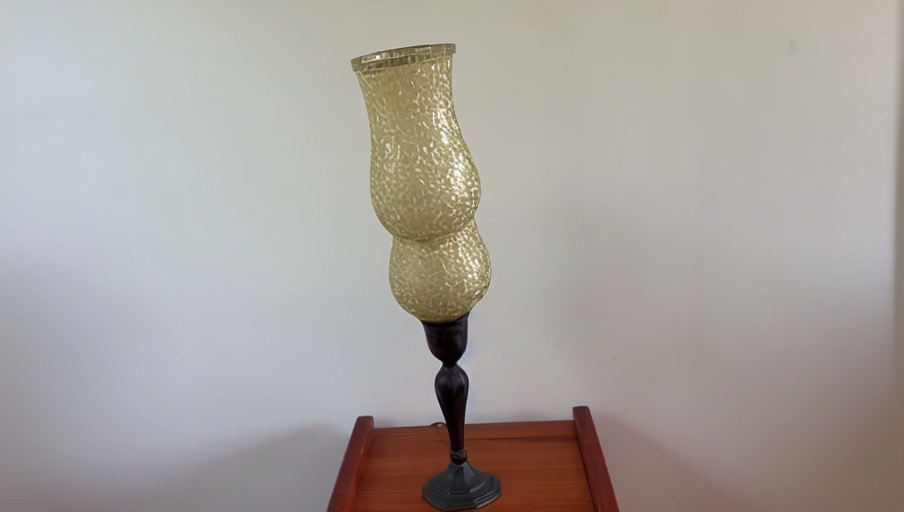} &
    \includegraphics[width=0.19\linewidth]{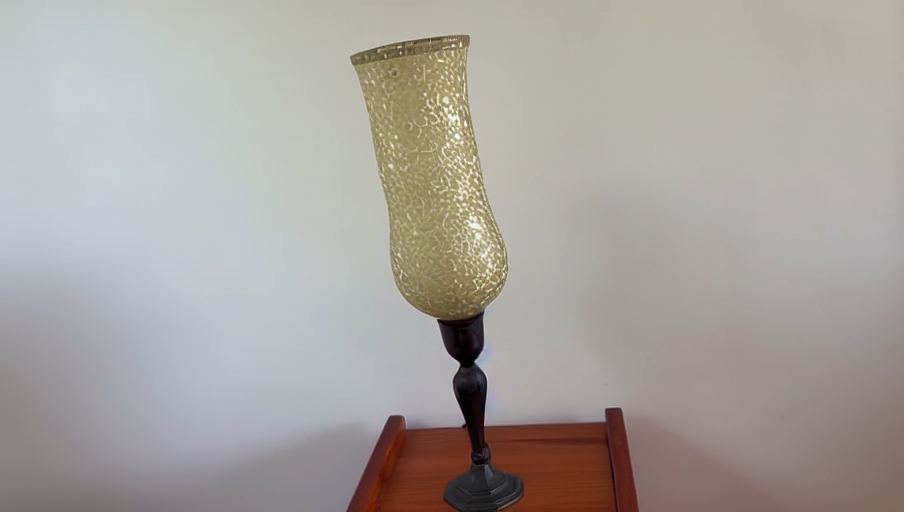} \\

    \raisebox{21pt}{\rotatebox[origin=t]{90}{Ours w/ 50}} &
    \includegraphics[width=0.19\linewidth]{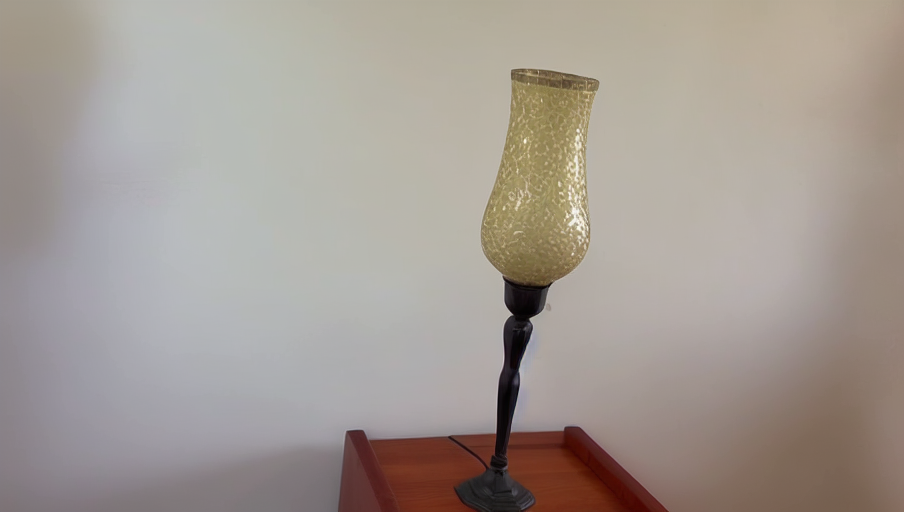} &
    \includegraphics[width=0.19\linewidth]{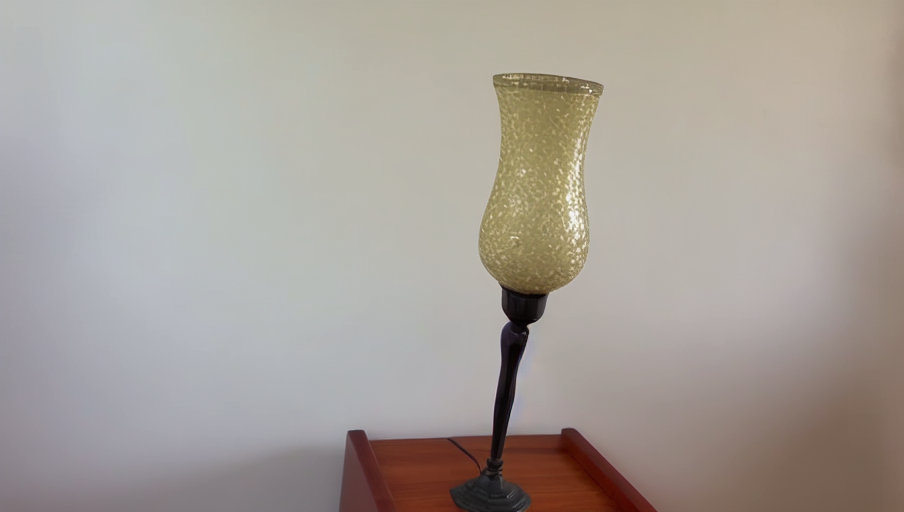} &
    \includegraphics[width=0.19\linewidth]{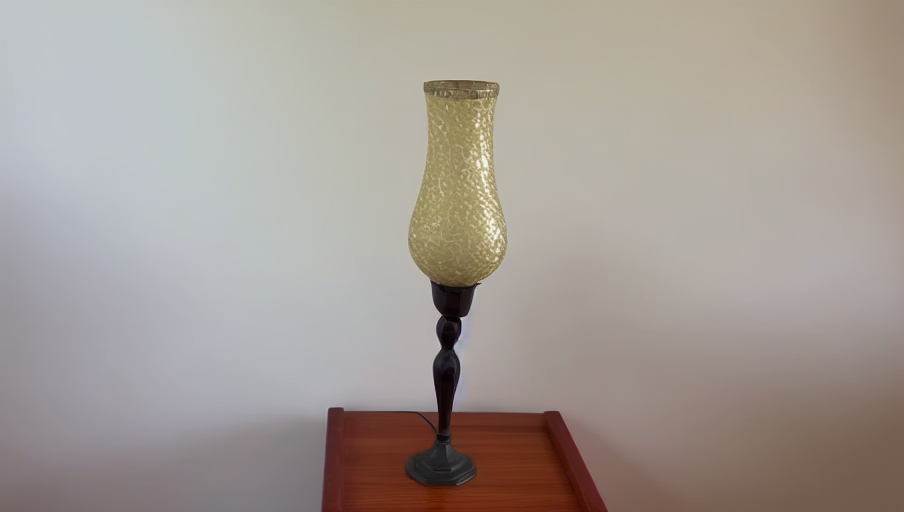} &
    \includegraphics[width=0.19\linewidth]{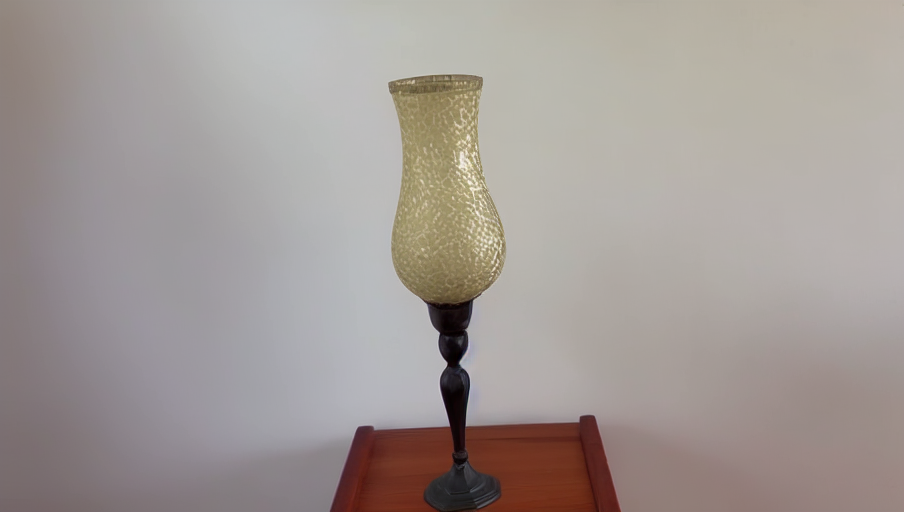} &
    \includegraphics[width=0.19\linewidth]{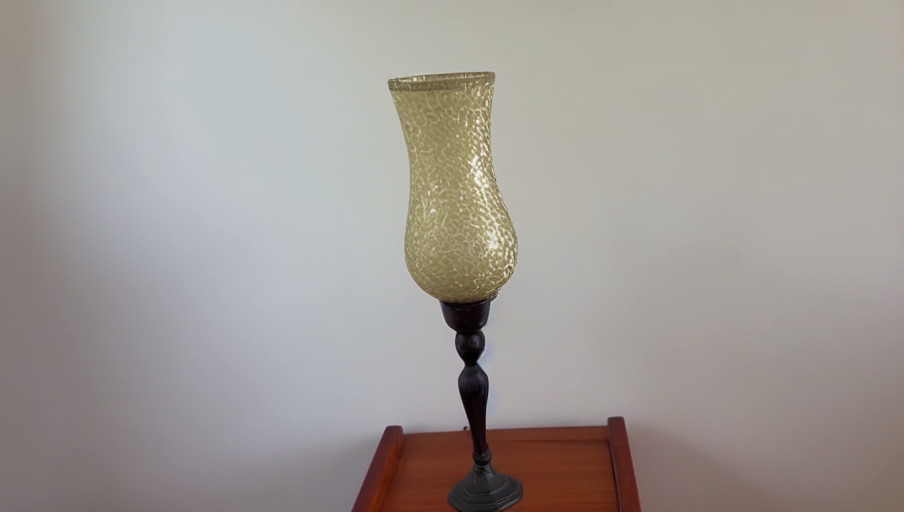} \\

    \raisebox{20pt}{\rotatebox[origin=t]{90}{Ours w/ 100}} &
    \includegraphics[width=0.19\linewidth]{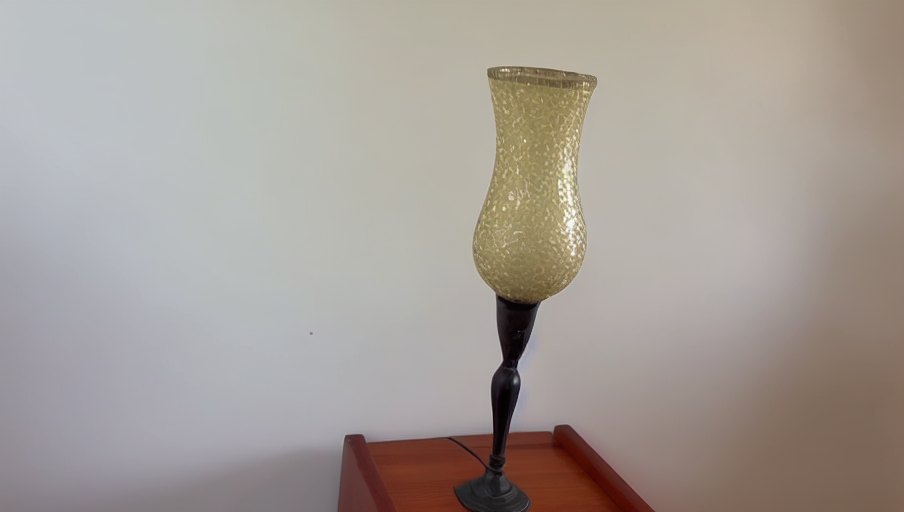} &
    \includegraphics[width=0.19\linewidth]{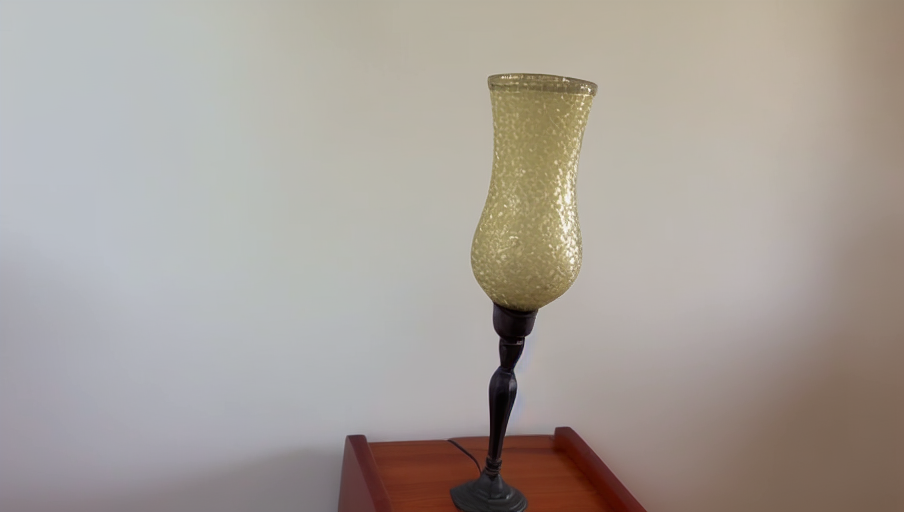} &
    \includegraphics[width=0.19\linewidth]{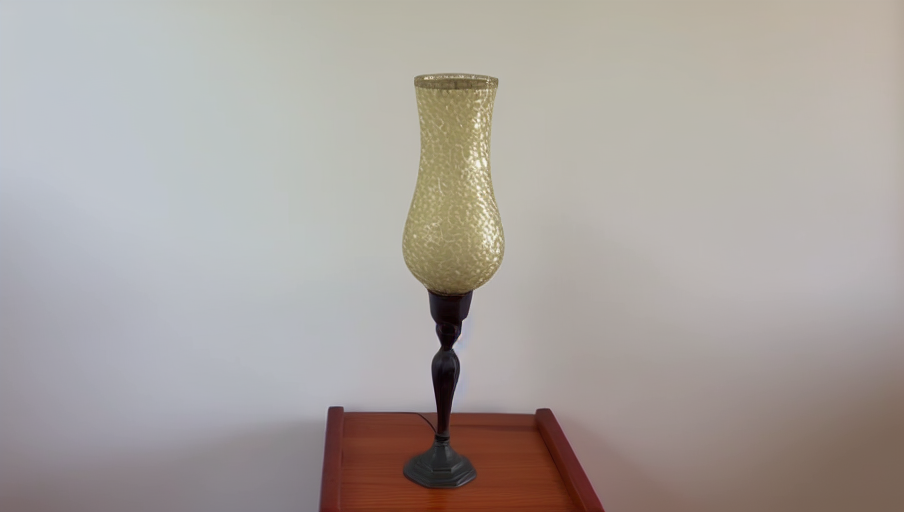} &
    \includegraphics[width=0.19\linewidth]{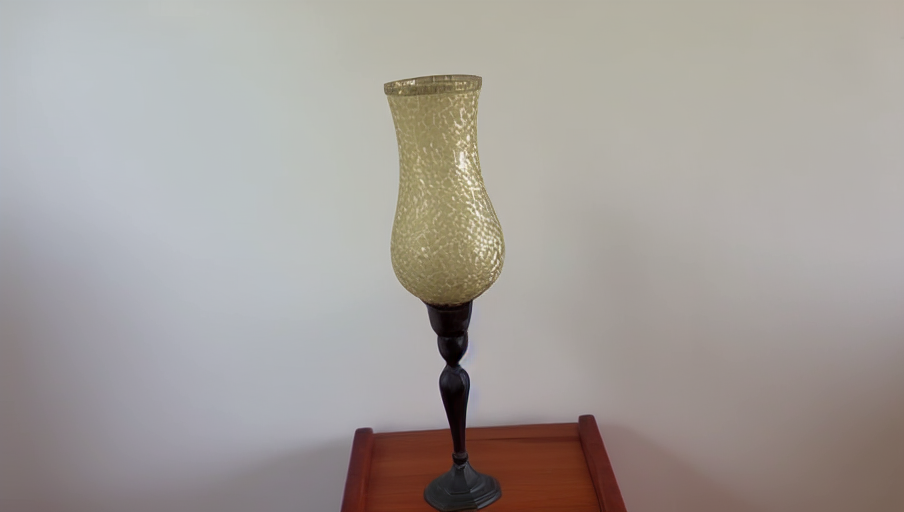} &
    \includegraphics[width=0.19\linewidth]{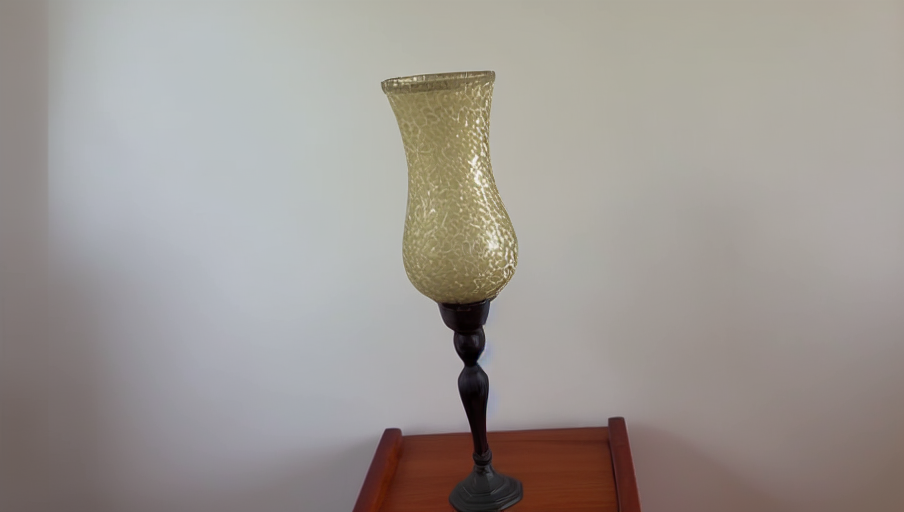} \\

    \raisebox{20pt}{\rotatebox[origin=t]{90}{Ours w/ 150}} &
    \includegraphics[width=0.19\linewidth]{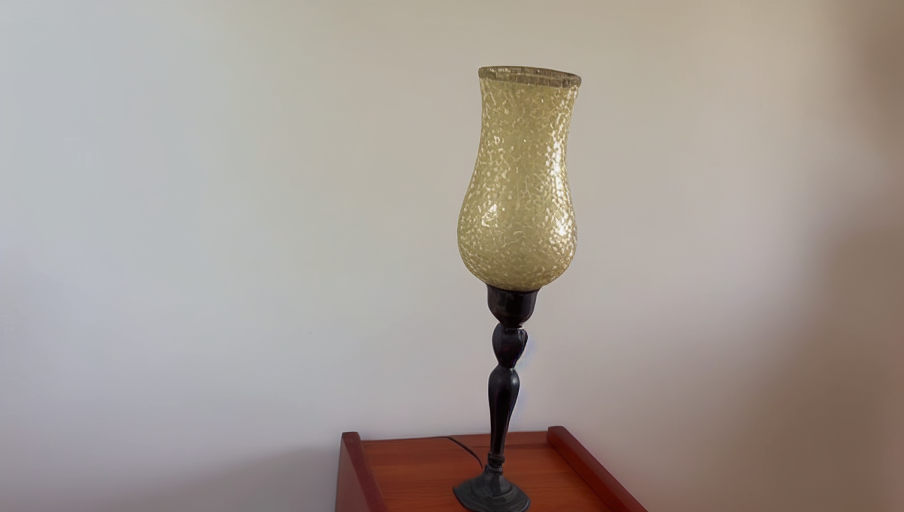} &
    \includegraphics[width=0.19\linewidth]{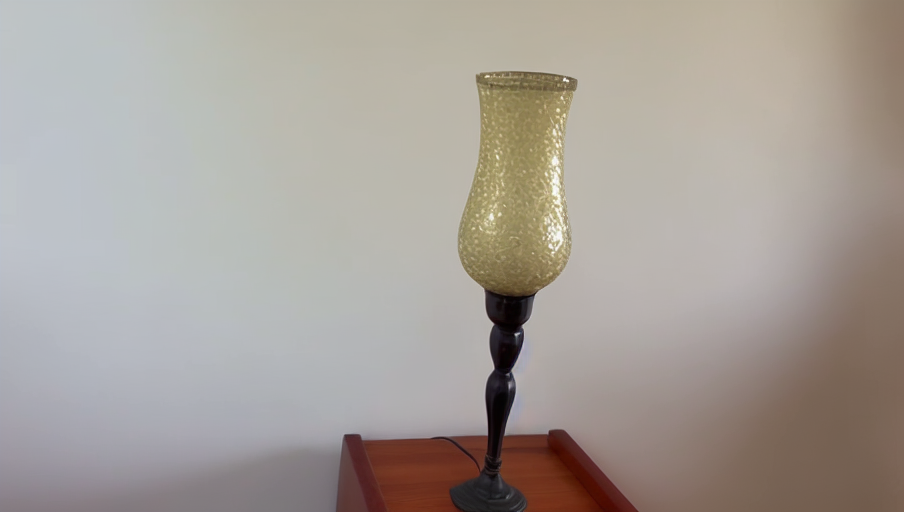} &
    \includegraphics[width=0.19\linewidth]{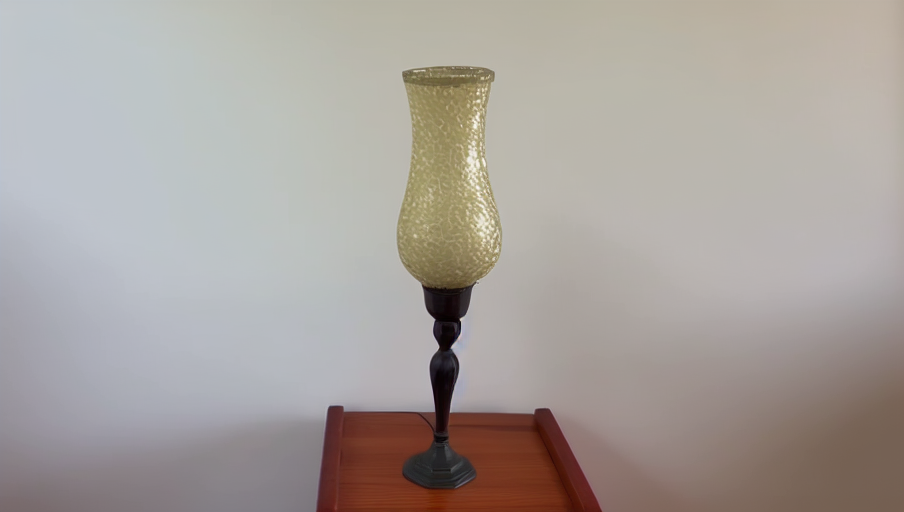} &
    \includegraphics[width=0.19\linewidth]{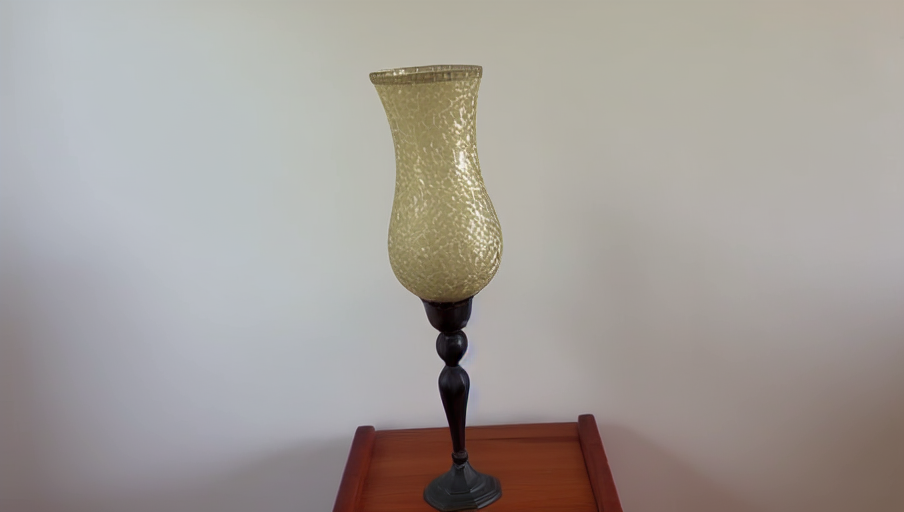} &
    \includegraphics[width=0.19\linewidth]{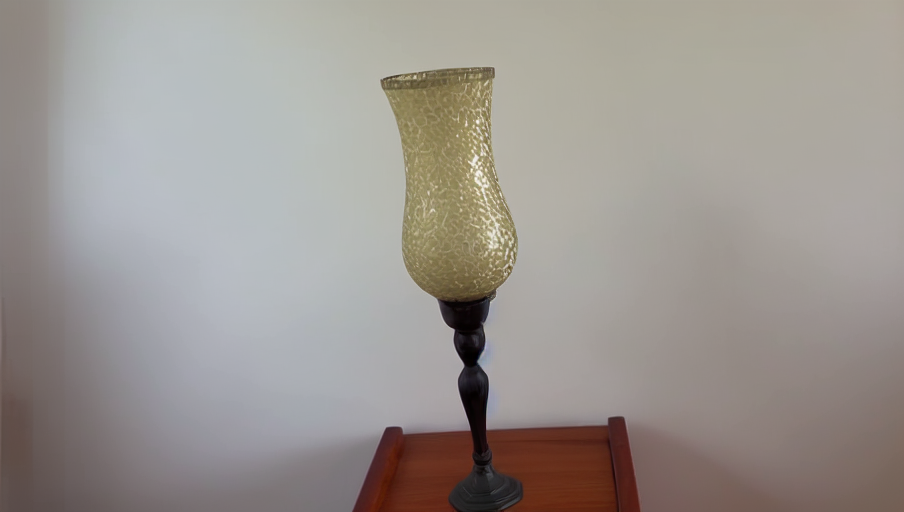} \\

    \raisebox{20pt}{\rotatebox[origin=t]{90}{Ours w/ 200}} &
    \includegraphics[width=0.19\linewidth]{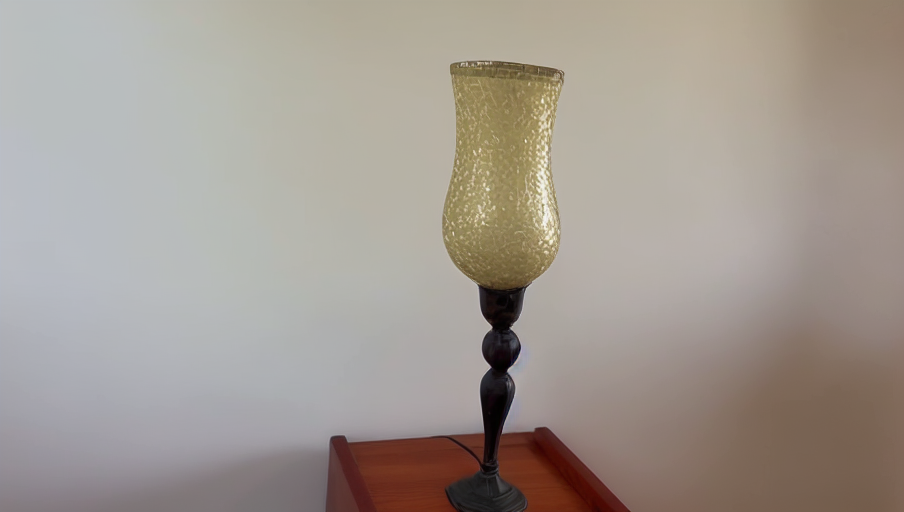} &
    \includegraphics[width=0.19\linewidth]{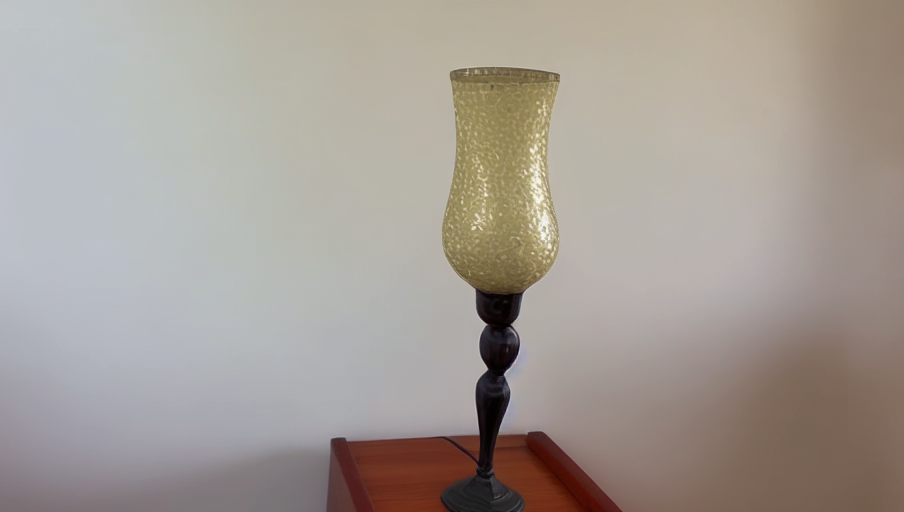} &
    \includegraphics[width=0.19\linewidth]{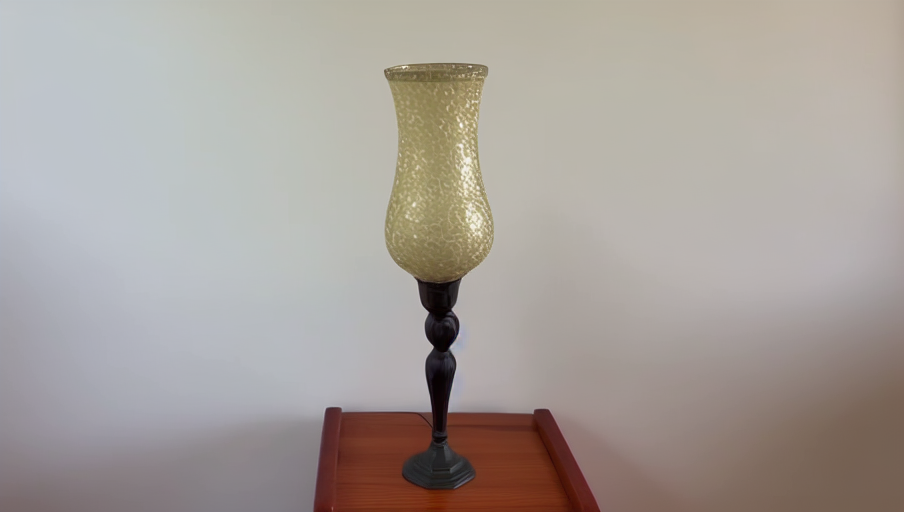} &
    \includegraphics[width=0.19\linewidth]{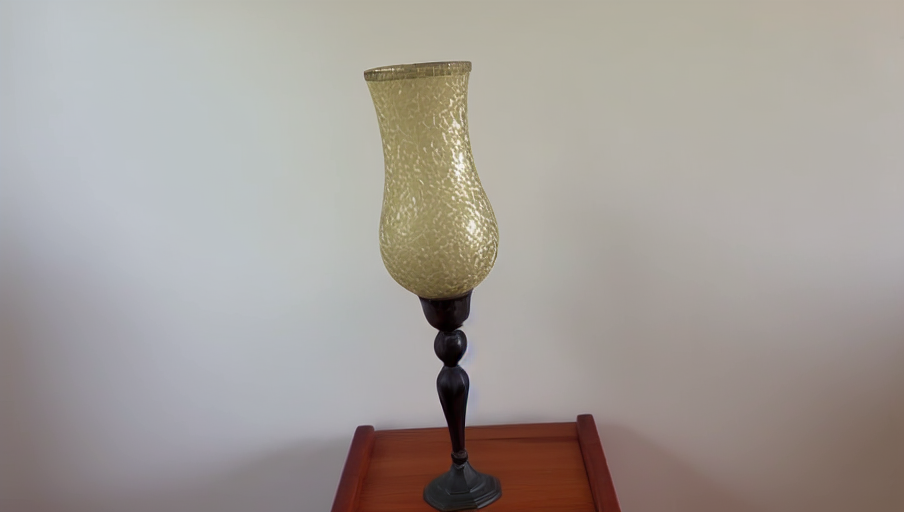} &
    \includegraphics[width=0.19\linewidth]{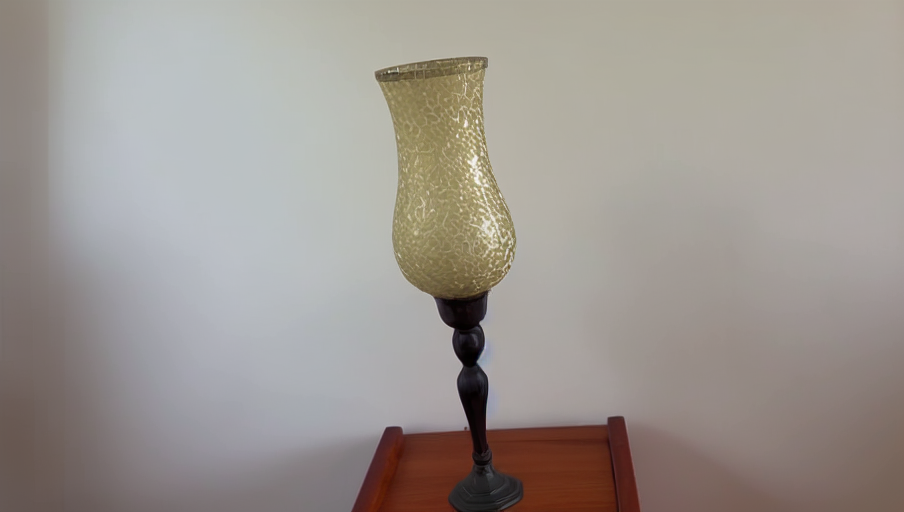} \\

    \raisebox{20pt}{\rotatebox[origin=t]{90}{Ours w/ 250}} &
    \includegraphics[width=0.19\linewidth]{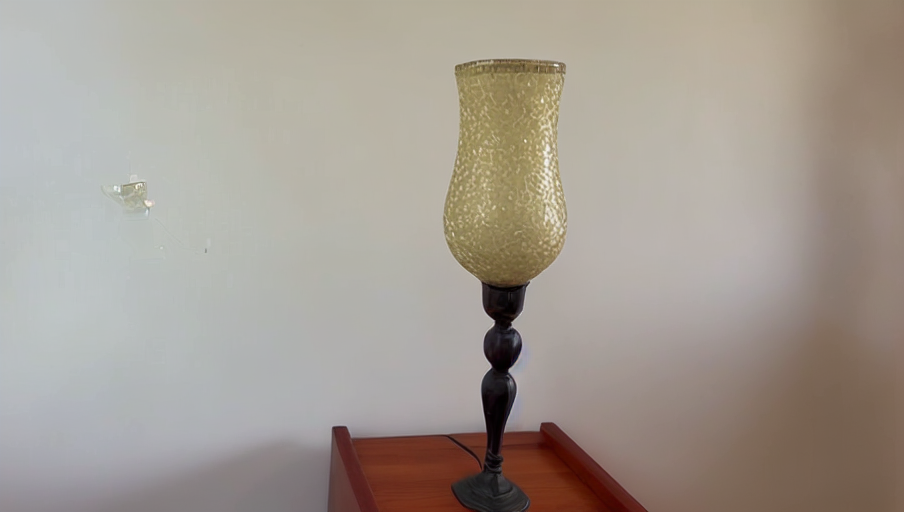} &
    \includegraphics[width=0.19\linewidth]{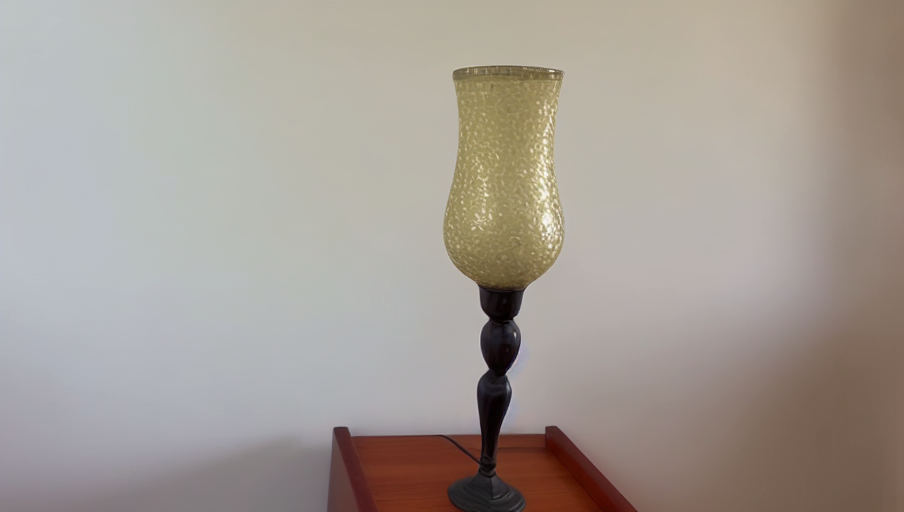} &
    \includegraphics[width=0.19\linewidth]{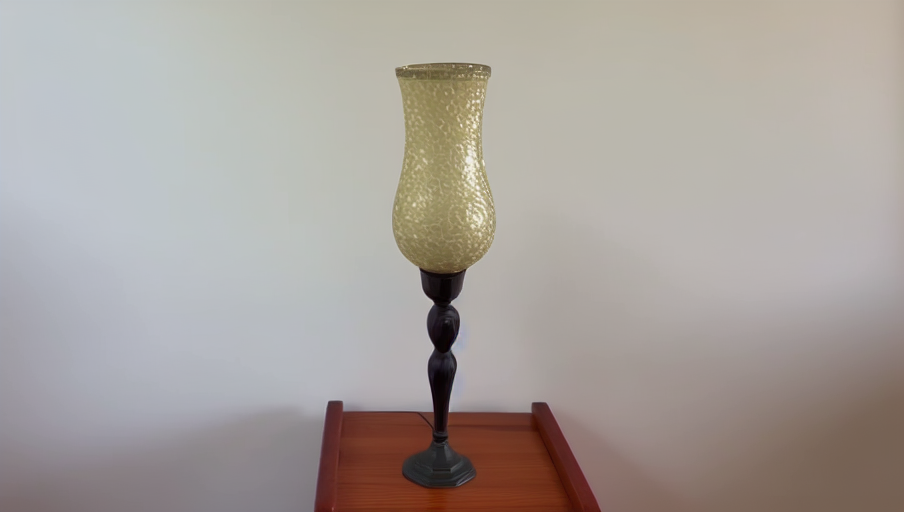} &
    \includegraphics[width=0.19\linewidth]{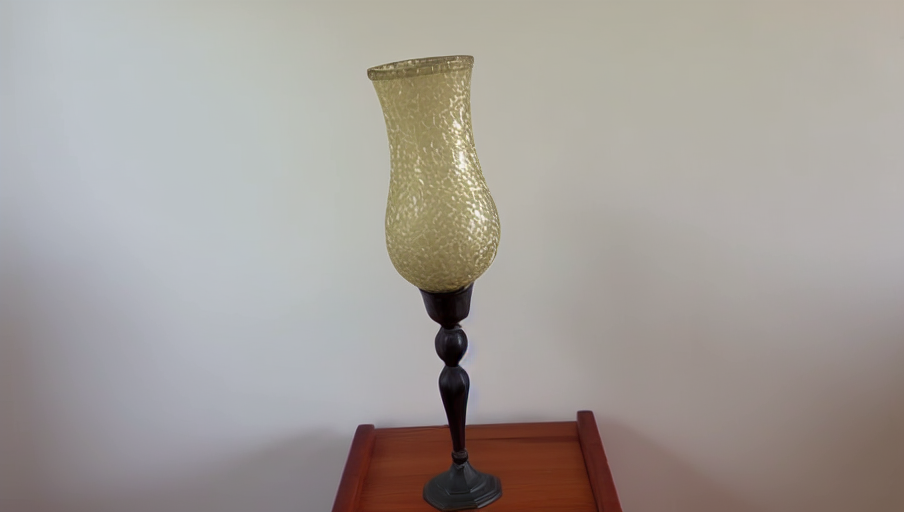} &
    \includegraphics[width=0.19\linewidth]{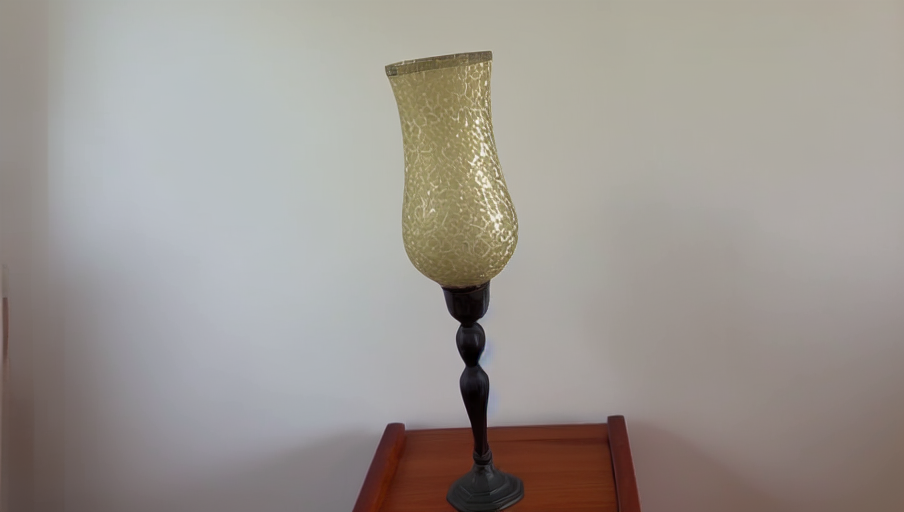} \\

    \raisebox{20pt}{\rotatebox[origin=t]{90}{Ours w/ 305}} &
    \includegraphics[width=0.19\linewidth]{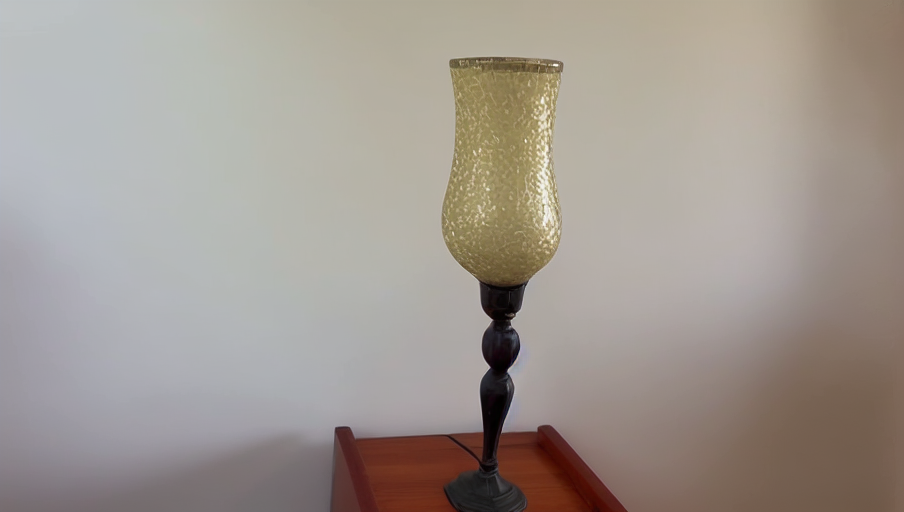} &
    \includegraphics[width=0.19\linewidth]{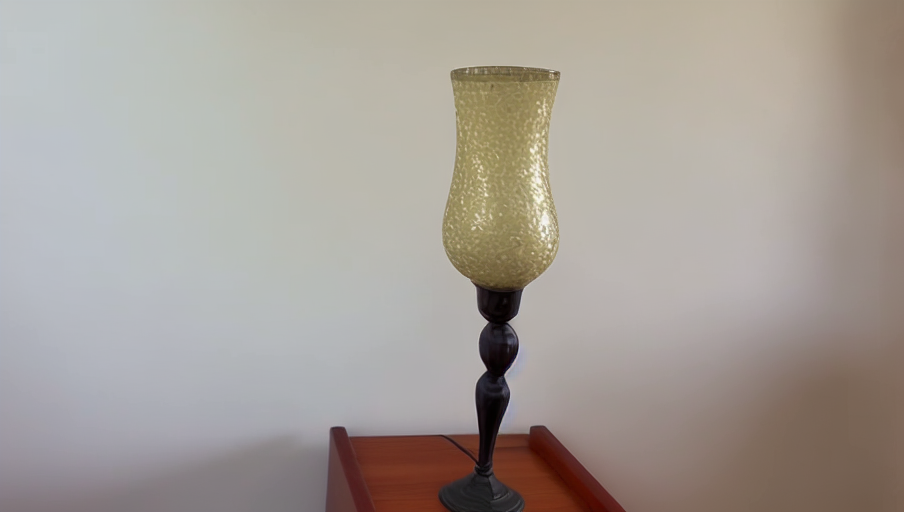} &
    \includegraphics[width=0.19\linewidth]{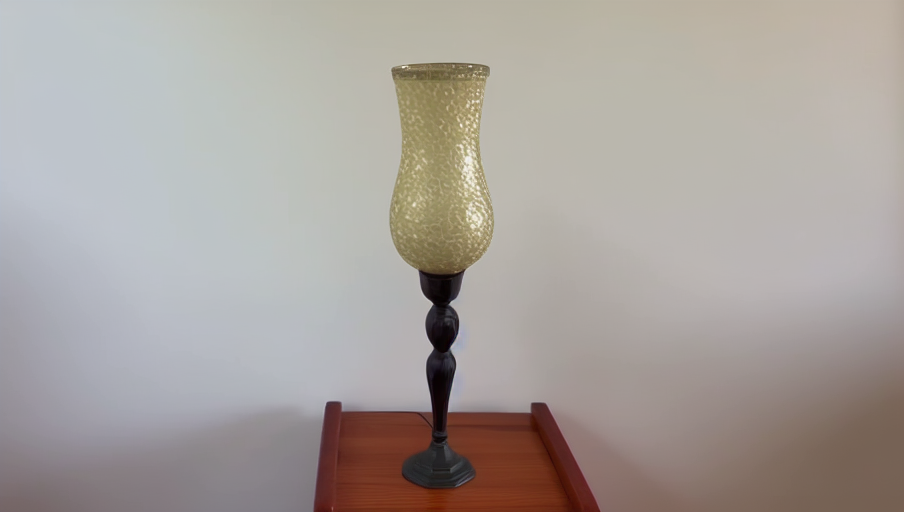} &
    \includegraphics[width=0.19\linewidth]{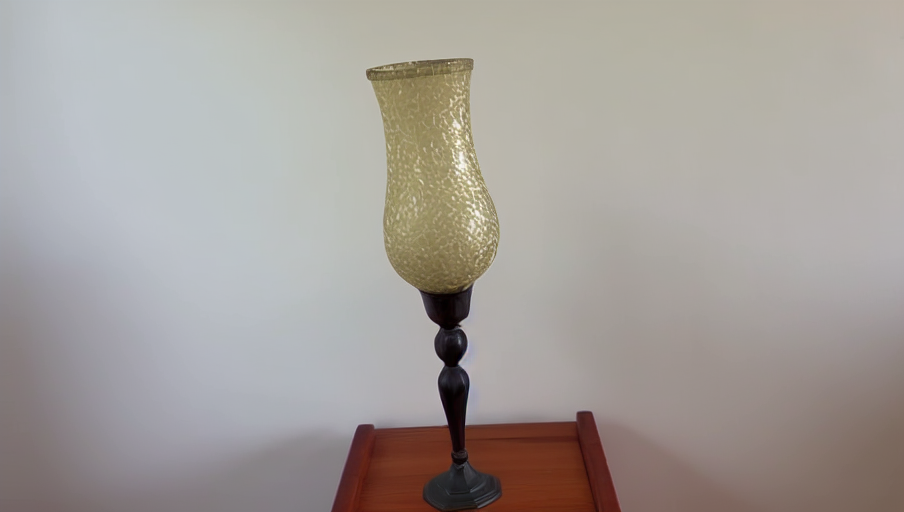} &
    \includegraphics[width=0.19\linewidth]{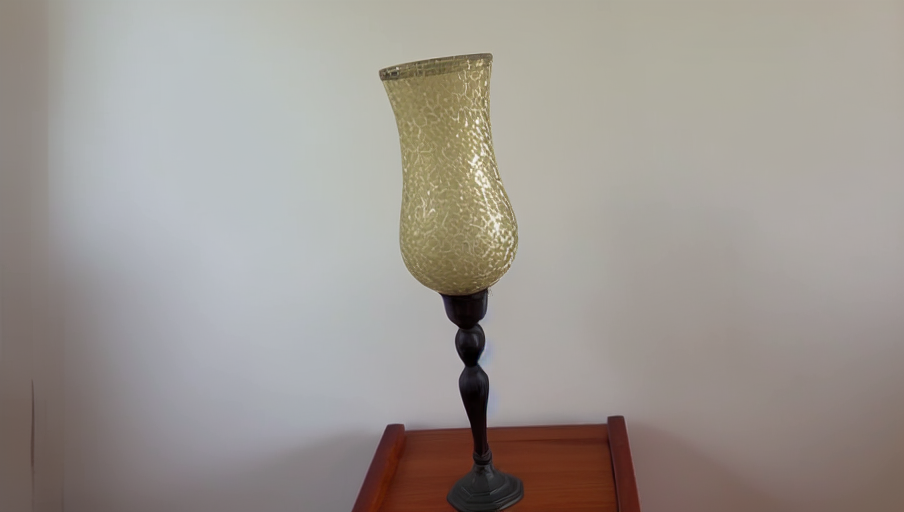} \\

\end{tabular}
\caption{
Evaluating our approach on a varying number of images in the dataset.
}
\label{fig:supp-stress-test}
\end{figure*}

For completeness, we analyze the effect of increasing the input set size on our approach. Specifically, we apply our method for the lamp scene, with sets of sizes 50, 100, 150, 200, 250, 305. 
To create these sets, we randomly sample images from the original set.
The results are shown in Figure~\ref{fig:supp-stress-test}.
As can be seen in the results, even with a set of 50 images we are able to achieve much more consistent results compared to MasaCtrl~\cite{cao2023masactrl} which edits each image in isolation. Our results are improved when increasing the dataset size, and are highly consistent with a set containing 200 images.

\section{Limitations}

\begin{figure}[tbh]
\centering
\scriptsize{
\setlength{\tabcolsep}{1pt}
\begin{tabular}{c c c c c}

    \includegraphics[height=0.32\linewidth]{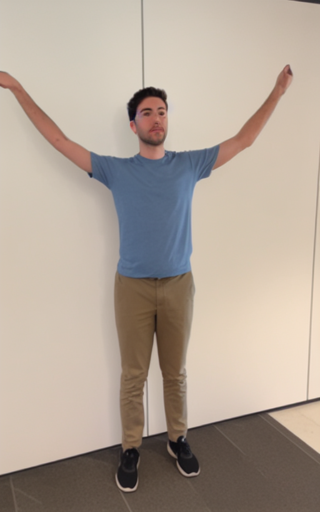} &
    \includegraphics[height=0.32\linewidth]{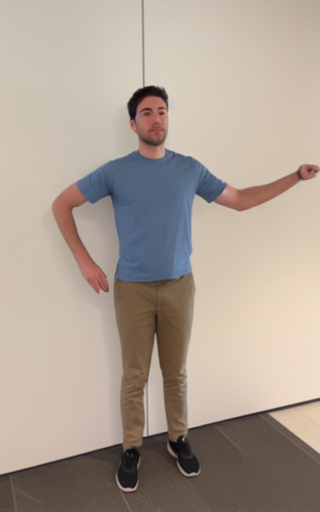} &
    \includegraphics[height=0.32\linewidth]{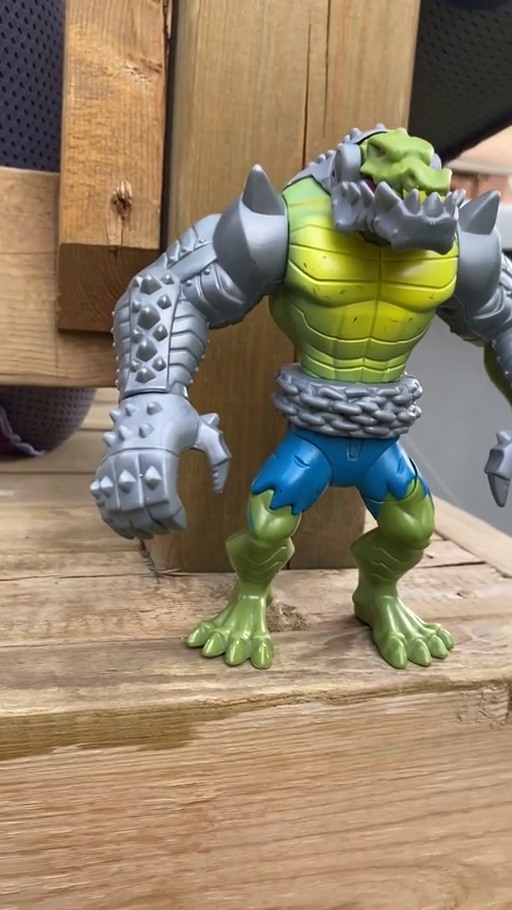} &
    \includegraphics[height=0.32\linewidth]{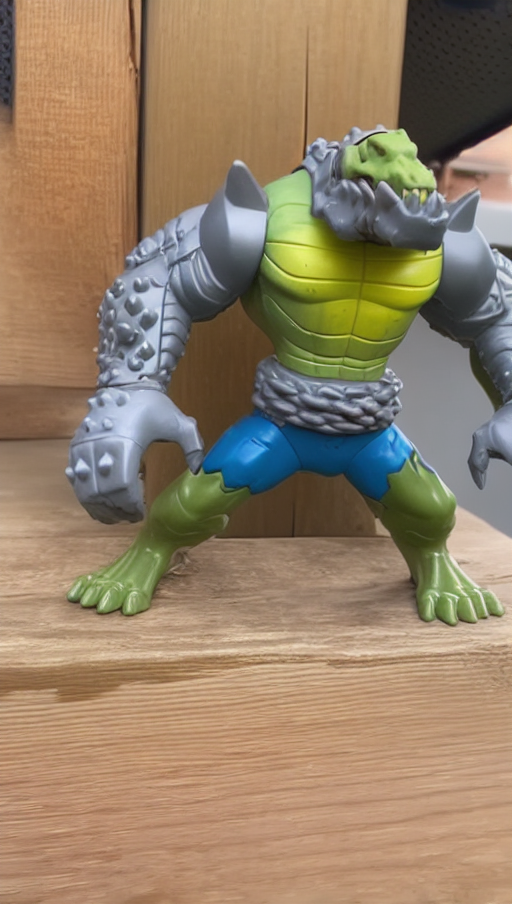} &
    \includegraphics[height=0.32\linewidth]{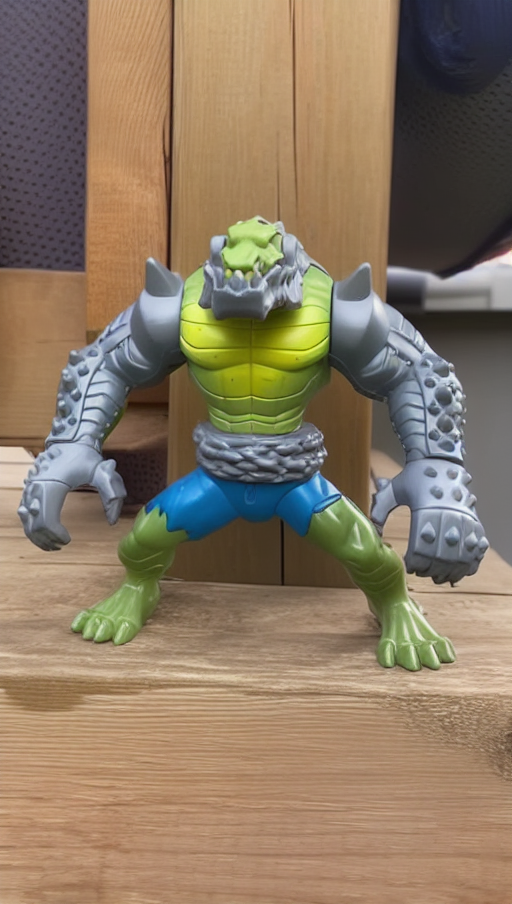} \\

    Edit 1 & Edit 2 & Original image & Edited view 1 & Edited view 2

\end{tabular}
}
\caption{
Demonstrating the limitations of our method.
}
\label{fig:supp-limit}
\end{figure}

As discussed in the paper, our method has several limitations. We demonstrate them in Figure~\ref{fig:supp-limit}. 

First, as can be seen in the two edited images of the person, his hand does not have fingers, and the palm is of low visual quality.

The alligator toy is high-detailed, as can be seen by the texture of his different body parts (\eg, arms and face). Our method struggles to make all these fine details consistent, as can be seen in the number of fingers in the toy's feet. In the original image, the toy has four fingers in the right foot. While in the edited image of view 2 our method generates four fingers in this foot, it generates five fingers in view 1. This may be a resolution problem, as we ensure the consistency of the queries which are lower resolution from the generated image itself. In addition, it can be seen that the deck behind the toy looks different between the original and edited images.

\section{Running Time}
The running time of our method is linear in the number of images in the multi-view set. The typical running time of the entire pipeline for a dataset of 300 images is roughly 8.5 hours on an NVIDIA RTX A5000 GPU.
This includes the time to invert the images, the denoising of the multi-view set, and the training of the QNeRFs along the process.

In comparison, IN2N-CN~\cite{instructnerf2023, zhang2023adding} takes roughly 5.5 hours on the same dataset and GPU. Note that IN2N-CN is slower than the vanilla IN2N, because MasaCtrl~\cite{cao2023masactrl} requires DDIM inversion~\cite{song2021denoising} in each dataset update iteration.

CSD-CN~\cite{kim2023collaborative} takes about three days on an NVIDIA H100 GPU with the same dataset.

\end{document}